\documentclass[a4paper,preprint,12pt]{elsarticle}

\usepackage{amssymb,amsmath}
\usepackage{epsfig}
\usepackage{epsf}
\usepackage{float}
\usepackage{a4wide}
\usepackage{graphpap}
\usepackage{setspace}
\usepackage{color}
\usepackage[ruled,titlenotnumbered]{algorithm2e}

\usepackage{latexsym}
\usepackage{graphicx}
\usepackage{array}
\usepackage{hhline}
\usepackage{fullpage}          
\usepackage{epsfig}

\usepackage[latin1]{inputenc}
\usepackage{mathrsfs}
\usepackage{amssymb}
\usepackage{verbatim}
\usepackage{amsmath}

\usepackage{subfigure}

\usepackage{caption}
\usepackage{multirow}

\def\RR{\mathbb R}

\journal{Computer Vision and Image understanding}
\begin{document}

\begin{frontmatter}

\title{Unsupervised edge map scoring: a statistical complexity approach.
}

\author[a]{Javier Gimenez}
\ead{jgimenez@mate.uncor.edu}
  \author[b]{Jorge Martinez}
  \ead{martinez@uns.edu.ar}
\author[a]{Ana Georgina Flesia}
\ead{flesia@famaf.unc.edu.ar}

\address[a]{Facultad de Matem\'atica, Astronom\'\i a y F\'\i sica, Universidad Nacional de C\'ordoba.  Ing. Medina Allende s/n, Ciudad Universitaria CP 5000, C\'ordoba, Argentina. }
\address[b]{Departamento de Matem\'atica, Universidad Nacional del Sur,  Avenida Alem 1253. 2do Piso
Bah\'{\i}a Blanca, B8000CPB, Argentina.}

\begin{abstract}

We propose a new Statistical Complexity Measure (SCM) to qualify edge maps without Ground Truth (GT) knowledge. The measure is the product of two indices, an \emph{Equilibrium} index $\mathcal{E}$  obtained by projecting the edge map into a family of edge patterns, and an \emph{Entropy} index $\mathcal{H}$, defined as a function of the Kolmogorov Smirnov (KS) statistic.

This new measure can be used for performance characterization which includes: (i)~the specific evaluation of an algorithm (intra-technique process) in order to identify its best parameters, and (ii)~the comparison of different algorithms (inter-technique process) in order to classify them according to their quality.

 Results made over images of the South Florida and Berkeley databases show that our approach significantly improves over Pratt's Figure of Merit (PFoM) which is  the objective reference-based edge map evaluation standard,  as it takes into account more features in its evaluation.


\end{abstract}
\end{frontmatter}

\section{Introduction}
In most image processing techniques,  the detection and handling of the edge structure of the input image is very important.
From object detection to image transmission, the quality of edge manipulation takes great part in the success of the processing.
Nevertheless, there is no universal definition of the notion of edge.
For  Abdou and Pratt, an edge is defined as a local  change in luminance  or discontinuity in the luminance intensity of the image \cite{Abdou1979} while Kitchen and Rosenfeld  pointed out that the edge concept depends on the type of processing and analysis in which it is involved  \cite{Kitchen1981}.  

Therefore, many  researchers have designed optimal  Edge Detection Algorithms (EDA) related to different properties of the edge structure, but only a few have studied how to measure the edge strength and quality of general edge maps~\cite{Papari2011}. Effective and objective Edge Detection (ED) evaluation measures must be developed in order to assess EDA performance.

In general, ED evaluation measures can be classified due to the need of a reference map called Ground Truth (GT) (supervised  or unsupervised measures) and the type of score that they output, quantitative or qualitative.
Some well known examples of quantitative supervised measures, also called  discrepancy measures, are Pratt's Figure of Merit (PFoM) \cite{Pratt1978}, Kappa index \cite{Cohen1960},  and  Baddeley's Delta Metric (BDM) \cite{Baddeley1992}. A comparison between these discrepancy measures and some other supervised statistical measures were performed in \cite{Lopez_Molina}. Two main conclusions were drawn from their experiments: i) up to date, there is no convincing solution for edge image comparison or quality evaluation;  ii) the biases of the measures can be helpful in applications where there is a particular interest in penalizing or ignoring some specific kind of error. A supervised quality metric for binary documents based on structural pixel matching, taking into account global edge structures introducing a smoothness term in the matching function was proposed in \cite{Jang}. Examples of binary documents are text files either photocopied, faxed or scanned, with fast publishing resolution. In this kind of binary documents, bad visual word understanding is not always related to classical low scoring. PFoM is known to give high scores to lighter maps, with high rate of false negatives, but it is not acquiescent to human perception~\cite{Fernandez2004}.

 Without the guide of a GT, assessing edge maps quality is a more difficult task. The unsupervised ED measures that are found in the literature look for specific characteristics of the input edge map, such as coherence,  \cite{Bryant1979}, continuity, \cite{Kitchen1981}, smoothness and good continuation, \cite{Zhu1996, Heath1997},  or an specific pattern identification, \cite{Bernsen1991}, among others \cite{Nercessian2009, Konishi2003}.  Bower et al. studied the bias introduced by the search of only one characteristic~\cite{Bowyer1999}. They reported a similar conclusion to the one given by \cite{Lopez_Molina}:  there is no unique solution; moreover, selected best maps are qualitatively different,  and bias can not be estimated without further assumption of the error incurred.

  Recently, Yitzhaky and Peli  proposed an unsupervised evaluation procedure of ED techniques based on the consensus approach \cite{Yitzhaky2003}. Using the correspondence between  different standard EDA results, an estimated best edge map (\emph{consensus map}) was
obtained and later used as an estimated ground truth (EGT). Correspondence was computed by using both a receiver operating characteristics (ROC)
analysis and a Chi-square test for standard binary outputs, considering a trade off between structure and
noisiness in the detection results.  Fernandez-Garcia et al. provided a definition of \emph{consensus edge map} that is close to the notion of confidence set.  They argued that in order to compare ED procedures, it is not esencial to use the best and exact GT; rather it is only necessary to use a reliable EGT that allows correct classification or ranking of the EDA to be obtained, \cite{Fernandez2008}.
They also noted that their approach may be used to evaluate detections from different EDA (inter-technique performance characterization) only if these detectors aim at the same output format. Our proposed evaluation methods also take into account this assumption.

The consensus approach suffers from bias regarding the generation of the candidate edge maps used to define the EGT. If the majority of the edge maps
considered are not of adequate quality or fail to extract certain edge structures which are detected by only a small selection of the edge maps, this will be reflected both in the consensus EGT and in the quality of the evaluation methodology derived from it. In a sense, it penalizes algorithms that do not agree with the failures of the other algorithms.

 In this paper we define a new non-reference measure that does not depend, directly or indirectly, on GT data. As the previous measures, it can be used for ED performance characterization which includes: (i) the specific evaluation of an algorithm (intra-technique process) in order to identify its best parameters, and (ii) the comparison of different algorithms with the same output format (inter-technique process) in order to classify them according to their quality.

 Our proposal, denoted Statistical Complexity Measure (SCM)  searches for a compromise between two extreme values in the space of edge maps: a map with few edge points in a  perfect shape (\emph{Equilibrium}) and with  many edge points randomly located (\emph{Information}).
 The new measure is the product of two indices, an \emph{Equilibrium} index $\mathcal{E}$  obtained by combining  local correlation between the edge map and a family of predetermined edge patterns and an \emph{Entropy} index $\mathcal{H}$, defined as a function of the Kolmogorov Smirnov (KS) statistic. SCM gives value between zero and one, being  zero the minimum and one the maximum quality.

Konishi et al. defined an statistical ED algorithm which relies on Chernoff information and entropy of  probability distributions conditional to edge and non-edge state \cite{Konishi2003}. Their validation experiments studied elements similar to the indices that are part of our Complexity Measure. They also noted that maps with scattered random points may give high information regardless the real structure of the image, but the combinations of shape seeking measures with entropy functionals greatly reduces  the probability of such anomalies.



The paper is organized as follows. In Section \ref{supervizado}, a cosine based discrepancy measure $\mathcal{Q}_{\mathfrak{B}}$ to score a map against a collection of hand-made GT (supervised case), or against a collection of fixed significant patterns (unsupervised case) is introduced. In Section \ref{no_supervizado} the concepts behind the  \emph{Equilibrium} index $\mathcal{E}$ and \emph{Entropy} index $\mathcal{H}$ are introduced, and  the final SCM $\mathcal{C}$ as the product of both indices is defined. Experiments and results are discussed in Section \ref{experimentos} and conclusions and comments are left for Section  \ref{conclusion}.

\section{Cosine-based Similarity Measure (CSM).}
\label{supervizado}

 In this section, a cosine-based similarity measure $\mathcal{Q}_{\mathfrak{B}}$ is introduced as an intermediate step in the definition of the final measure $\mathcal{C}$, along with necessary notation.
\begin{figure}[h]
 \[\begin{array}{cccc}
 \includegraphics[scale=0.15]{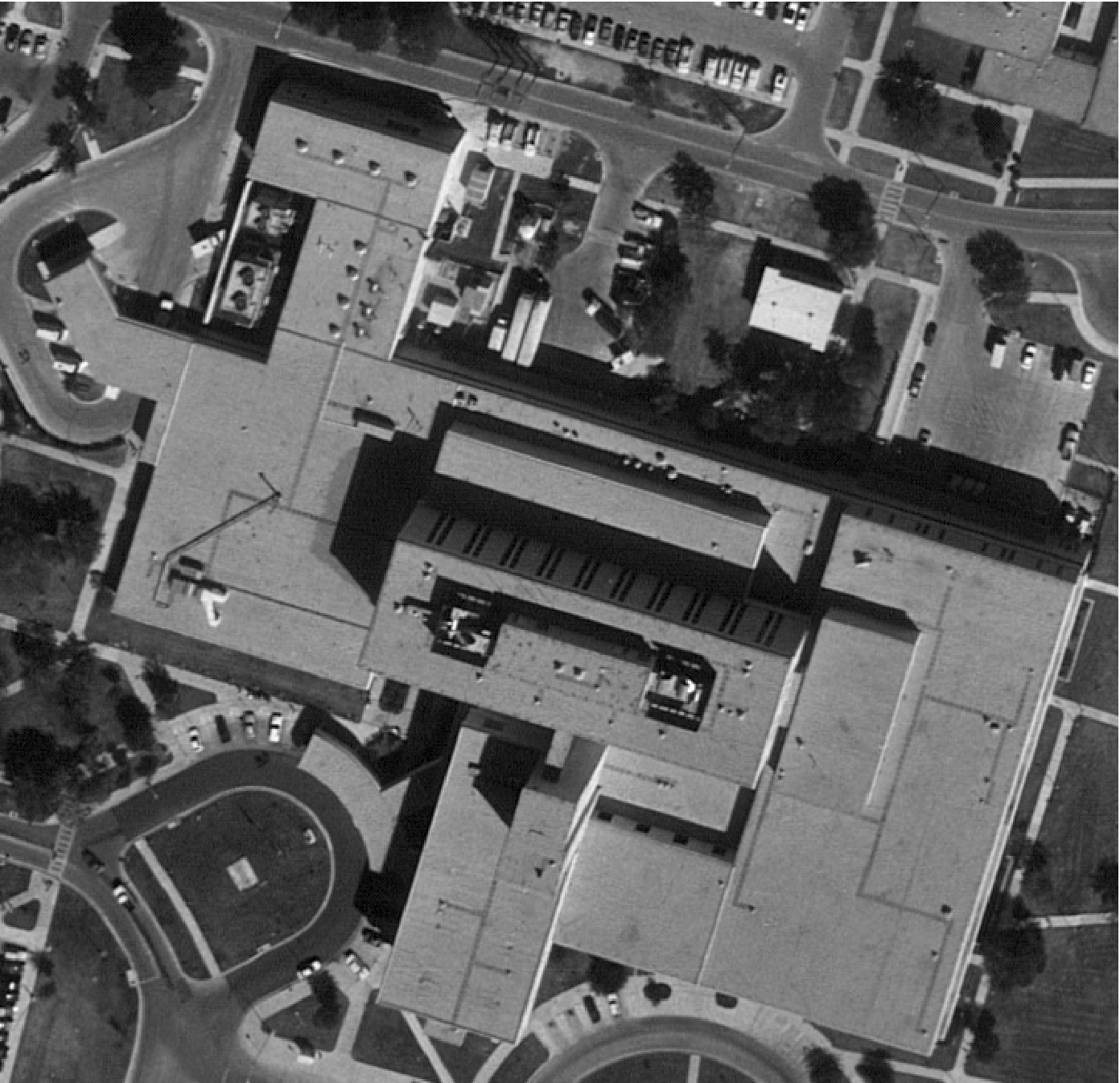}&\includegraphics[scale=0.225]{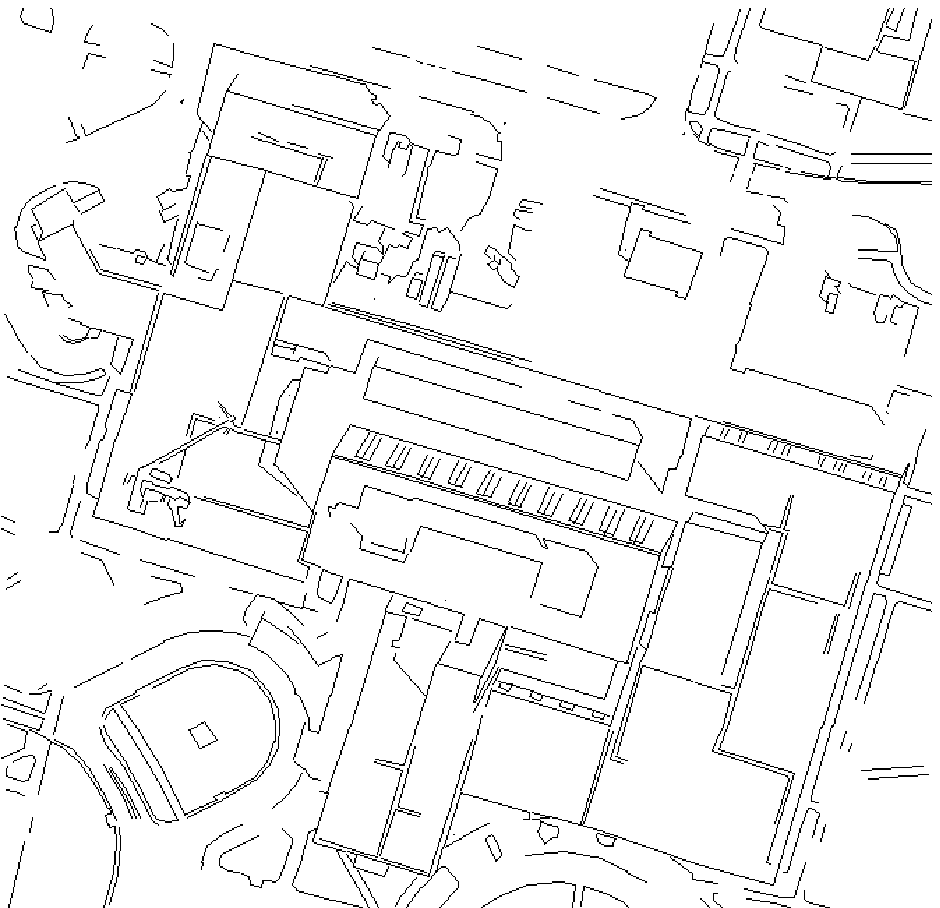}&\includegraphics[scale=0.225]{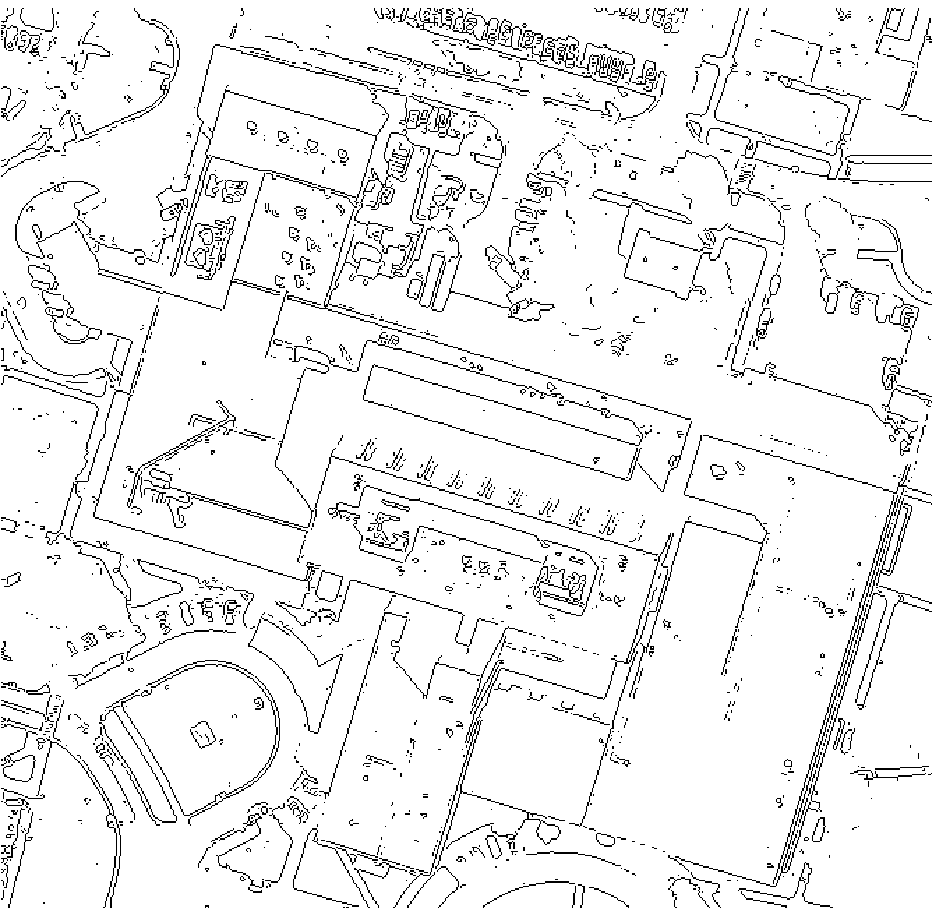}\\
(a)&(b)&(c)\\
 \end{array} \]
 \vspace{-0.5cm}
\caption{Evaluation of CSM $(a)$ Original \emph{Large Building} image from South Florida database , $(b)$ GT, $(c)$ Sobel ED output (thresholding parameter $T=0.068$) with $\mathcal{Q}_{\mathfrak{B}}=0.3976$.}
 \label{grafico_ejemplo_supervisado}
\end{figure}
Let $I$ be an image, $\textbf{b}$ an edge map associated with $I$, (this is, $\textbf{b}$ is a binary image with the same size as $I$), and $\textbf{g}$ its GT (if it is available). Figure \ref{grafico_ejemplo_supervisado} shows an example of such images. A simple measure of similarity between $\textbf{b}$ and $\textbf{g}$ is the cosine of the angle  between them~\cite{Theodoridis2008}, when they are seen as 1-D vectors (concatenating all columns one under another).

 We define the index $\mathcal{Q}_{\mathfrak{B}}$  as the maximum of all similarity values
\begin{equation}\label{def_q-NS_muchasGT}
\mathcal{Q}_{\mathfrak{B}}\left(\textbf{b}\right) =\max_{1 \leq i \leq n}\mathcal{Q}(\textbf{g}_{i},\textbf{b})
=\max_{1 \leq i \leq n}\frac{\textbf{g}^{T}_{i}\textbf{b}}{\left\|\textbf{g}_{i}\right\| \left\|\textbf{b}\right\|}\;.
\end{equation}
being $\left\|\textbf{x}\right\| =\sqrt{\textbf{x}^{T}\textbf{x}} $, and   $\mathfrak{B}=\left\{\textbf{g}_{1},\textbf{g}_{2},\ldots,\textbf{g}_{n}\right\}$ a collection of GT images (that could have only one element).

The Cauchy Schwartz inequality implies that the index is upper bounded by one thus attaining such bound only when the map is optimal $\left(\textbf{b} \in \mathfrak{B}\right)$. Since edge maps and GT images are binary images, the index is lower bounded by 0, attained only  in the absence of any similarity (when $\textbf{b}$ is orthogonal to $\mathfrak{B}$).  When no GT is available, predefined edge patterns may be  locally sought in the edge map with this measure.

\section{A statistical complexity measure (SCM)}
\label{no_supervizado}

In this section, a new SCM in the context of unsupervised evaluation of edge maps is introduced. This new measure can be used to identify the optimal parameters of a given algorithm, but also to compare and rank the results of different algorithms. 
 In this framework, the concepts of \emph{Equilibrium} and \emph{Information} can be discussed and scoring indices for such qualities in edge maps can be proposed.

 Following the general structure of complexity measures described by \cite{Lopez1995}, SCM  is defined as
\begin{equation}
\mathcal{C}(\textbf{b})=\mathcal{E}(\textbf{b})\mathcal{H}(\textbf{b}),
\end{equation}
where $\mathcal{E}$ is an \emph{Equilibrium} index,  $\mathcal{H}$  is an \emph{ Entropy} index, and $\textbf{b}$ is an  edge map.
To define such indices,  the concepts of \emph{Equilibrium} and \emph{Information} in the context of ED must be discussed.
\begin{figure}[h]
 \[\begin{array}{cccc}
\includegraphics[scale=0.21]{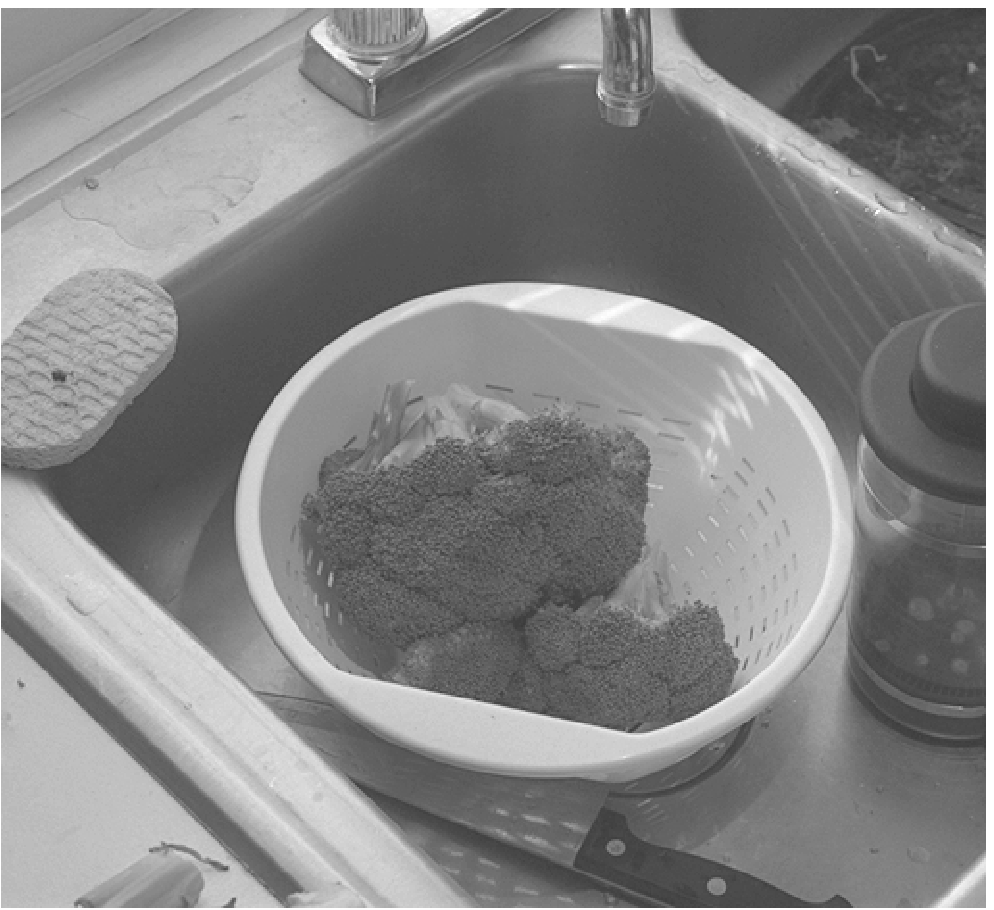}&\includegraphics[scale=0.22]{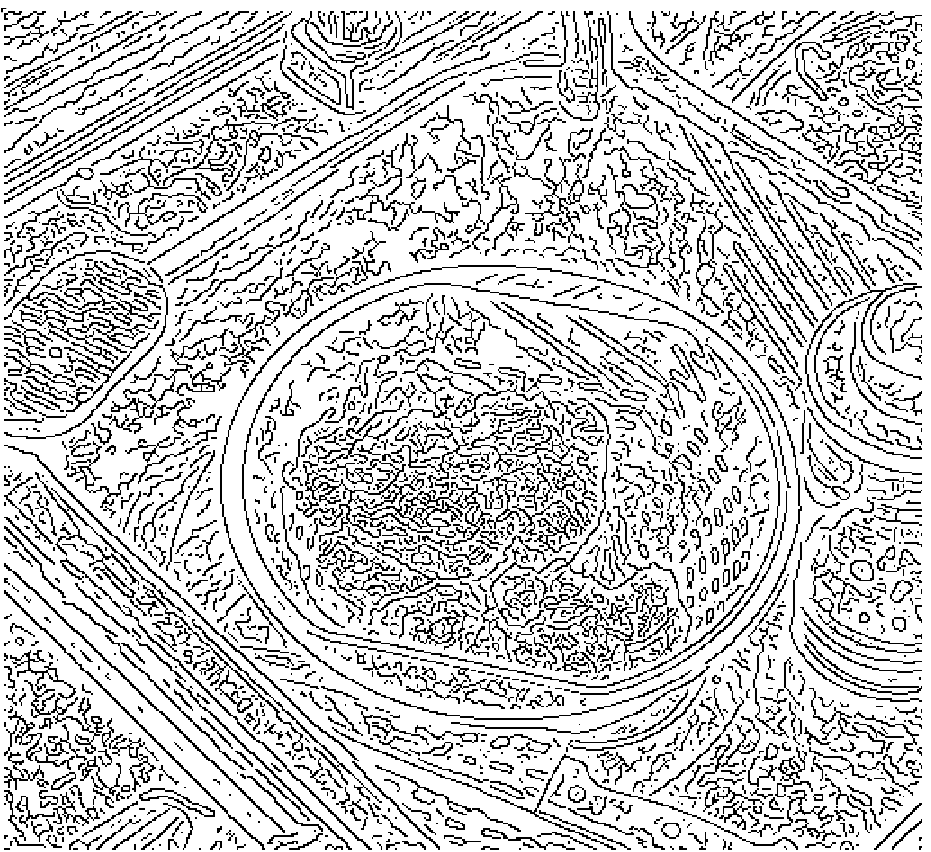}&
\includegraphics[scale=0.2]{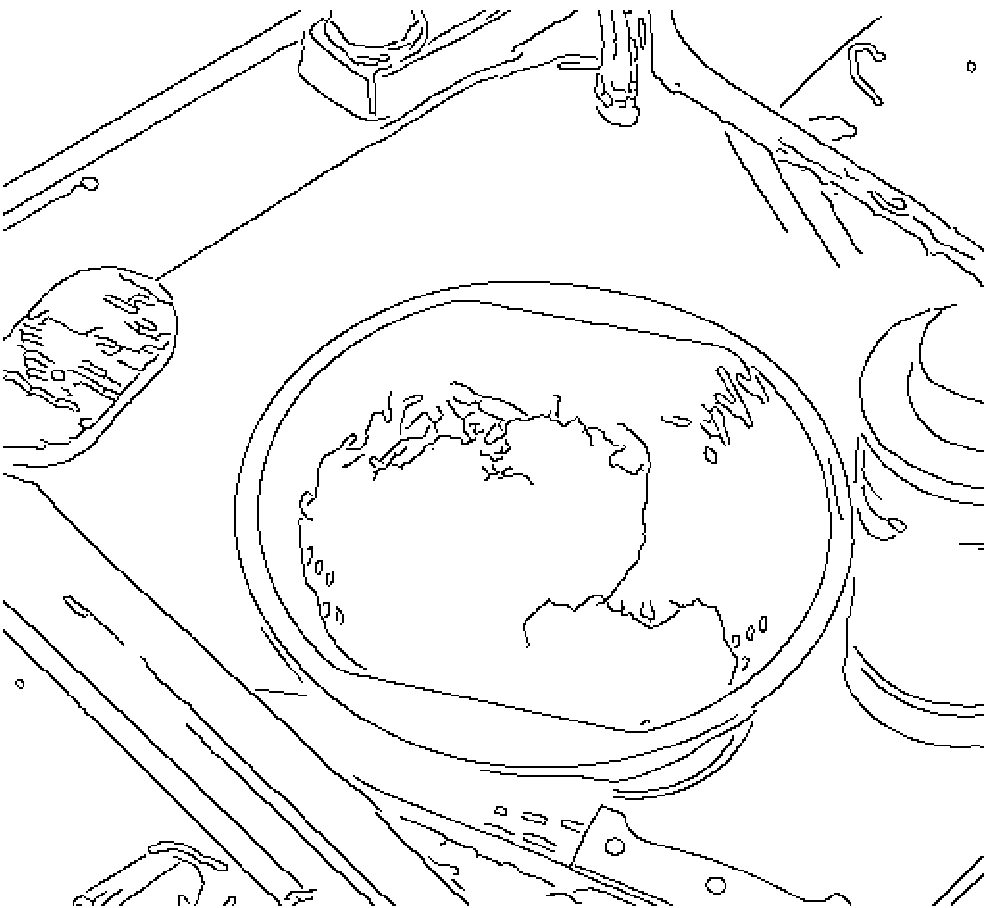}&\includegraphics[scale=0.2]{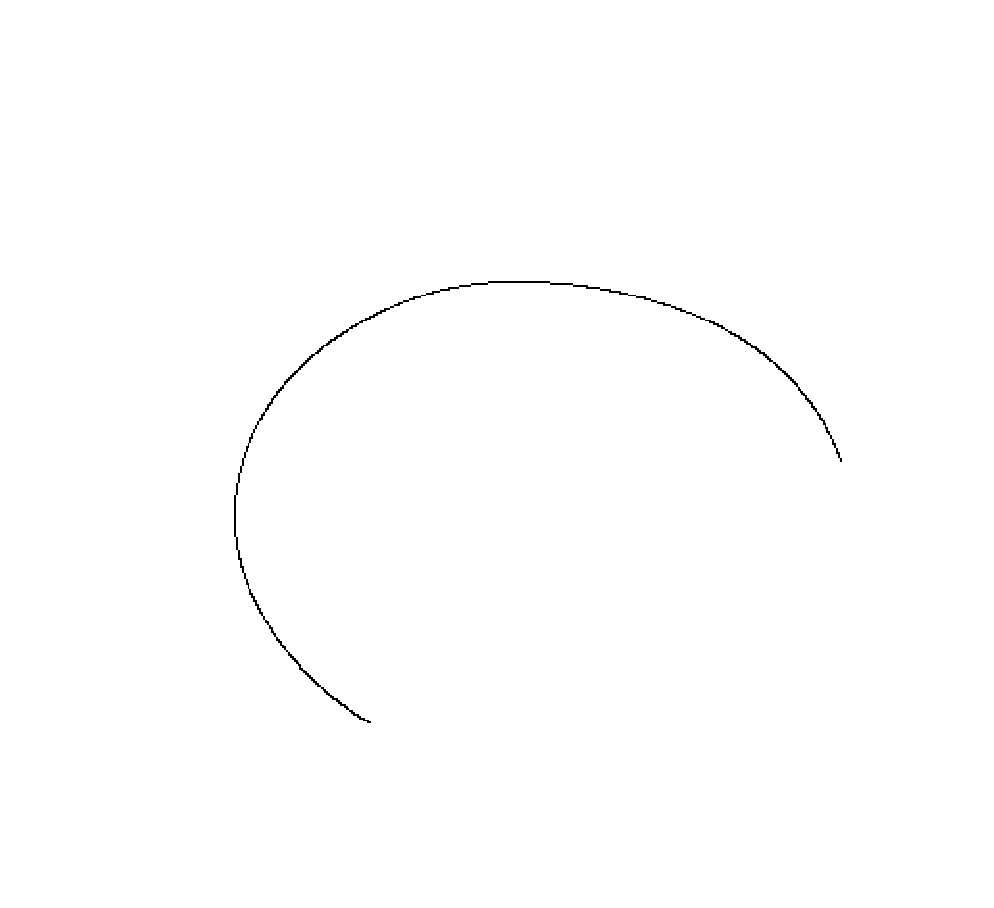}\\
(a)&(b)&(c)&(d)\\
 \end{array} \]
 \vspace{-0.5cm}
 \caption{ $(a)$ Original \emph{215} image from the South Florida database, $(b)$, $(c)$ and $(d)$ Canny ED outputs with parameters: high threshold $T_h=0.01,0.19,0.99$, lower threshold  $T_l=0.4 \times T_h$; standard deviation $\sigma=\sqrt2$, respectively.}
 \label{grafico_explicacion}
\end{figure}
An edge map is well balanced (reached \emph{Equilibrium}) if it is structurally simple. In this sense, the map in  panel $(d)$ of Figure~\ref{grafico_explicacion} is better balanced   than the edge map in panel $(c)$ and in turn, the one in $(c)$ is  better balanced  than the one in $(b)$.
Regarding \emph{Entropy}, one map has more information than another if the discontinuities, textures and shapes of the analyzed image are better  characterized. The overabundance of information produces chaotic (cluttered) edge maps like in $(b)$, but the absence of information produces poor edge maps like in $(d)$.  Thus, \emph{Equilibrium} and \emph{Information}  are two complementary concepts, and the \emph{Complexity} searches for a balance point between them.

Thus, to quantify the \emph{Equilibrium} of an edge map, we measured  local correlation against a family of specific edge patterns assessing the correct identification and value of the usual local characteristics of edges.  Since \emph{Entropy} should measure the amount of information of a system, which is maximized when the system reaches a random state,   we assess the randomness of an edge map with an index based on  contrasting the statistical distribution of the spatial edge positions against the bidimensional uniform distribution.

\subsection{Equilibrium index}

Abdou and Pratt introduced in their seminal paper the notion of Figure of Merit in order to score fragmented, offset and smeared edge patterns in comparison with the ideal edges present in the GT~\cite{Abdou1979}. The \emph{Equilibrium} index should perform a similar task in the unsupervised case, thus the GT will be replaced by a family $\mathfrak{B}$ of carefully chosen binary edge patterns.

Abusing notation, let  $\mathfrak{B}= \{\textbf{b}_{1},\textbf{b}_{2},\ldots,\textbf{b}_{n}\}$ be a collection of $N\times N$ edge patterns transformed into column vectors. Sliding a $N\times N$ window over the edge map $\textbf{b}$, centered in each edge pixel position $k$,  edge sub-maps  $\textbf{b}_{\left(k\right)}$ are extracted and transformed into column vectors. The CSM of each sub-maps with respect to the family of edge patterns $\mathfrak{B}$ is computed by
\begin{equation}
\label{qk}
\mathcal{Q}_{\mathfrak{B}}\left(\textbf{b}_{\left(k\right)}\right)=\max_{1\leq j\leq n}\frac{\textbf{b}_j^T\textbf{b}_{\left(k\right)}}{\|\textbf{b}_{j}\|\left\|\textbf{b}_{\left(k\right)}\right\|}\;.
\end{equation}

The \emph{Equilibrium}  of $\textbf{b}$ with respect to the family of edge patterns $\mathfrak{B}$ is defined as the average of the local CSM computed only on edge pixels $k$,
\begin{equation}\label{qbarra}
\mathcal{E}\left(\textbf{b}\right)=\frac{1}{|E_\textbf{b}|}\sum^{|E_\textbf{b}|}_{k=1}\mathcal{Q}_{\mathfrak{B}}\left(\textbf{b}_{\left(k\right)}\right),
\end{equation}
where $E_\textbf{b}$ is the set of all edge pixels in the binary edge map $\textbf{b}$, and $|E_\textbf{b}|$ is the cardinal number of such set.

\subsubsection{A family of edge patterns}

\begin{figure}[h]
\includegraphics[scale = 0.8]{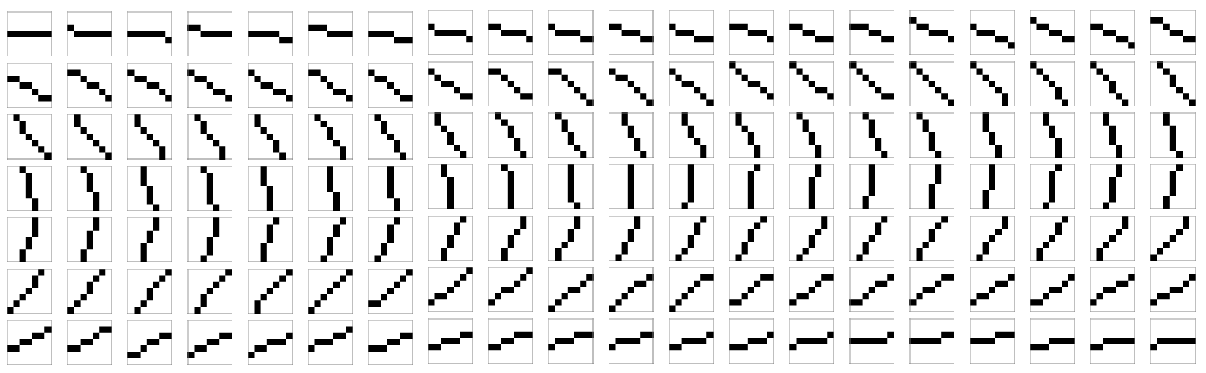}
\vspace{-0.5cm}
 \caption{A family of lineal edge $7\times 7$ pixel patterns obtained with Bresenham's algorithm.}
 \label{patrones7x7}
\end{figure}

The family $\mathfrak{B}$ of edge patterns could be very general, but in this paper, as in \cite{Kitchen1981}, only  line-like edge patterns are considered (Figure~\ref{patrones7x7}).
Being a line segment an essential primitive graphic,  it can be used to construct many other objects. Our line patterns are made with  an accurate and efficient raster line-generating algorithm defined in \cite{Bresenham1965}.  Bresenham stated that his line algorithms provide the best-fit approximations to the true lines by minimizing the error (distance) to the true primitive. Beginning with ray traces that go through the origin,  140 edge patterns of size $7\times7$ were constructed and stored in the present database (Figure~\ref{patrones7x7}).

\begin{figure}[h]
 \[\begin{array}{cc}
\begin{array}{c}
\includegraphics[scale = 0.25
]{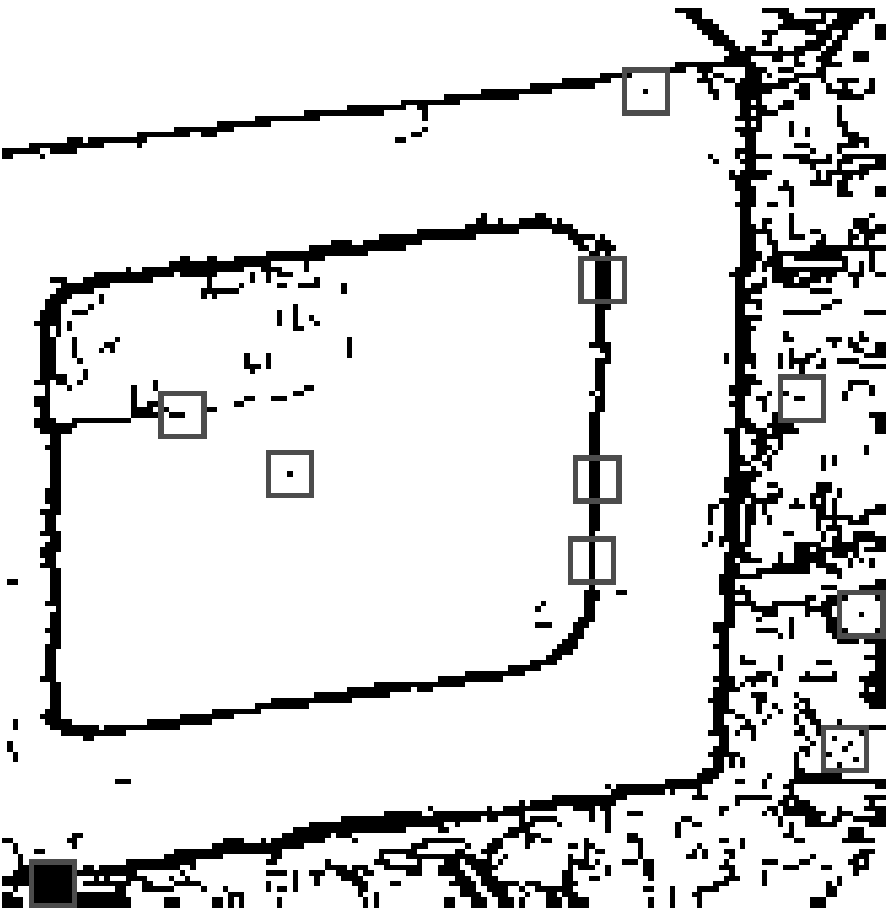}\\
(a)
 \end{array}&\begin{array}{ccccc}\includegraphics[scale = 0.1]{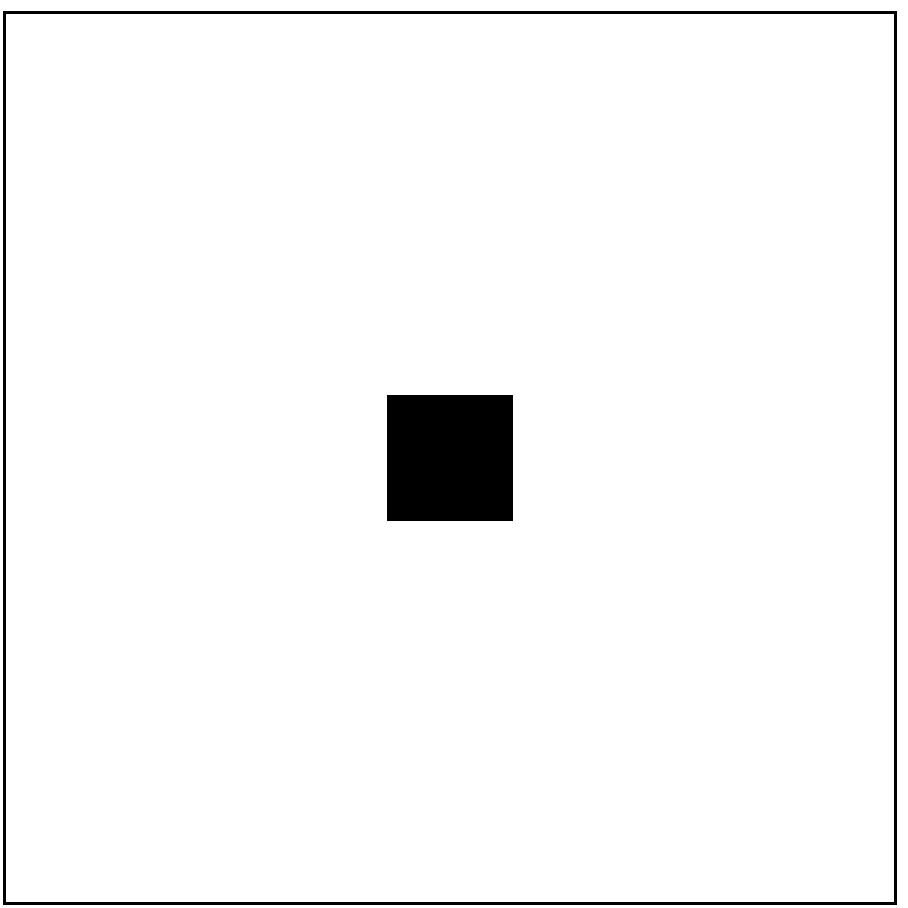}&\includegraphics[scale = 0.1]{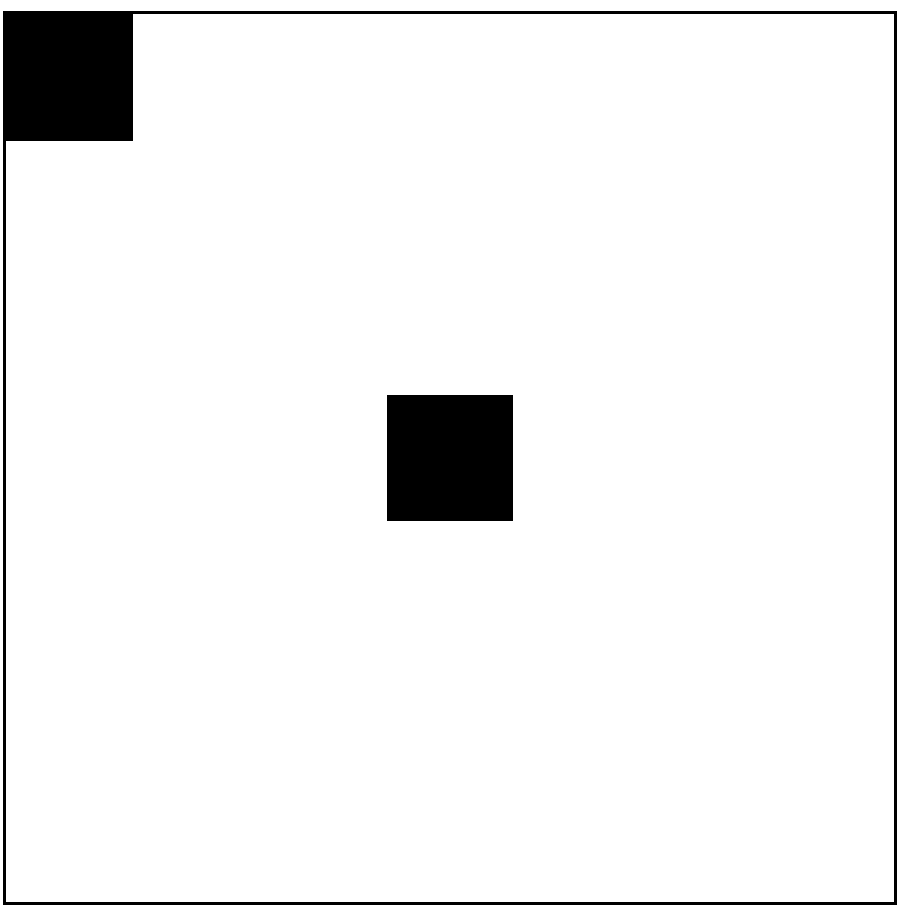}&\includegraphics[scale = 0.1]{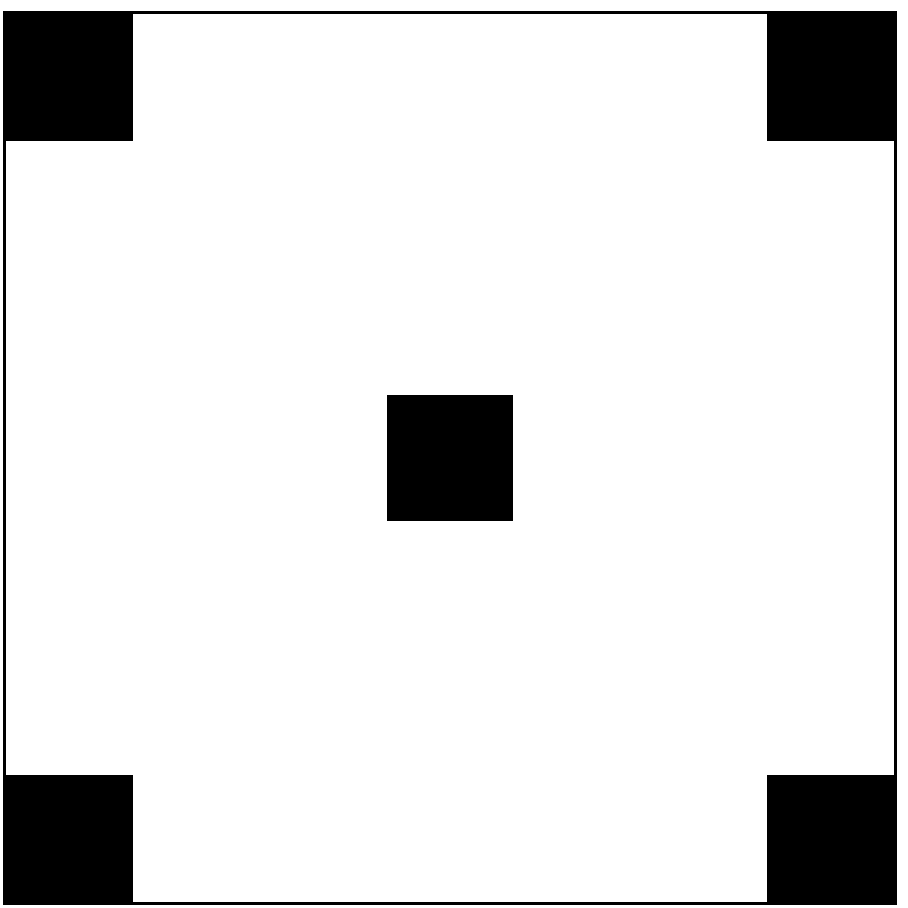}&\includegraphics[scale = 0.1]{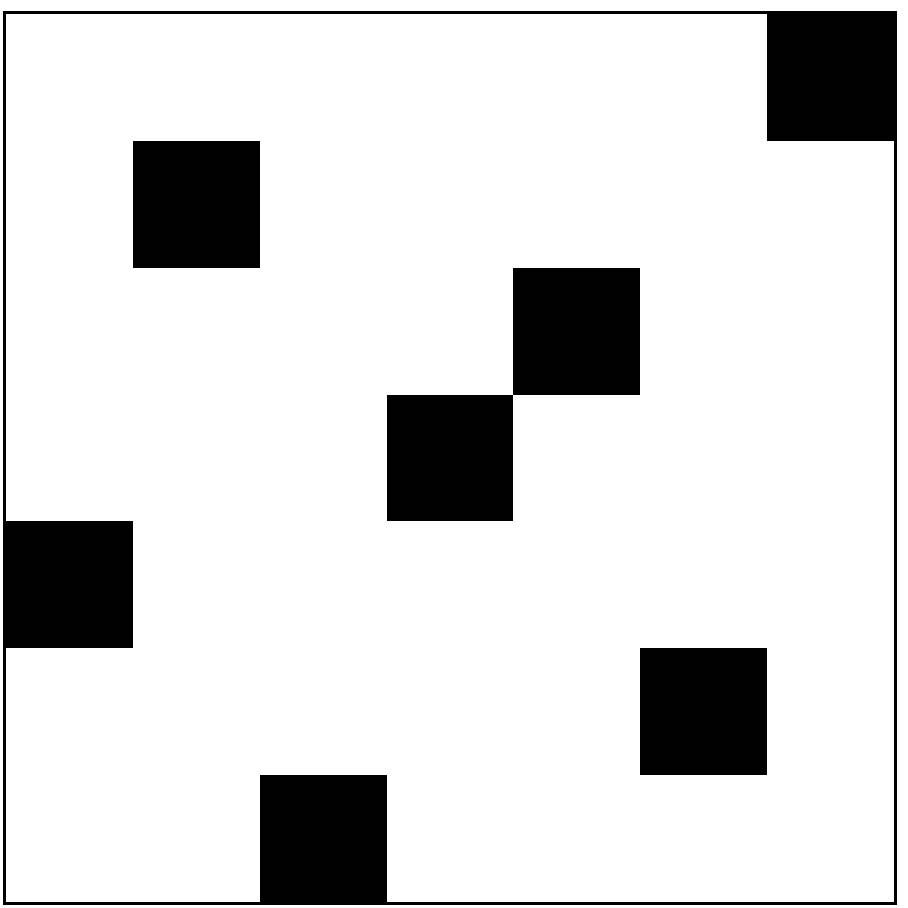}&\includegraphics[scale = 0.1]{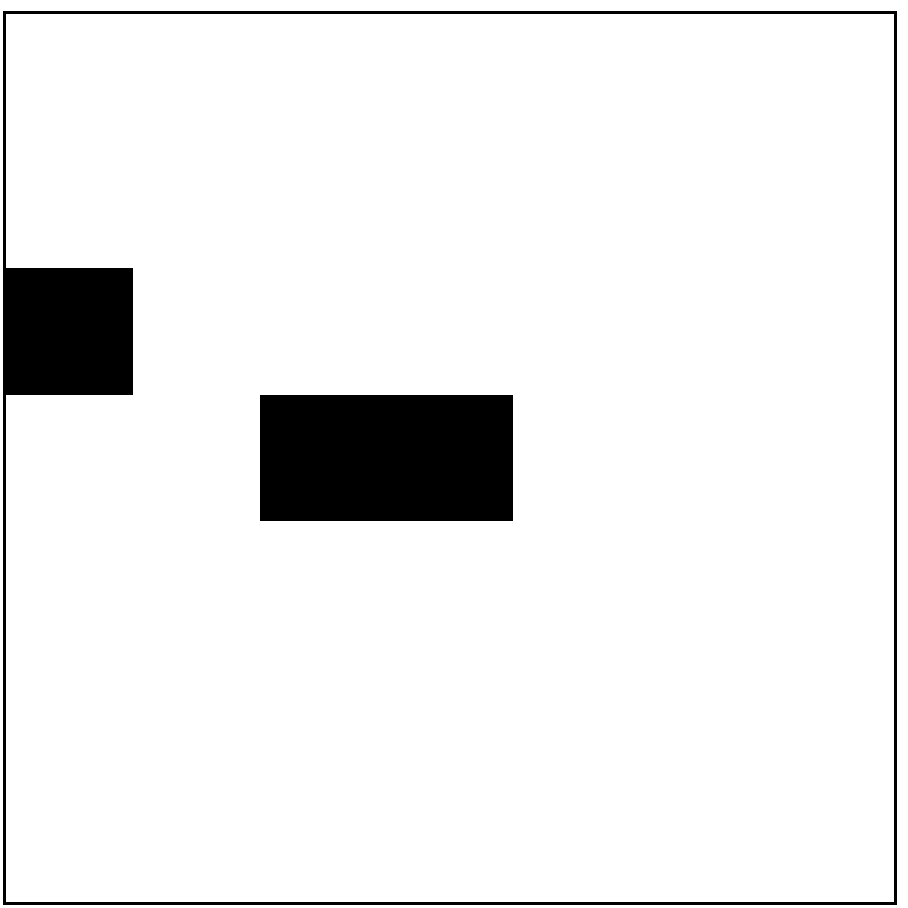}\\
(b)&(c)&(d)&(e)&(f)\\
\mathcal{Q}_{\mathfrak{B}}=0,378&\mathcal{Q}_{\mathfrak{B}}=0,534&\mathcal{Q}_{\mathfrak{B}}=0,507&\mathcal{Q}_{\mathfrak{B}}=0,428&\mathcal{Q}_{\mathfrak{B}}=0,654\\
\includegraphics[scale = 0.1]{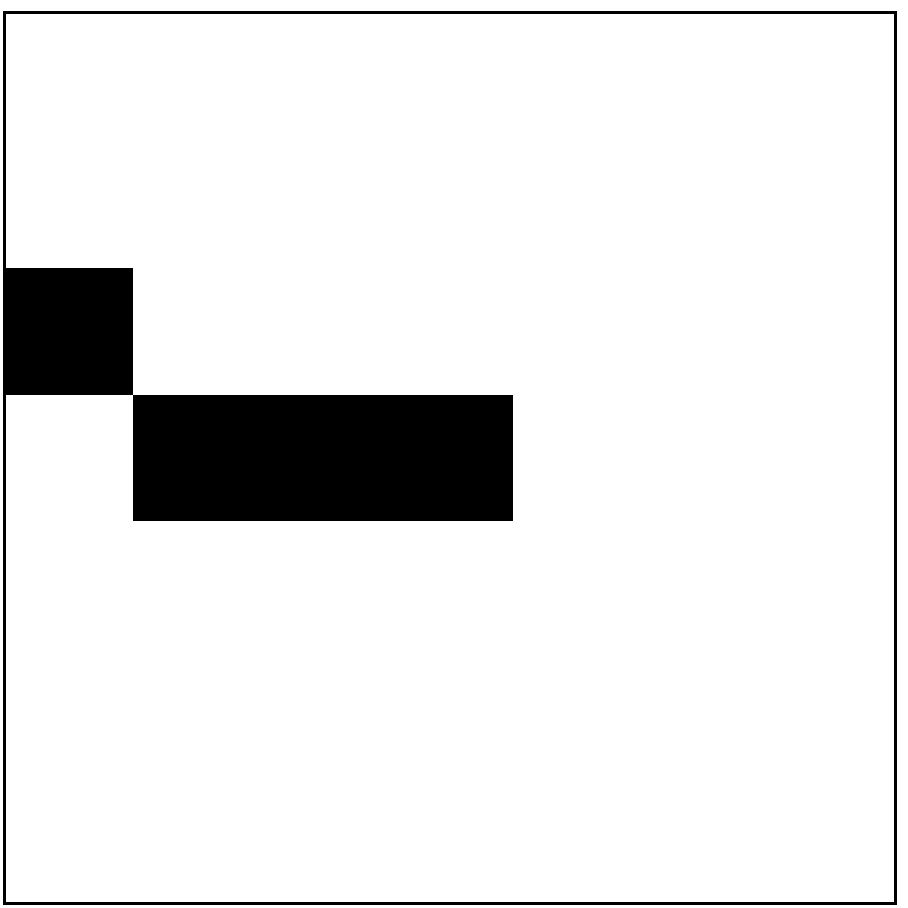}&\includegraphics[scale = 0.1]{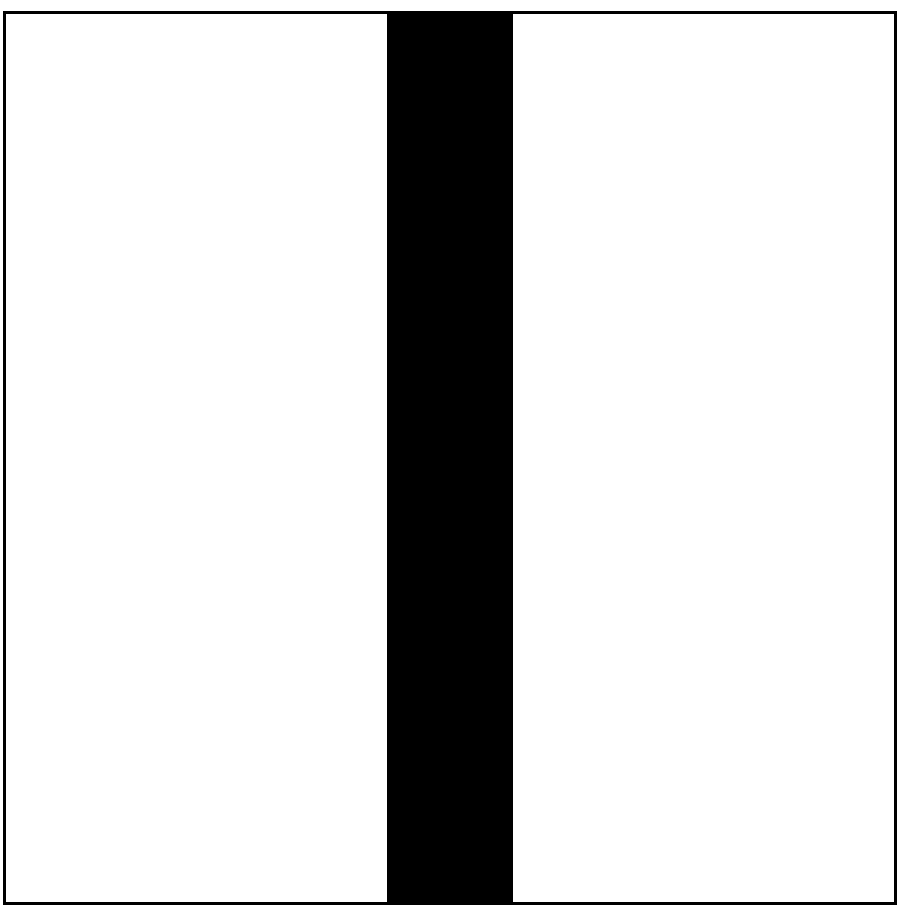}&\includegraphics[scale = 0.1]{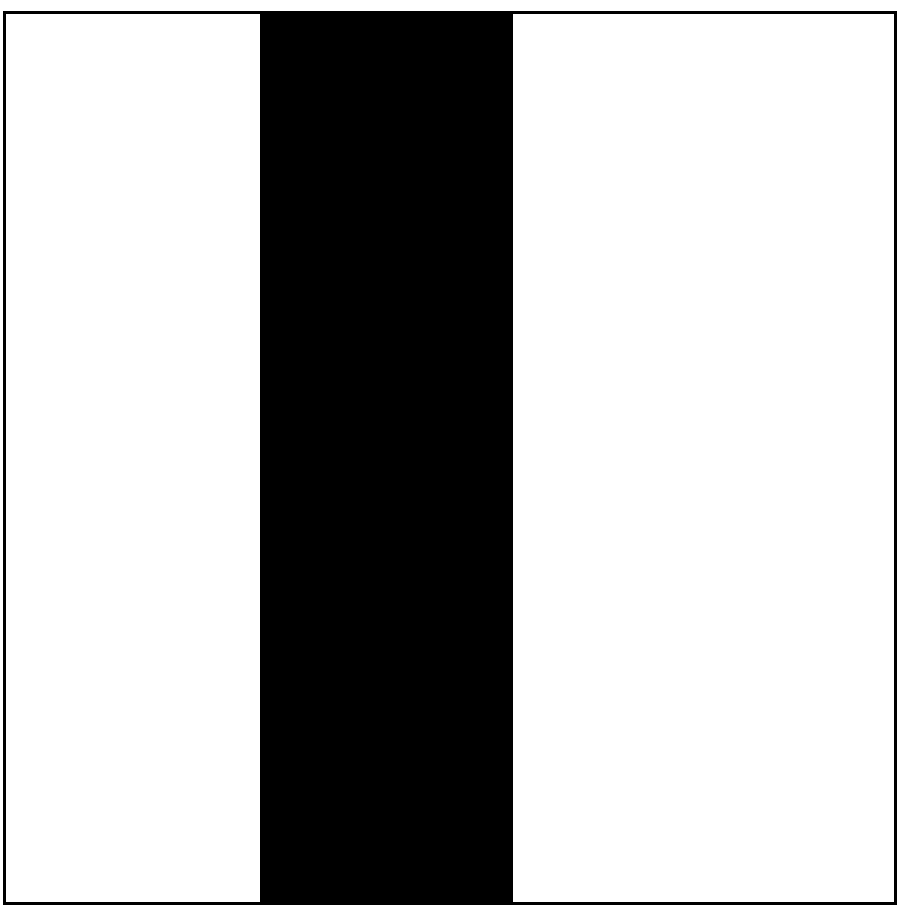}&\includegraphics[scale = 0.1]{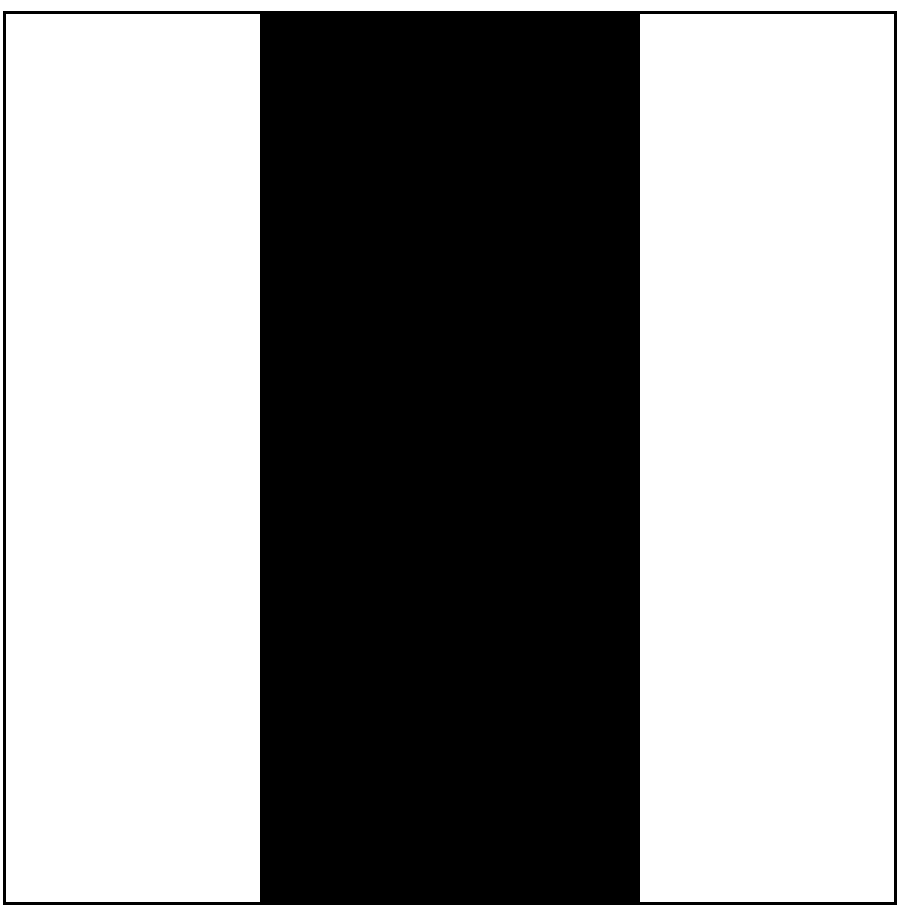}&\includegraphics[scale = 0.1]{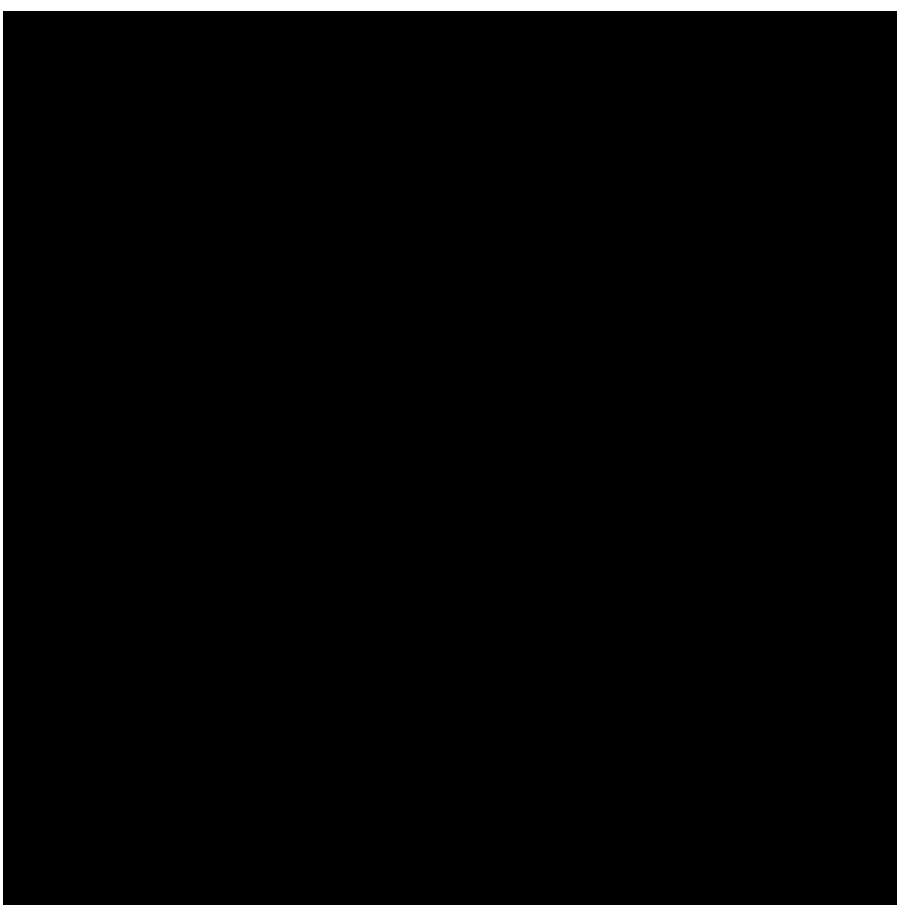}\\
(g)&(h)&(i)&(j)&(k)\\
\mathcal{Q}_{\mathfrak{B}}=0,755&\mathcal{Q}_{\mathfrak{B}}=1&\mathcal{Q}_{\mathfrak{B}}=0,707&\mathcal{Q}_{\mathfrak{B}}=0,577&\mathcal{Q}_{\mathfrak{B}}=0,378
\end{array}\end{array} \]
 \vspace{-0.5cm}
\caption{$(a)$ Edge map of image \emph{block}, $(b)$-$(k)$ windows of size $7\times7$ extracted from $(a)$.}
\label{fig:sample_subfigures}
\end{figure}

In  Figure~\ref{fig:sample_subfigures}, the values of $\mathcal{Q}_{\mathfrak{B}}$ on different patterns that appear in a Sobel edge map (computed from image \emph{block}) are shown. The edge pattern $(c)$, $(f)$, $(g)$ and $(h)$ show the performance of the index when the edges are close to line segments. The maps $(h)$-$(k)$ show the behavior of (Eq.~\ref{qbarra}) in presence of thick edges. The maximum value is reached in $(h)$, a pattern of a line of one pixel width. Noisy patterns $(b)$-$(e)$ reach an index value lower than 0.54.

\subsection{Information and the Kolmogorov Smirnov (KS) statistic}

Shannon stated that the information provided by an observation is proportional to how improbable it was \cite{shannon1948}. Relating this notion with the edge detection problem, having three points aligned in an edge map implies that the probability of having  a fourth point next to them is higher than the probability of having a point further away. Therefore, observing a point in a place with low probability gives more information than observing a point in an expected place. Randomness in space positions is at the core of the notion of \emph{edge information}.

Testing randomness in space  is a task usually done by testing the null hypothesis of uniform distribution in the unit square with the use the KS statistic. Such statistic takes values between  $0$ and $1$, rejecting the uniform hypothesis for values close to 1.

For a given edge map $\textbf{b}$, let $\phi=(\phi_1,\phi_2)$ be  an injective function that maps the edge positions $(i,j)\in E$ to the unit square $[0,1]\times [0,1]$, defined by $\phi(i,j) = (\frac{2i-1}{2N},\frac{2j-1}{2M})$ where $N \times M$ is the image size. Let $\mathcal{D}$ be the  KS bidimensional statistic defined by
\begin{equation}
\label{estadistico}
 \mathcal{D}(\textbf{b})=\displaystyle\max_{(x,y)\in\RR^2}|F_\textbf{b}(x,y)-F(x,y)|,
\end{equation}
where $F$ is the cumulative distribution function of an uniform distributed bidimensional vector, and $F_\textbf{b}$ the empirical distribution function of the sample  $\phi(E)$ given by
\begin{equation}
\label{empirica}
F_\textbf{b}(x,y)=\displaystyle\frac{|\{(i,j)\in E/\phi_1(i)\leq x,\phi_2(j)\leq y\}|}{|E|}\;.
\end{equation}

The  entropy measure $\mathcal{H}$ is defined as
\begin{equation}\label{entropia_mapa}
\mathcal{H}(\textbf{b})=1-\mathcal{D}(\textbf{b})\;.
\end{equation}
$\mathcal{D}(\textbf{b})$ was computed using the efficient algorithm by \cite{Justel1997}.
%

\section{Results and Analysis.}
\label{experimentos}

\subsection{Aim of the experiments}

 The computational experiments shown in this section are designed to investigate both the edge discrimination power of
 the concepts of \emph{Information} and \emph{Equilibrium} implemented by $\mathcal{H}$ and $\mathcal{E}$ and the use of $\mathcal{C}$ for ED performance characterization which includes: (i) the specific evaluation of an algorithm (intra-technique process) in order to identify its best parameters, and (ii) the comparison of different algorithms (inter-technique process) in order to classify them according to their quality.

  In order to do so, images from benchmark databases, compiled specifically for edge detection and object boundary detection  were selected, and for each image,  a database of edge maps was made by sampling the parameter space of several well known gradient based ED algorithms (EDA). On such database, $\mathcal{C}$ scoring is compared against reference-based measures, i.e. measures that take into account the GT provided by the image benchmark database.

 The reference-based measures considered are our $\mathcal{Q}_{\mathfrak{B}}$, defined in Section \ref{supervizado} and  the golden standard PFoM discrepancy measure given by
\[
 \mathcal{P}_\alpha(\textbf{g},\textbf{b})=\frac{1}{\max\{|E_{\textbf{b}}|,|E_{\textbf{g}}|\}}\sum_{k\in E_{\textbf{b}}} \frac{1}{1+\alpha d^2(k,E_{\textbf{g}})},
\]
where $\alpha=1/9$, $d$ is the Euclidean measure, $E_{\textbf{g}}$ and $E_{\textbf{b}}$ are the edge pixels subset of maps $\textbf{g}$ and $\textbf{b}$ respectively, \cite{Abdou1979, Lopez_Molina} .

 The performance of $\mathcal{C}$ in the intra technique evaluation process was done by studying the edge map selected for the maximum value of the scoring curves over the database and the actual scoring value. The former gives visual evidence and the later gives information about the balance between \emph{Equilibrium} and \emph{Entropy} (in the case of our unsupervised measure) and which of them is the closest to the GT in the case of PFoM and $\mathcal{Q}_{\mathfrak{B}}$.  The scoring curves $\mathcal{C}$, $\mathcal{E}$ and $\mathcal{H}$ also shed light on how each index reflects the degradations produced by excess or absence of edge pruning.

Several reference-based measures were compared in \cite{Lopez_Molina} by using a database of degraded images made by applying three different degradation operators to a single output of Canny EDA, with specific parameters selected to provide an overall good edge map related to the GT. The operators were addition of false positives, addition of false negatives, and diagonal displacements in a random fashion.

The experiments considered in this paper do not include random modifications in the map; all edges are true edges if they are considered in an appropriate scale. Also, all other EDA considered here, besides Canny, aim at the same output format as in  \cite{Yitzhaky2003, Fernandez2008}. They  are all gradient based EDA, thus they can be compared using quality curves \cite{Fernandez2004}.

Canny EDA was also used as a benchmark detector in  \cite{Martin2004}. The model presented as a baseline was Matlab's implementation of the Canny edge detector, with and without hysteresis. For both cases, standard deviation $\sigma$ was the only parameter to fit, since the thresholding parameters were considered parameters of the Precision-Recall (ROC) curve defined as reference-based evaluation methodology.

We follow the method of \cite{Fernandez2004} to describe the quality of an EDA that produces an edge map ${\textbf{b}(p)}$, being $p$ parameters in a one dimensional section $S$ of the EDA parametric space. Given the EDA, we produce an evaluation curve where each point on the curve is independently computed  by first setting the  parameters $p$ of the EDA to produce a binary map and then computing the evaluation measure on such map. When a single performance measure is required or is sufficient, the maximal value of the evaluation measure $\mathcal{M}$ \begin{equation}{\textbf{b}_S}=\arg \max_{p\in S} \mathcal{M}(\textbf{b}(p)),\end{equation}
is reported as a summary of the detector performance.

\subsection{Edge map database}

To construct the database, five gradient based EDA were considered: Canny \cite{Canny1986}, Prewitt \cite{Prewitt1970}, Sobel  \cite{Sobel1970} , Roberts  \cite{Roberts1963}  and Laplacian of Gaussian (LoG) \cite{Marr&Hildreth1980}, all provided by the  Matlab edge function from Matlab's Image Processing Toolbox.

According to Matlab's help, Canny EDA has three parameters: $\sigma$, Gaussian standard deviation of the derivative filter, and low and high hysteresis thresholding parameters. Edge thinning  by non maxima suppression was previously performed  to thresholding.
The LoG EDA has two parameters: $\sigma$ (standard deviation of the Gaussian Laplacian filter) and $T$ (thresholding parameter).
 Prewitt, Roberts and Sobel EDA are computed by  convolving the image with their corresponding gradient operators along the $x$ and $y$ direction. Thresholding is later applied to the gradient module. Images are not preprocessed with smoothing or denoising algorithms, since Matlab applies a private function to thinner edges and clean spurious points after thresholding.

For all images, a collection  of 100 edge maps were generated by moving each EDA  parameters as follows:
\begin{itemize}
\item Canny EDA: standard deviation is set at $\sigma=\sqrt2$, high threshold hysteresis parameter ($T_h$) is (equally spaced) sampled 100 times from zero to one, and  the low threshold parameter  ($T_l$) is set as $T_l=0.4\times T_h$.
\item  Sobel, Prewitt and  Roberts EDA:  threshold parameter ($T$) is sampled 100 times from 0.004 to 0.396. Edge thinning is turned on.

\item LoG EDA: standard deviation is set at $\sigma=\sqrt2$ and the threshold parameter ($T$) is sampled 100 times from 0.0004 to 0.0396.

\end{itemize}
The final database has 500 edge maps for each real image considered. Moving each EDA threshold from the smallest to the largest  value generally produces edge maps with a varying number of featured points. The maximum value of $\mathcal{C}$ over such EDA outputs selects the edge map with optimal balance between \emph{Equilibrium} and \emph{Information}.
 \begin{figure}[h]
 \[\begin{array}{cccc}
 \includegraphics[scale = 0.18]{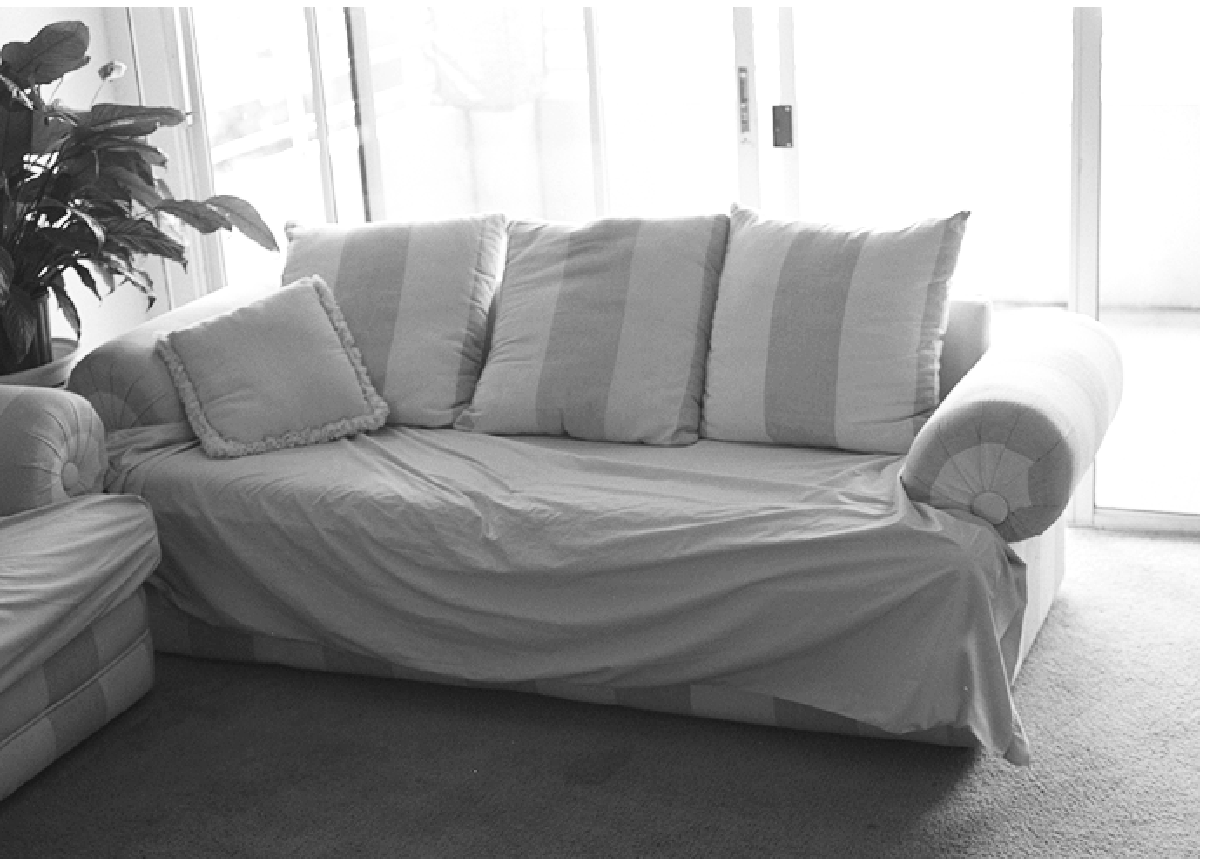}&\includegraphics[scale = 0.18]{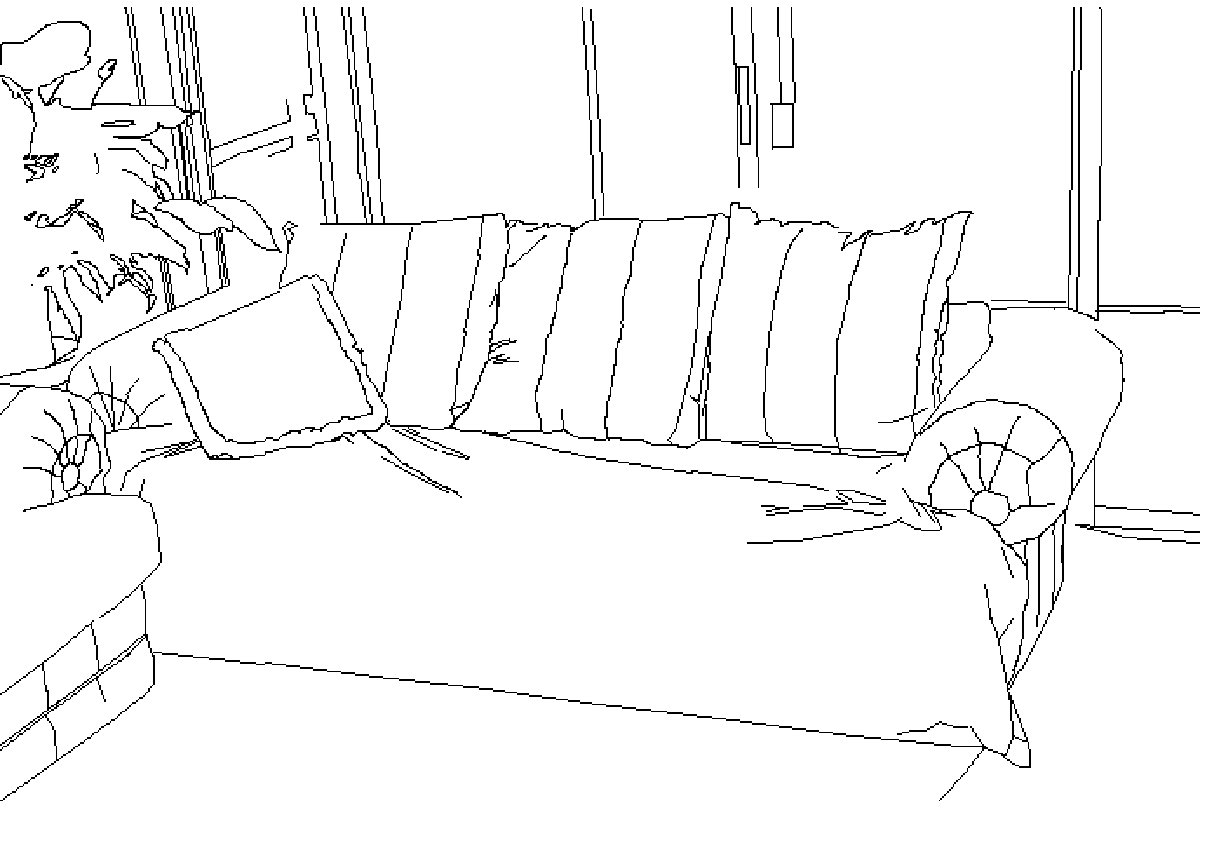}&\includegraphics[scale = 0.18]{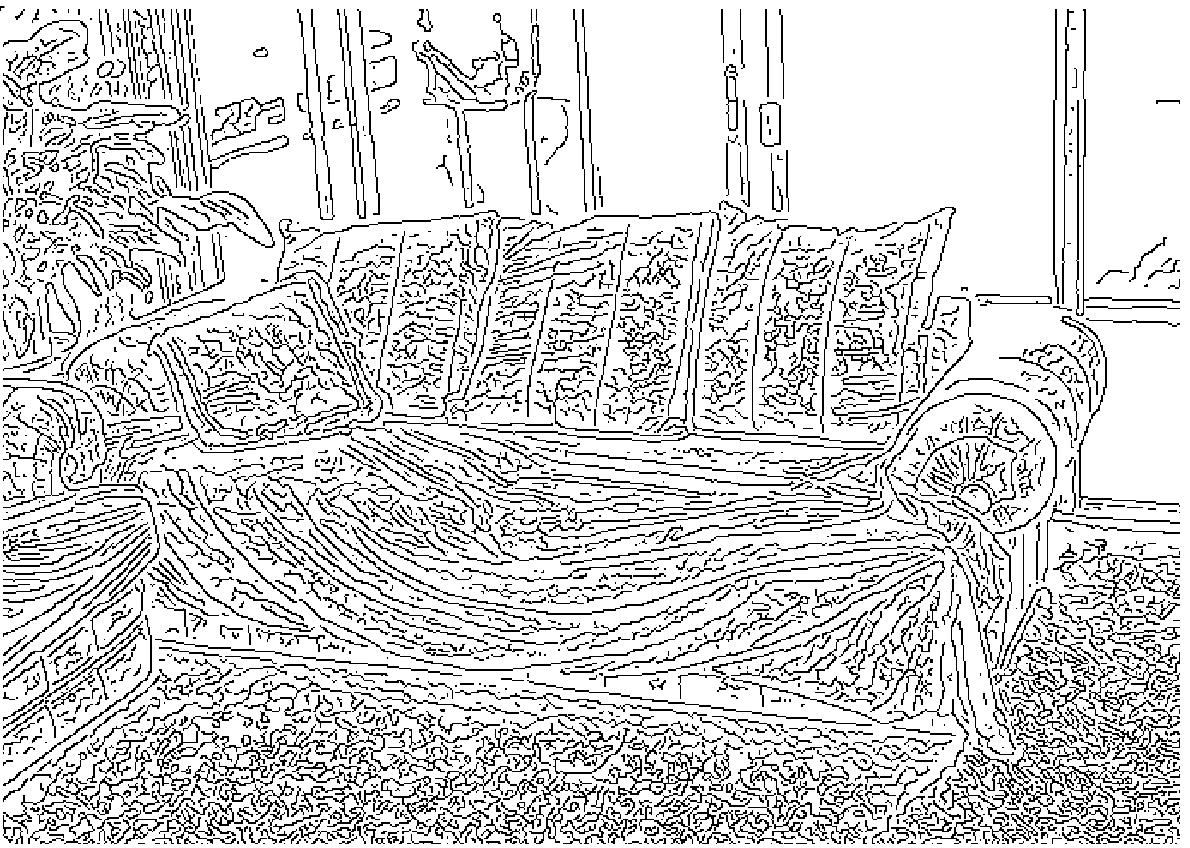}& \includegraphics[scale = 0.18]{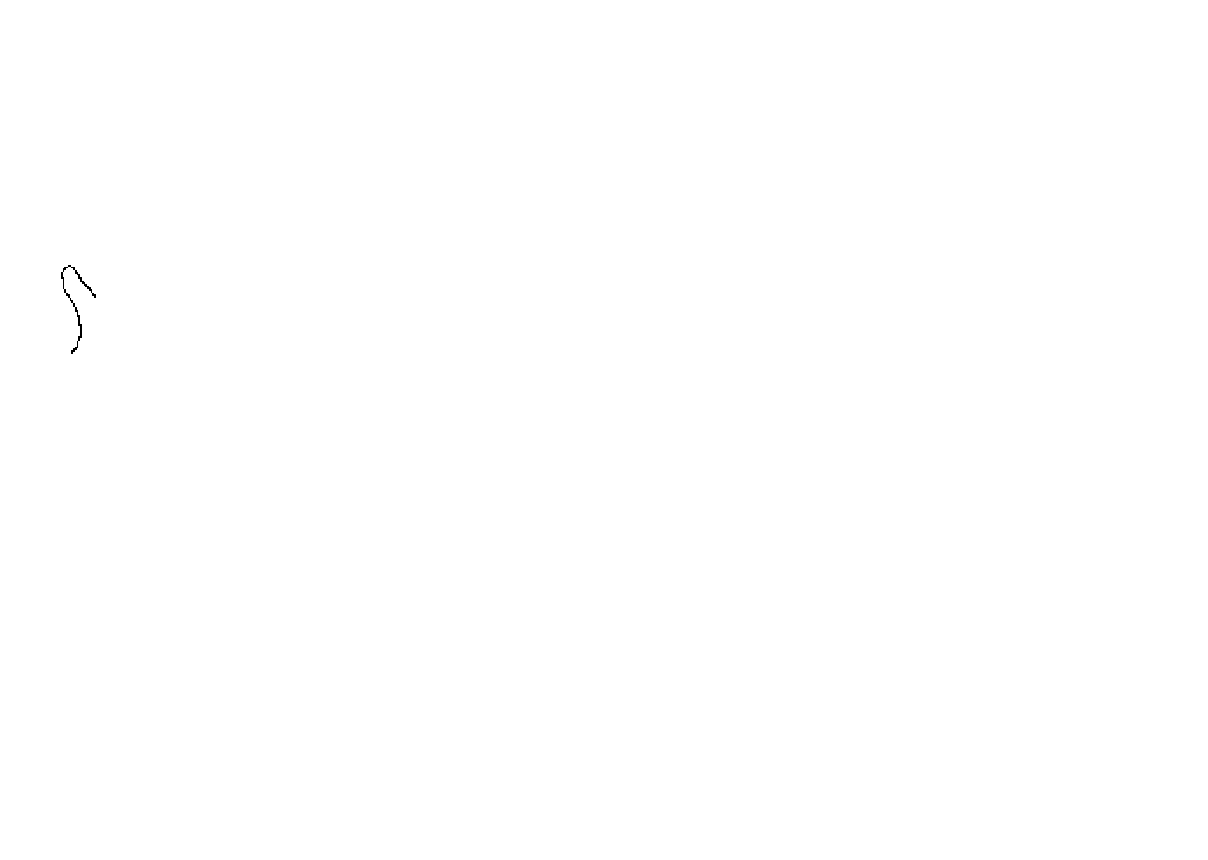}\\
(a)&(b)&(c)&(d)\\
 \end{array} \]
 \vspace{-0.5cm}
 \caption{Comparison among GT and extreme edge maps $(a)$ Original image \emph{109}, $(b)$ GT,  $(c)$-$(d)$ Canny's extreme edge maps with $T_h=0.01, 0.99$; $T_l=0.4\times T_h$; and $\sigma=\sqrt2$. }
 \label{grafico_CE_i109}
\end{figure}

\subsection{First experiment: South Florida database}
The South Florida Database is a public database provided by  \cite{Bowyer1999} specifically for edge detection assessment.  It largely consists of indoor images with little background texture. It includes two collections: one of 50 natural images  and other with $10$ aerial images. Each of the fifty images of the first collection contains a single object approximately centered in the image and set against a natural background. The second collection has images of man made constructions.

For each of the 60 images, we constructed a database of $500$ edge maps and all measures   $\mathcal{E}$, $\mathcal{H}$, $\mathcal{C}$, $\mathcal{Q}_{\mathfrak{B}}$ and PFoM were computed on them, the latest two by using the GT available in the database.

\begin{figure}[h!]
 \centering
 \[\begin{array}{cccc}
\includegraphics[scale = 0.2]{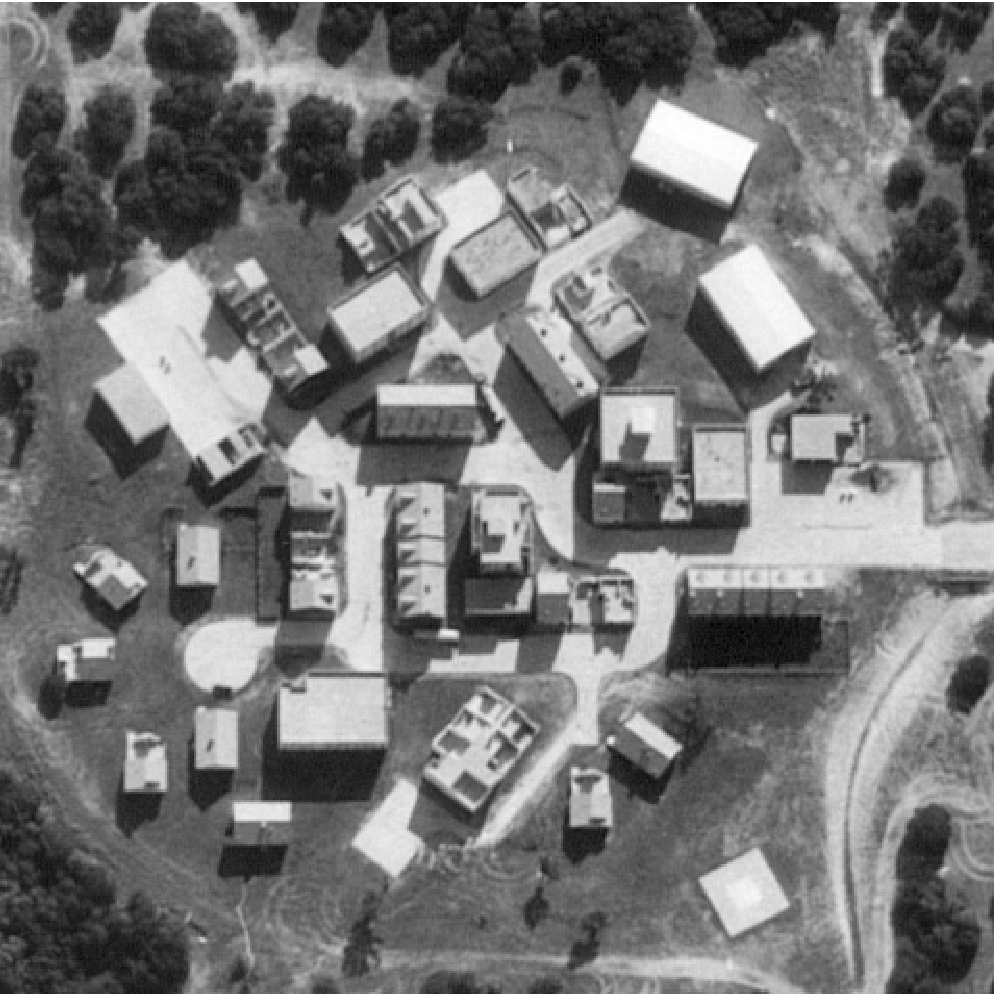}&\includegraphics[scale = 0.2]{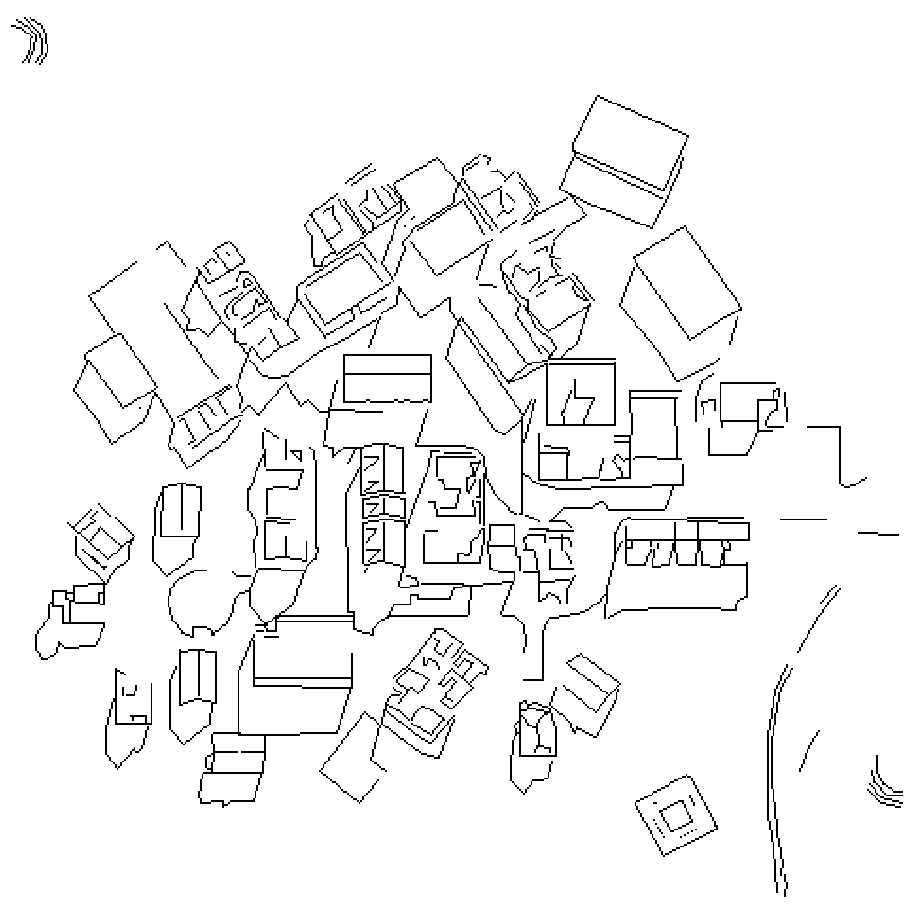}&&\\
\includegraphics[scale = 0.2]{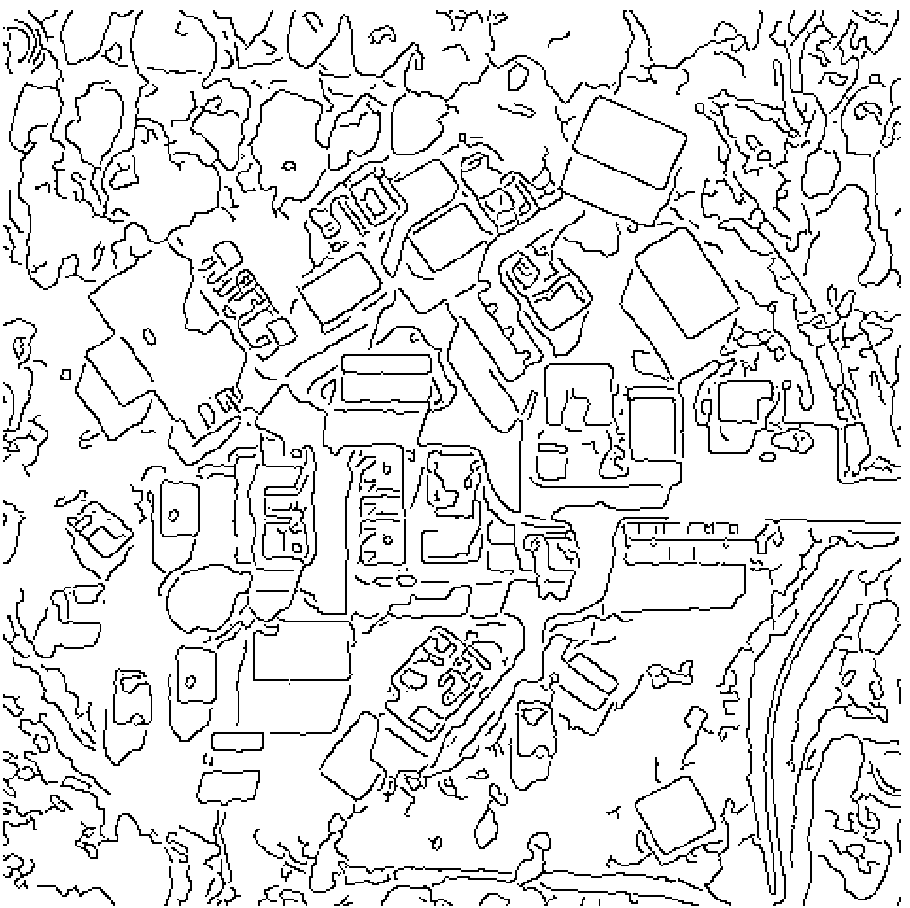}&\includegraphics[scale = 0.2]{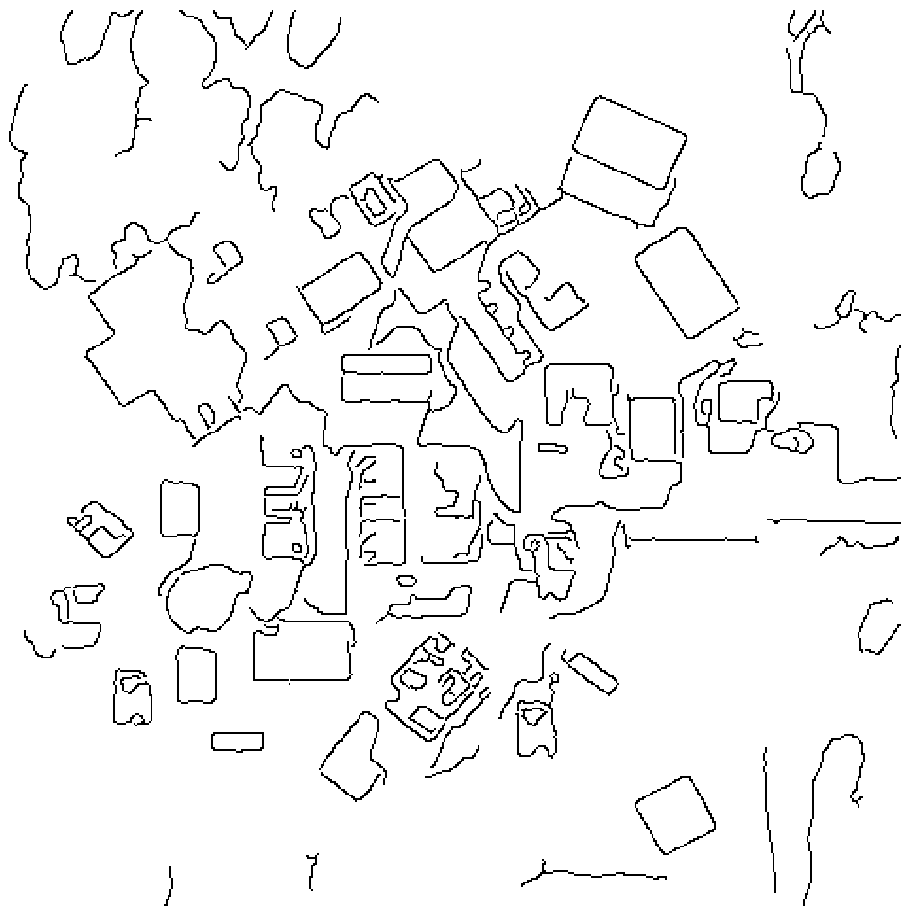}&\includegraphics[scale = 0.2]{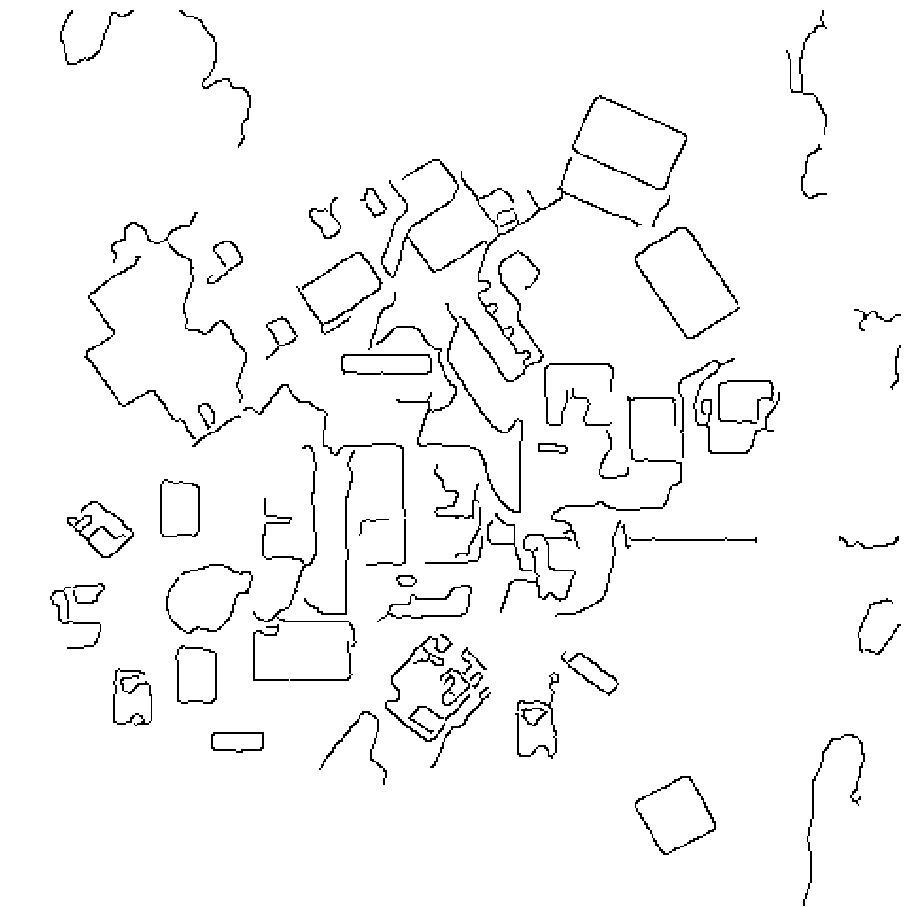}&\\
\includegraphics[scale = 0.2]{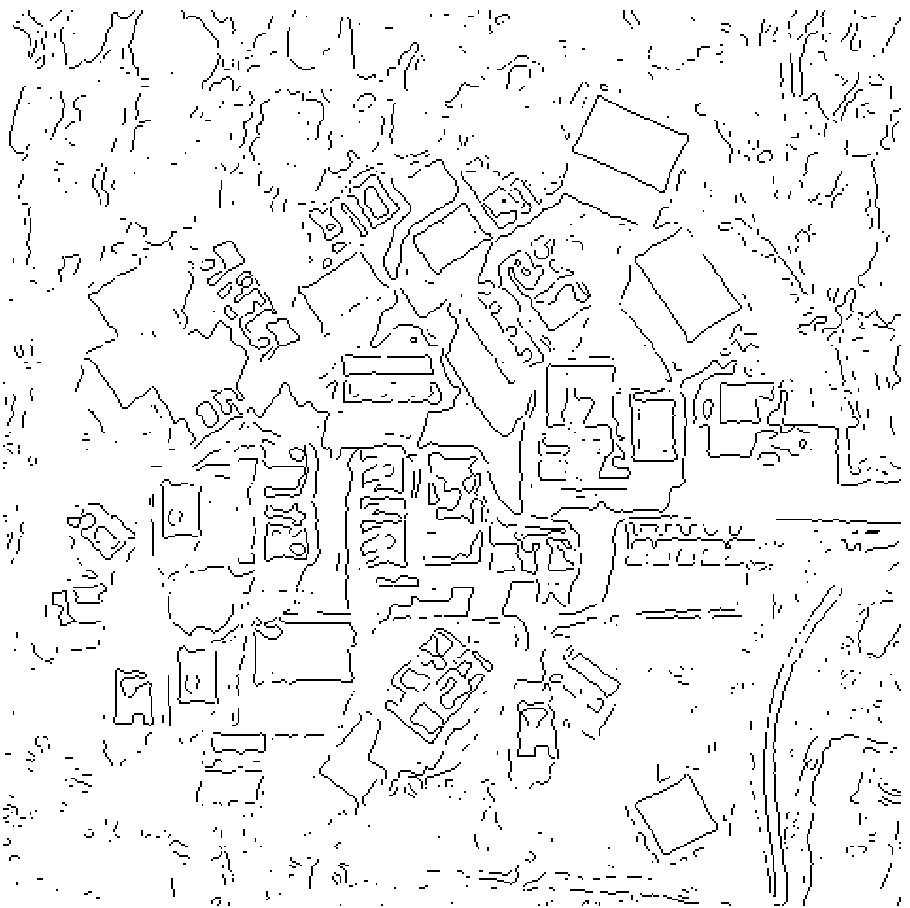}&\includegraphics[scale = 0.2]{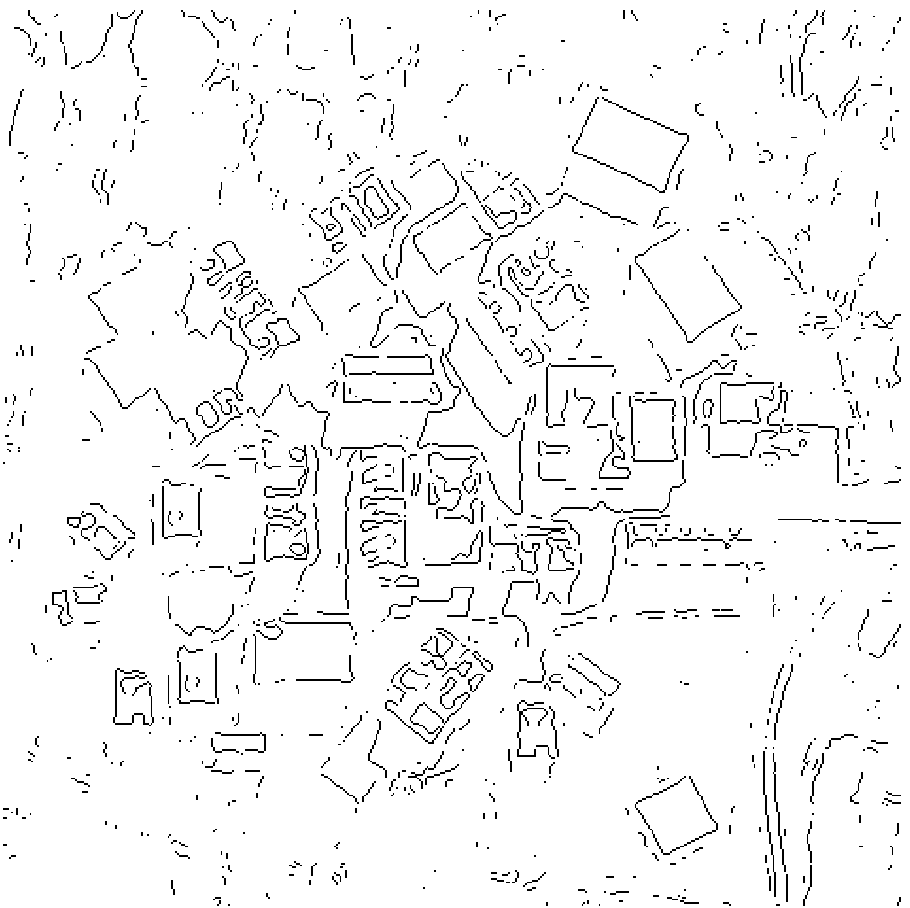}&\includegraphics[scale = 0.2]{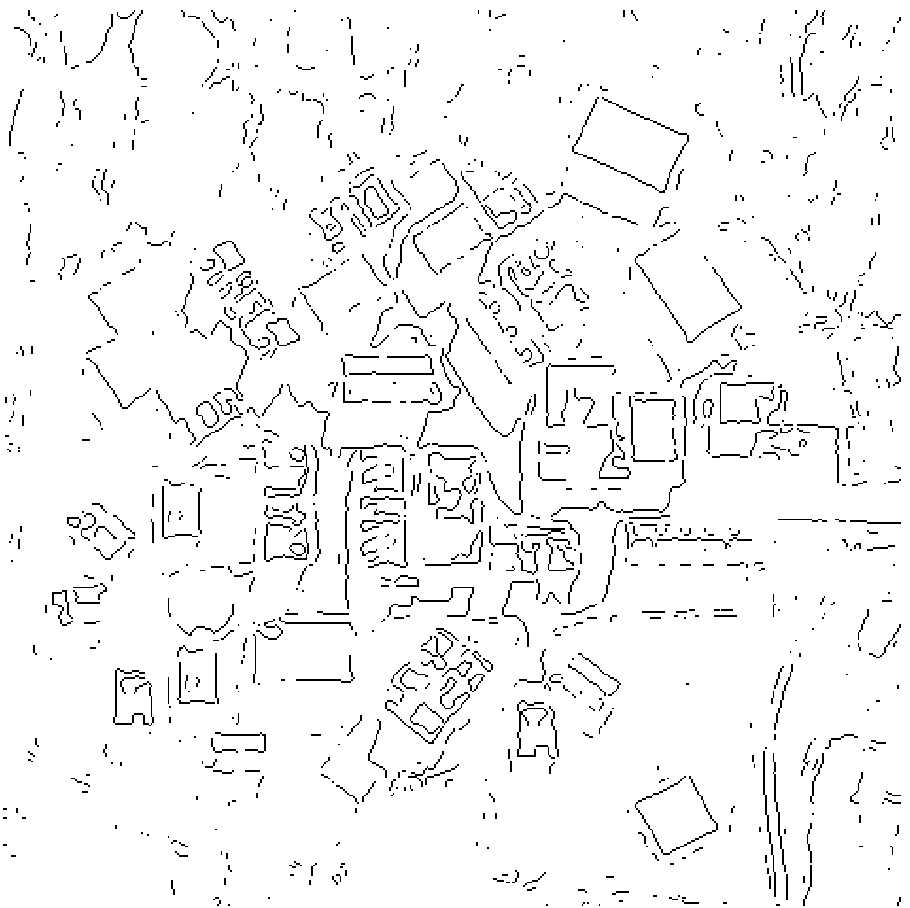}&\\
\includegraphics[scale = 0.2]{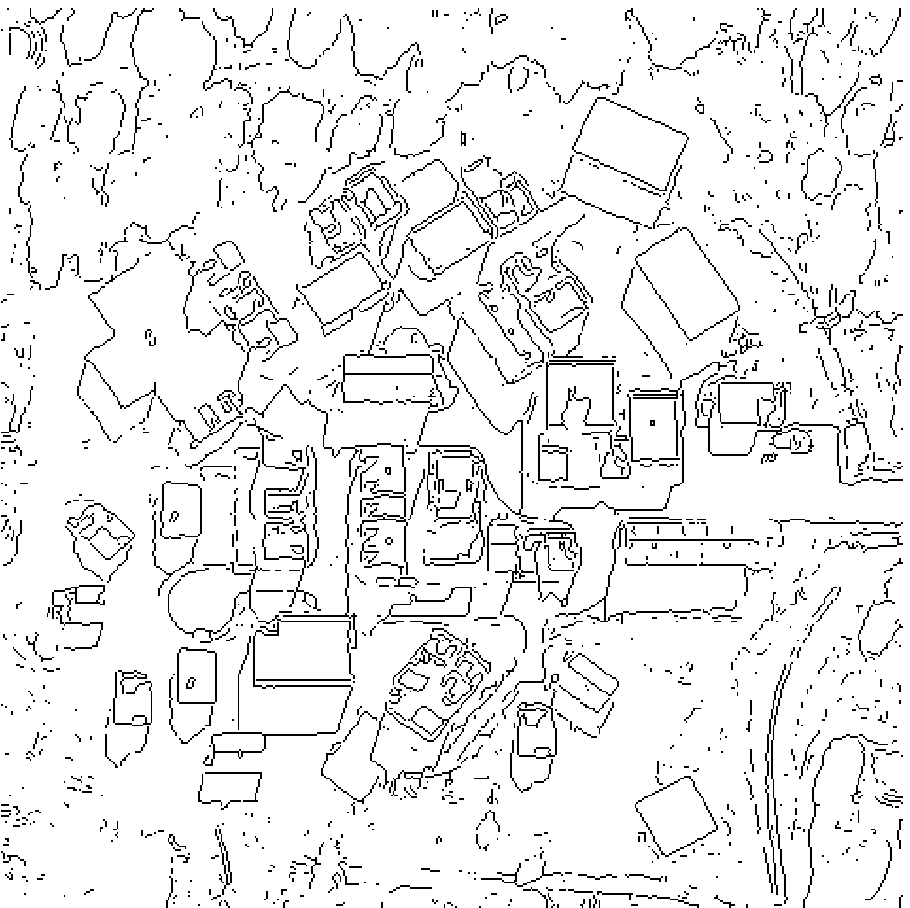}&\includegraphics[scale = 0.2]{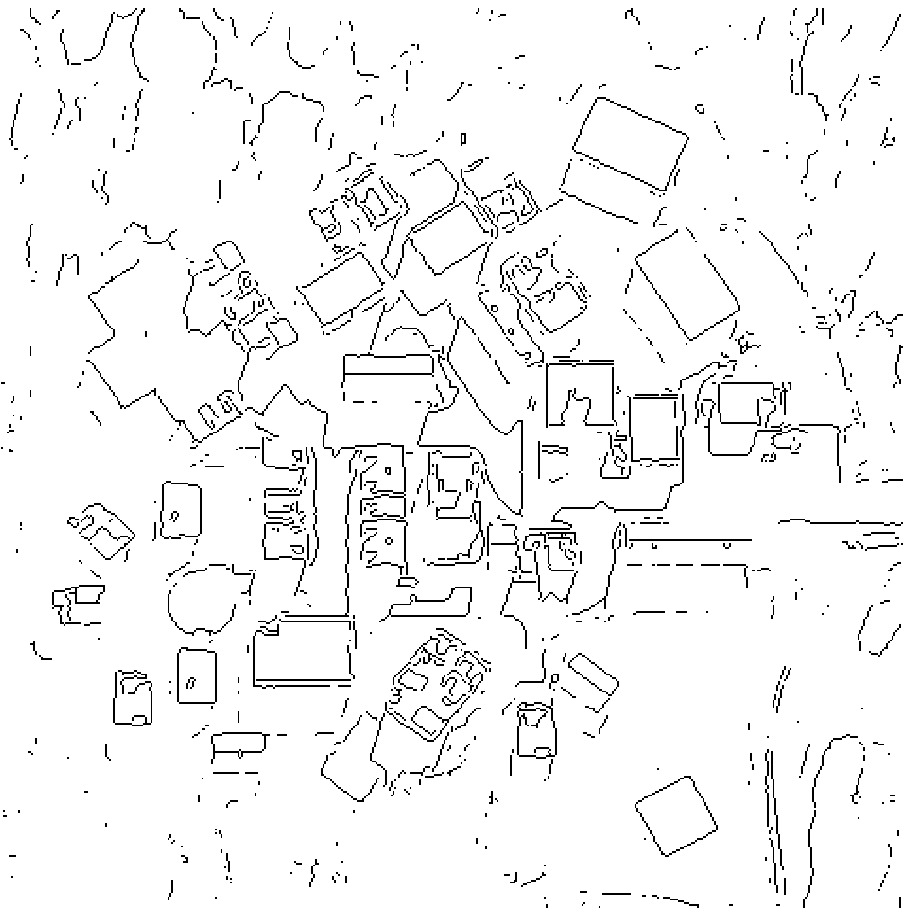}&\includegraphics[scale = 0.2]{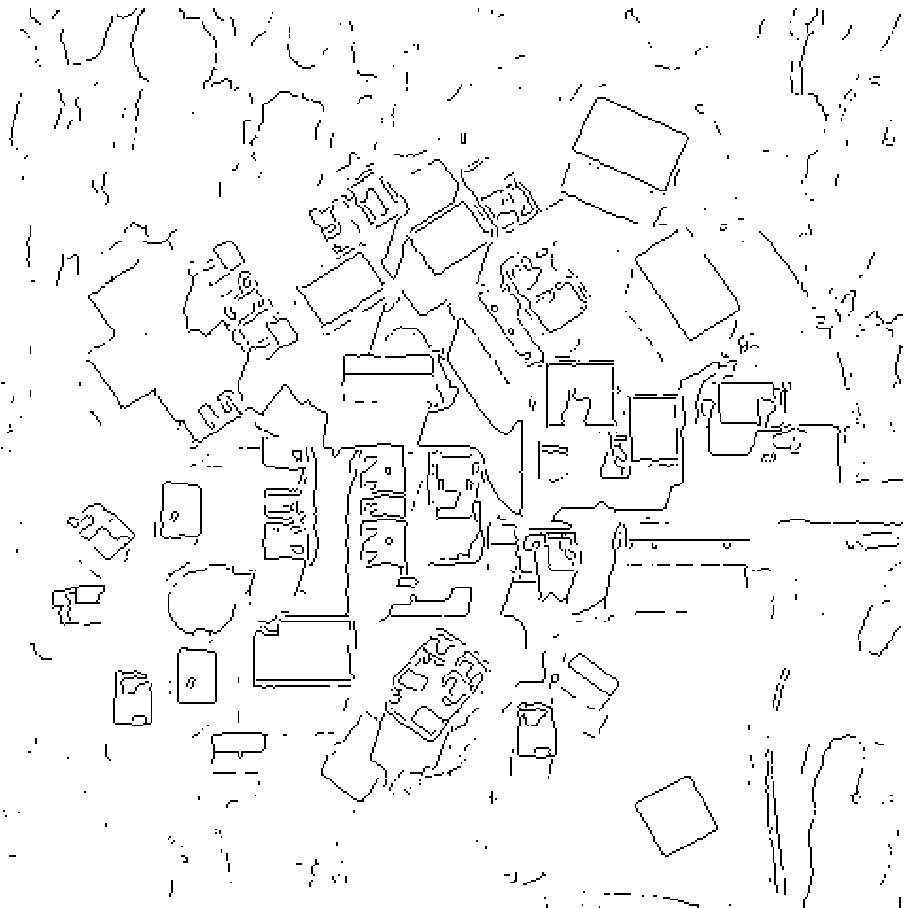}&\\
\includegraphics[scale = 0.2]{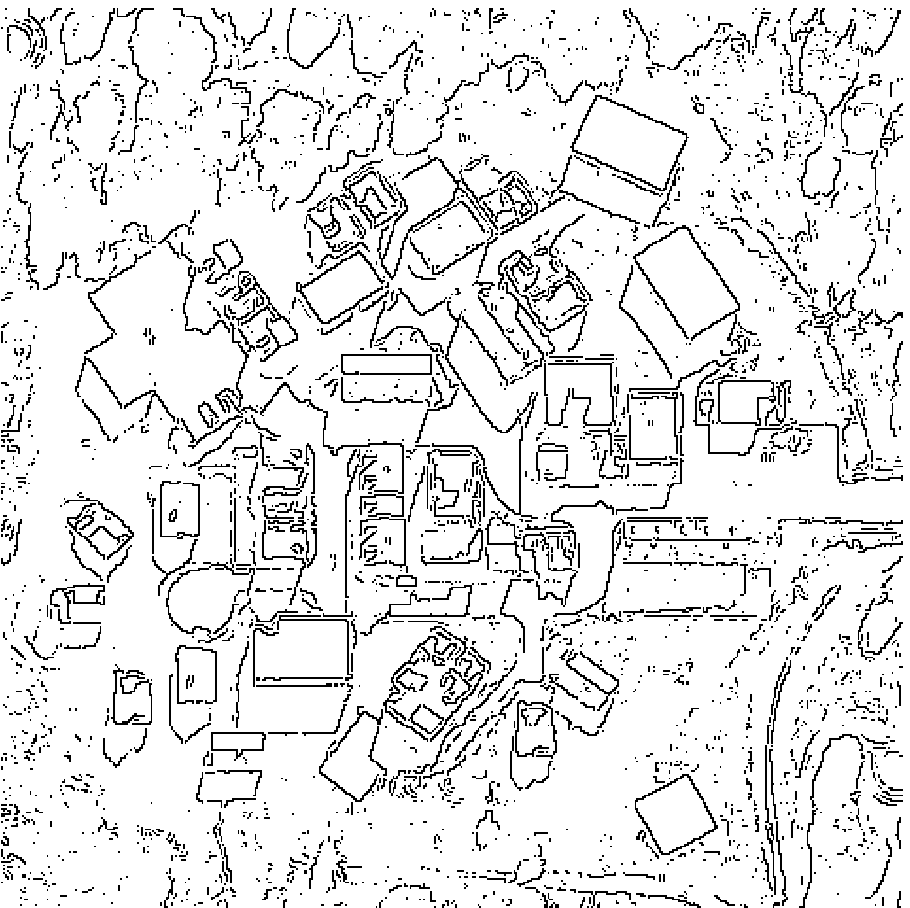}&\includegraphics[scale = 0.2]{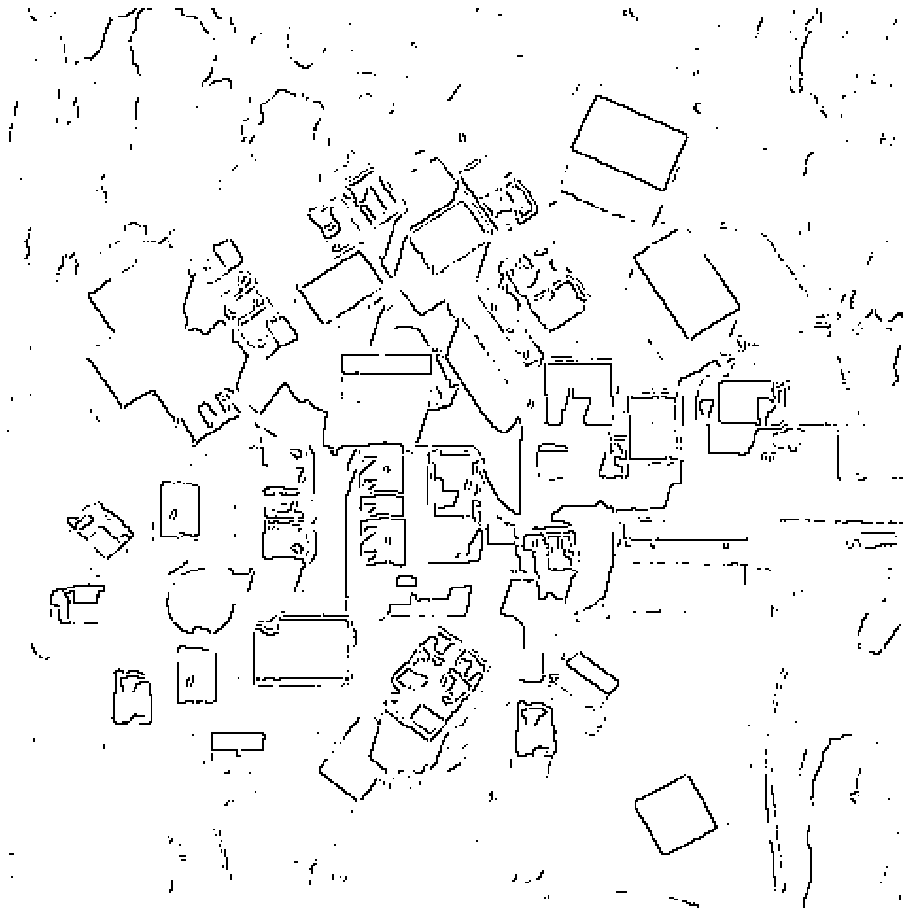}&\includegraphics[scale = 0.2]{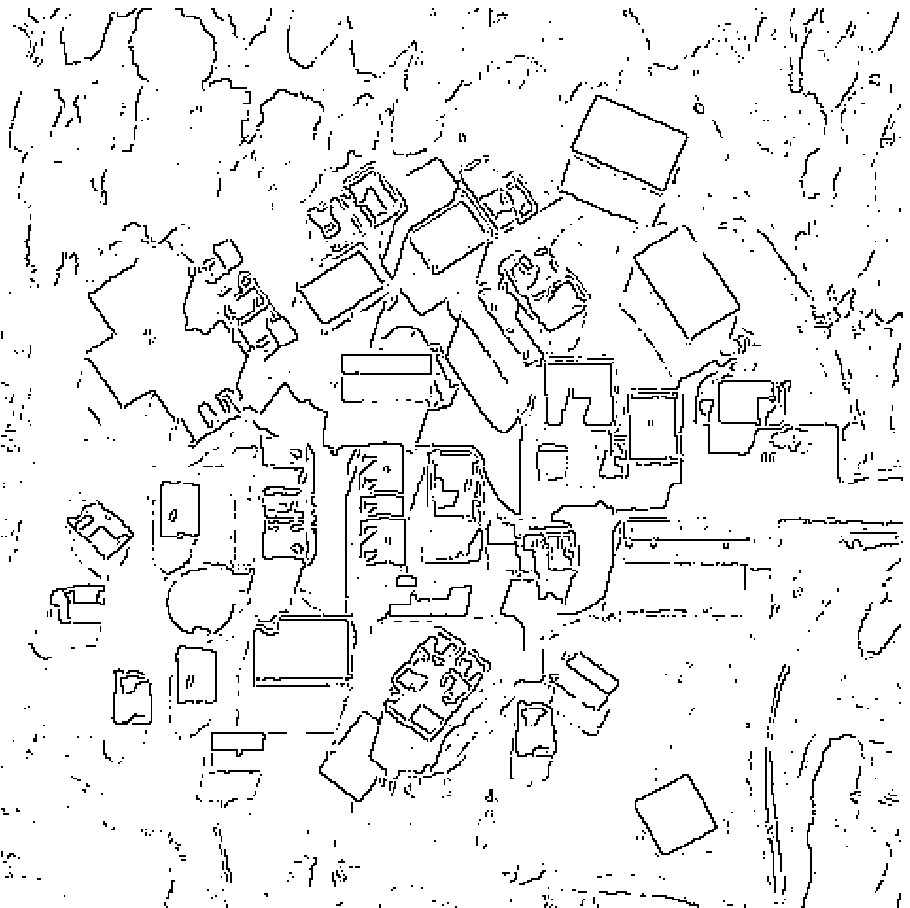}&\\
\includegraphics[scale = 0.2]{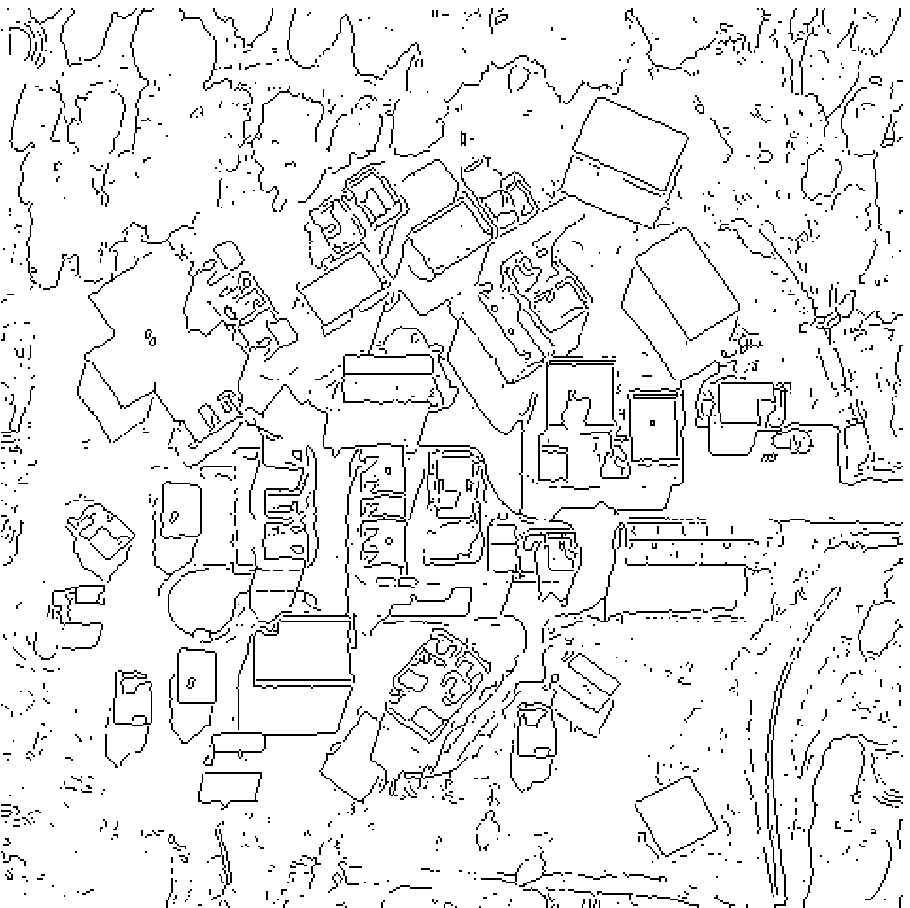}&\includegraphics[scale = 0.2]{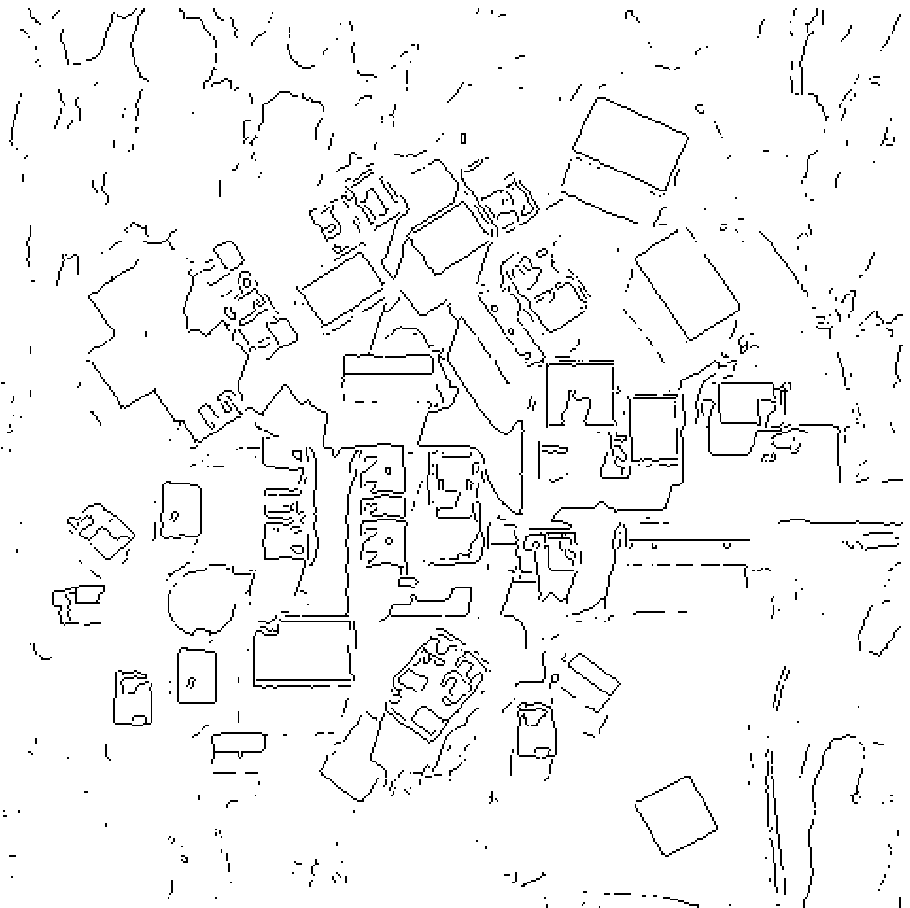}&\includegraphics[scale = 0.2]{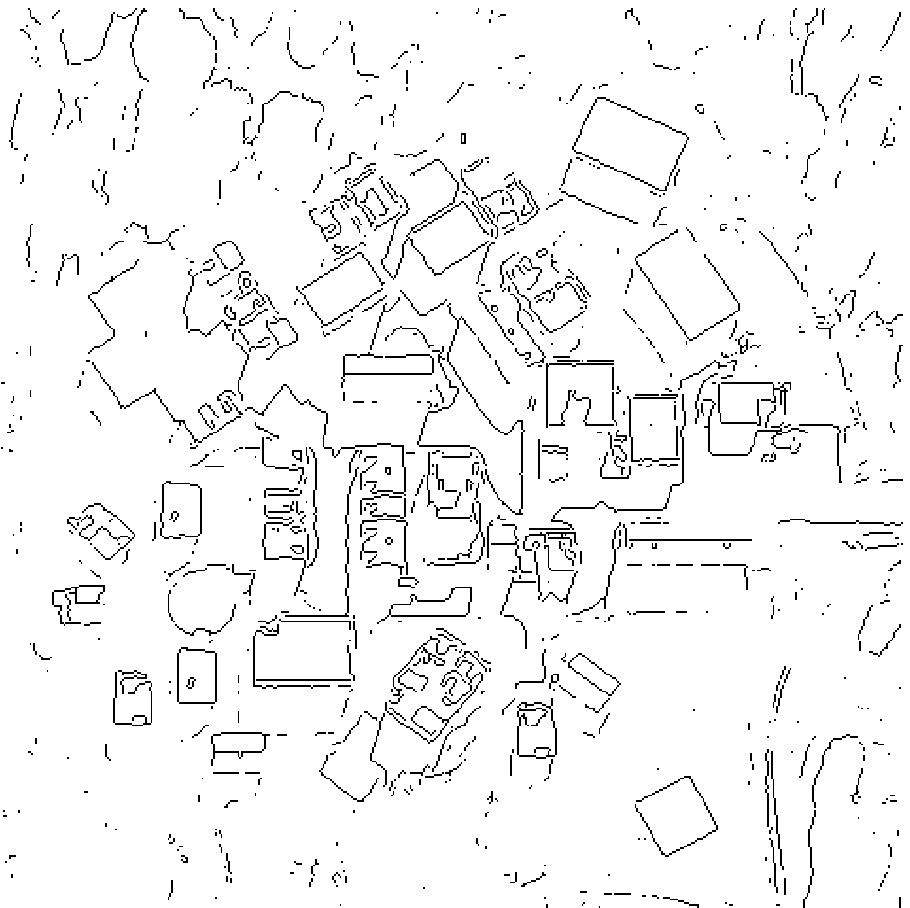}&\\
\end{array} \]
 \vspace{-0.5cm}
 \caption{ Best EDA map selection according to $\mathcal{C}$, PFoM and $\mathcal{Q}_{\mathfrak{B}}$, from a collection of edge maps of image \emph{woods}, made with different EDA. First row: Original image \emph{woods} and GT. From the second to the last row, each column corresponds to best edge map according to $\mathcal{C}$, PFoM and $\mathcal{Q}_{\mathfrak{B}}$ obtained  by different EDA:  Canny; LoG; Prewitt; Roberts and Sobel, respectively.}
 \label{woods2}
\end{figure}

We selected two images for a qualitative performance discussion: image \emph{109}, good quality grayscale indoor image with a central object (Figure~\ref{grafico_CE_i109} $(a)$) and image \emph{woods}, a good quality outdoors aerial image which depicts several buildings surrounded by woods and  country roads (Figure~\ref{woods2}). The first image has little texture while the second  contains a lot of vegetation (e.g., grass, shrubs, trees), which corresponds to texture in the image. The GT depicts edges only related  to object boundary, which do not include the trees present in the image. Overall, this image is a challenge for edge detectors, in particular, for those which only use grayscale information.

In Figure~\ref{grafico_CE_i109}, image \emph{109} and its GT are shown along with two extreme edge maps computed with Canny EDA,  with $T_h=0.01$ and $T_h=0.99$. The first edge map has  many texture details transformed in short edges,  and the second has almost no edges. The other $98$ edge maps are comprised in between these two extreme edge maps.
\begin{figure}[h!]
\[\begin{array}{cc}
 \includegraphics[scale = 0.3]{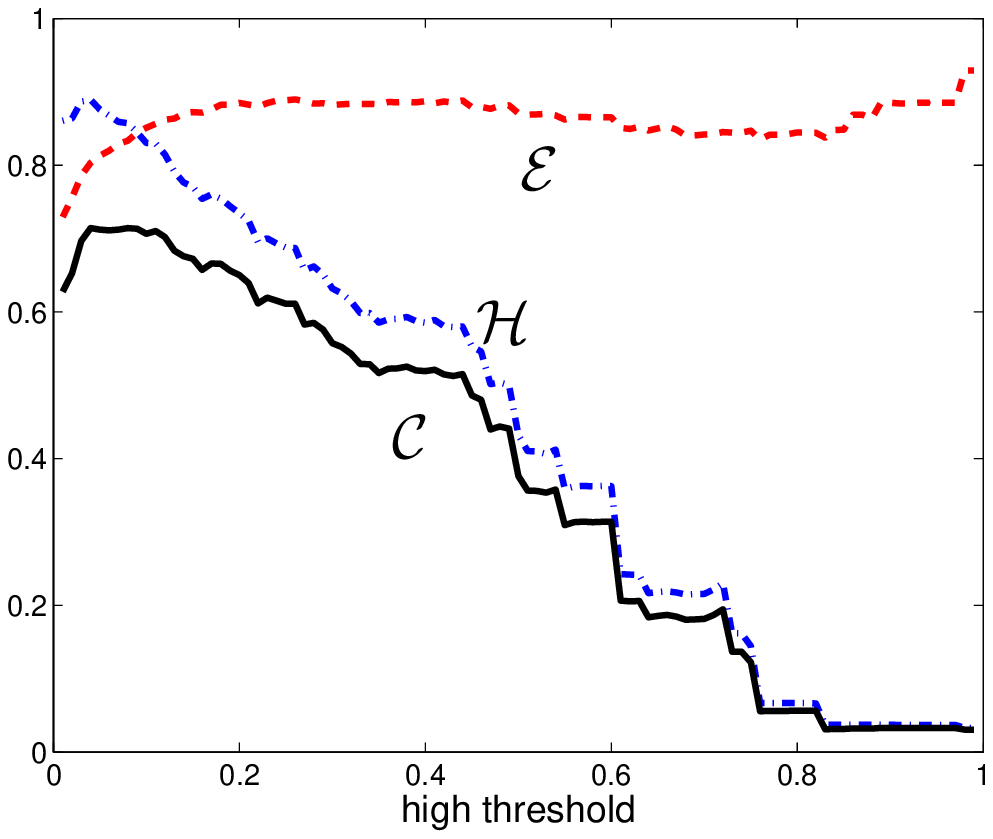}&\includegraphics[scale = 0.3]{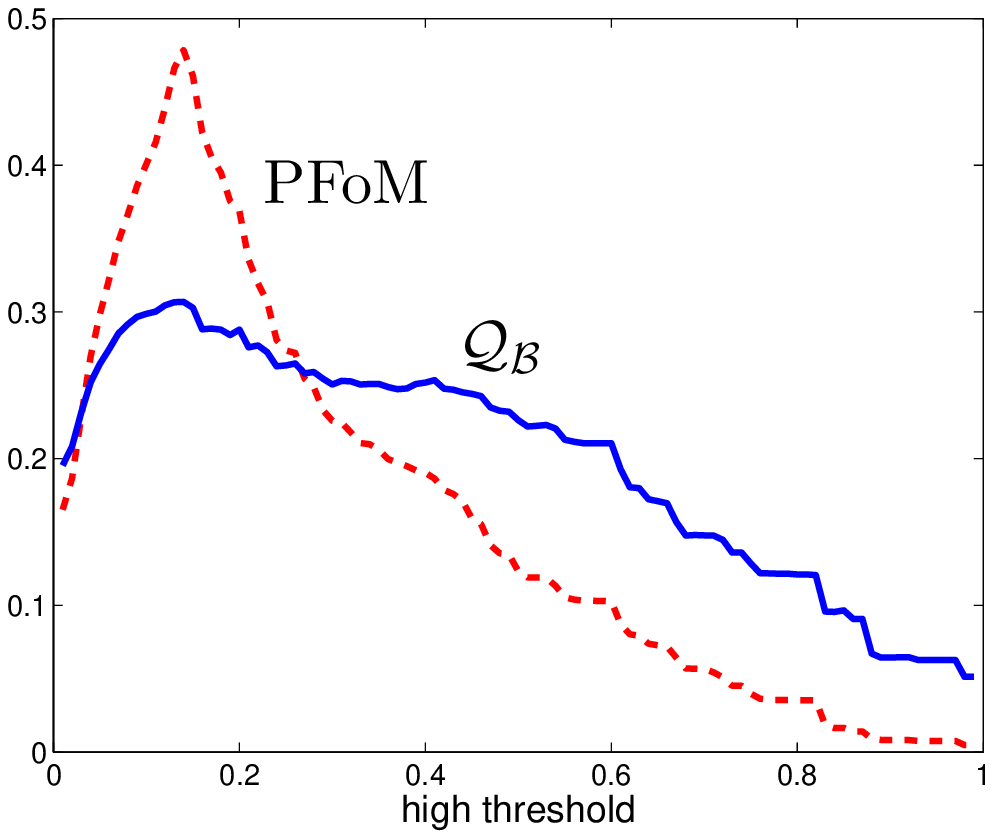}\\
(a)&(b)\\\check{}
\includegraphics[scale = 0.3]{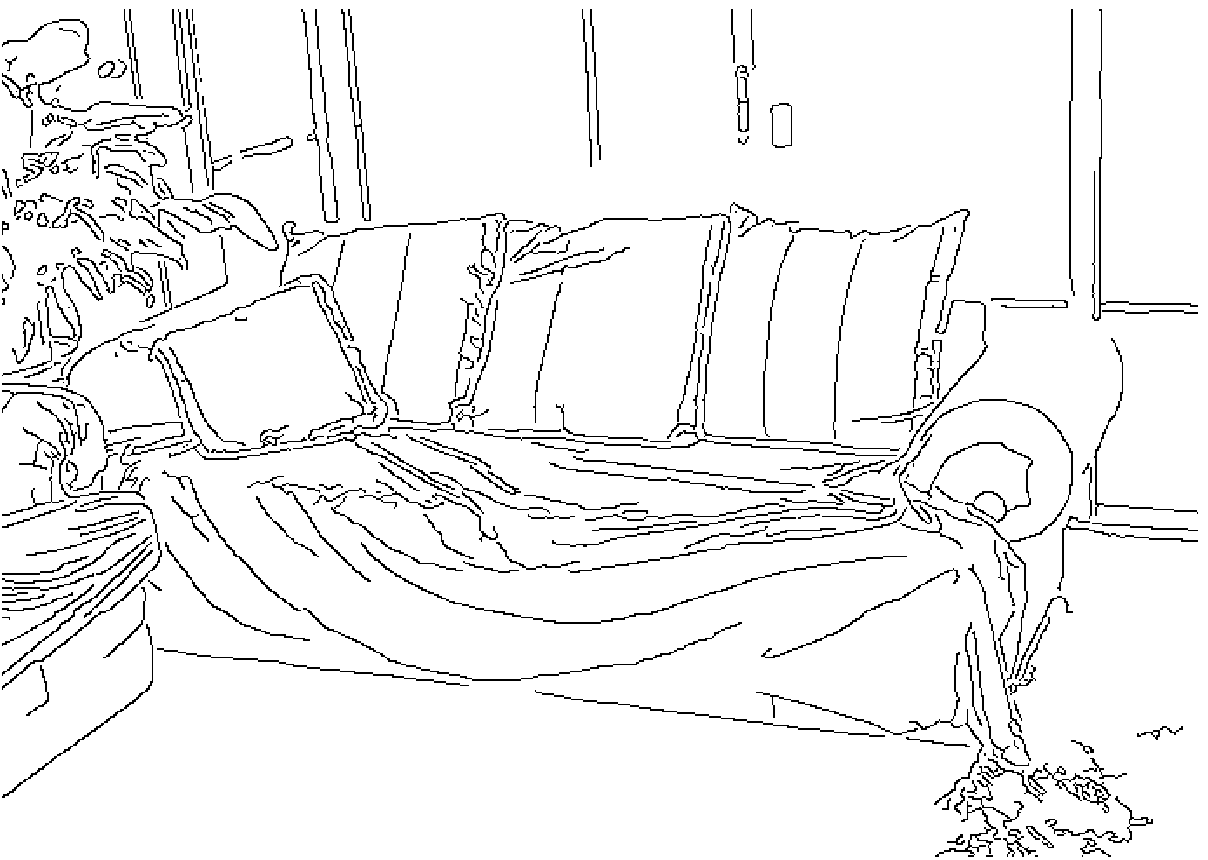}&\includegraphics[scale = 0.3]{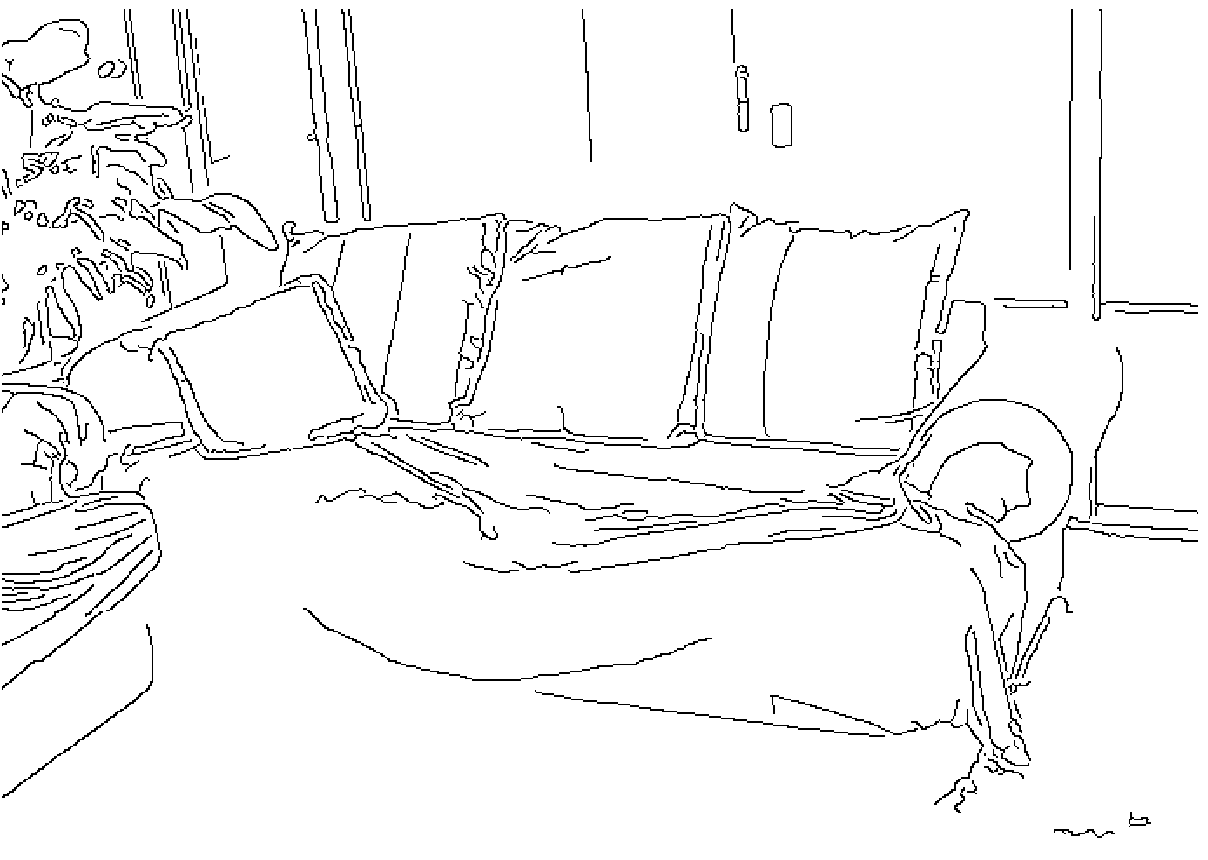}\\
(c)&(d)\\
\end{array} \]
\vspace{-0.5cm}
\caption{ Canny EDA quality curves of image \emph{109}. $(a)$  Plot of $\mathcal{E}$, $\mathcal{H}$ and $\mathcal{C}$ vs high threshold $T_h$, $(b)$ Plot of  PFoM and $\mathcal{Q}_{\mathfrak{B}}$ vs $T_h$, $(c)$~Best map according to $\mathcal{C}$ (score $0.714$) and $(d)$~Best map according to  PFoM (score $0.4784$) and $\mathcal{Q}_{\mathfrak{B}}$ (score $0.3068$). Canny EDA parameters are high threshold $T_h$=0.08 and 0.14 respectively, low threshold $T_l=0.4 \times T_h$ and $\sigma=\sqrt2$. }
\label{grafico_CE_i109_2}
\end{figure}

The evaluation curves, constructed with the values of the  $\mathcal{E}$, $\mathcal{H}$ and $\mathcal{C}$ over the collection of EDA outputs as a function of the $T_h$ values, are shown in Figure~\ref{grafico_CE_i109_2} along with a  plot of PFoM and  $\mathcal{Q}_{\mathfrak{B}}$ over the same parameter range. The $\mathcal{C}$ evaluation curve displays the usual behavior of complexity measures; it shows a peak when both measures are balanced.
\begin{figure}[h!]
 \centering
 \[\begin{array}{cc}
\includegraphics[scale = 0.4]{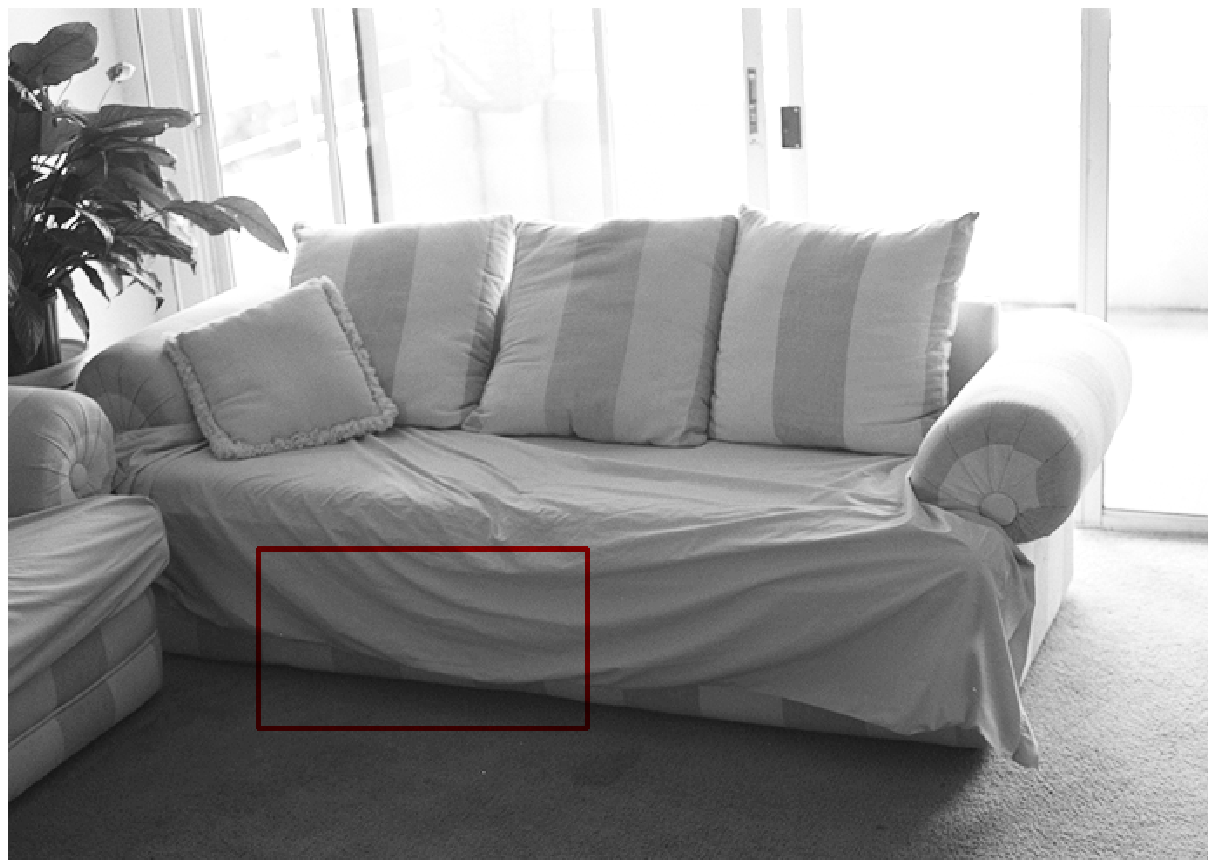}&\includegraphics[scale = 0.3]{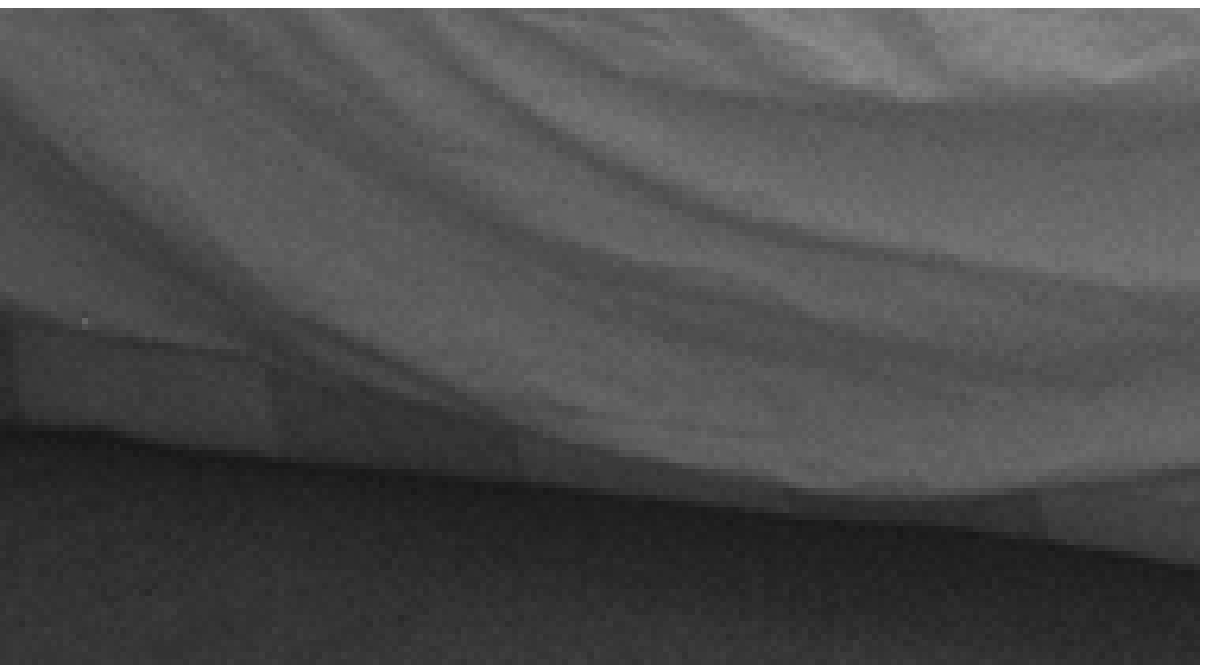}\\
(a)&(b)\\
\includegraphics[scale = 0.3]{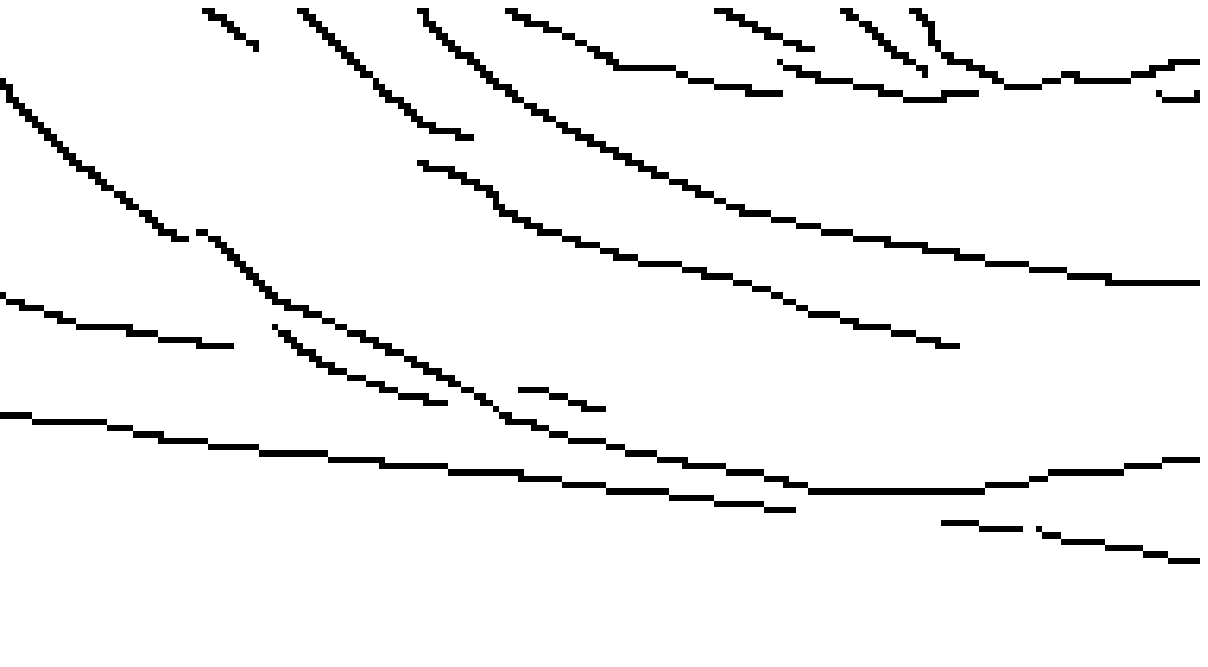}&\includegraphics[scale = 0.3]{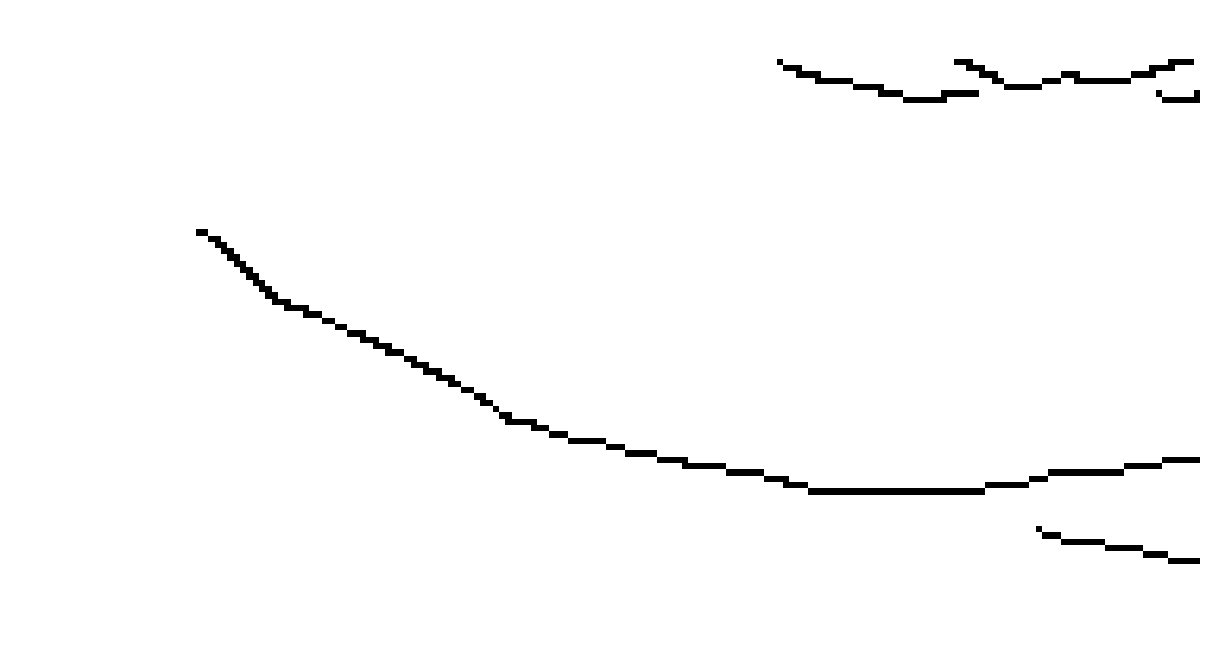}\\
(c) &(d)\\
\end{array} \]
 \vspace{-0.5cm}
\caption{ $(a)$~Image \emph{109}; $(b)$ Enlarged view of the  marked region in $(a)$; $(c)$~Enlarged view of the region shown in $(b)$, extracted from best Canny map according to $\mathcal{C}$ ; $(d)$~Enlarged view of the region shown in $(b)$, extracted from best Canny map according to PFoM.  }
\label{grafico_CE_i109_3}
\end{figure}
Qualitative comparisons between best edge maps according to $\mathcal{C}$ and PFoM are made, see Figure~\ref{grafico_CE_i109_3}. The boundary in the lower part of the coach is missing in the edge map selected by PFoM, (Figure \ref{grafico_CE_i109_3}~$(d)$). Our measure selected a more defined edge map, i.e. with more edge points, (Figure \ref{grafico_CE_i109_3}~$(c)$), thus showing a more defined contour around the couch.   Also, comparing the  thresholds value, the $\mathcal{C}$ measure selected a map with $T_h=0.08$ and PFoM  a map with higher threshold,  $T_h=0.14$.  PFoM was reported as a measure with a bias towards false negatives, i.e. it gives higher scores to edge maps with few edges \cite{Lopez_Molina}. This account for the missing  edge boundary points around the main object in the PFoM edge map.

The second image, \emph{woods}, provides a challenge to all the detectors. The edge maps selected by $\mathcal{C}$, PFoM and  $\mathcal{Q}_{\mathfrak{B}}$ are shown in Figure~\ref{woods2}.  Maps are qualitatively different, selected from different quartiles of the parameter range. Reference-based measures select: PFoM to Robert EDA as the best edge detector  and $\mathcal{Q}_{\mathfrak{B}}$ to Sobel EDA. Table~\ref{tabla_pratt_q_all_ed} shows their parameters value and scores. The scores are low, i.e. indicating low agreement between the outputs and GT.

\begin{table}[h]
\centering
\begin{tabular}{|c||c|c||c|c|}
  \hline
         EDA        & EDA optimal parameters               &   PFoM            & EDA optimal parameters &  $\mathcal{Q}_{\mathfrak{B}}$ \\
\hline
  Canny                       &$T_l$=0.228 $T_h$=0.57  & 0.5082            &$T_l$=0.264  $T_h$=0.66    &  0.3257  \\
  LoG                         &$T$=0.0096                      & 0.4545            &$T$=0.01                        &  0.2728  \\
  Prewitt                     &$T$=0.0096                                & 0.5479            &$T$=0.1                                    &  0.3961   \\
  Roberts                     &$T$=0.0096                                & \textbf{0.5565}   &$T$=0.072                                  &  0.3583   \\
  Sobel                       &$T$=0.1                                   & 0.5526            &$T$=0.096                                  &  \textbf{0.3978}   \\
            \hline
\end{tabular}
\caption{Maximum scores of PFoM and $\mathcal{Q}_{\mathfrak{B}}$  evaluation curves for all EDA, computed over image \emph{woods}.(Maximum scores by column are highlighted in bold typeface.)}
\label{tabla_pratt_q_all_ed}
\end{table}

\begin{table}[h]
\centering
\begin{tabular}{|c|p{3cm}|c|c|c|c|c|}
  \hline
         EDA              & $\mathcal{C}$-based optimal parameters                 &    $\mathcal{E}$     &$\mathcal{H}$   & $\mathcal{C}$      & PFoM           &$\mathcal{Q}_{\mathfrak{B}}$   \\
				\hline
  Canny                             &$T_l$=0.076 $T_h$=0.19   & 0.7472               &\textbf{0.9646} &\textbf{ 0.7208}    &0.3061          &0.2876          \\
  LoG                               &$T$=0.0076                      & \textbf{0.7760}               &0.9201 & 0.7139             &0.3951          &0.2654          \\
  Prewitt                           &   $T$=0.064                               & 0.7384               & 0.9262         & 0.6839             &\textbf{0.4197} &\textbf{0.3787} \\
  Roberts                           & $T$=0.0052                                & 0.6515               & 0.9368         & 0.6104             &0.3649          &0.3348          \\
  Sobel                             &  $T$=0.064                                & 0.7326               & 0.9293         & 0.6808             &0.4137          &0.3775          \\
  \hline
\end{tabular}
\caption{ $\mathcal{E}$, $\mathcal{H}$ and $\mathcal{C}$ scores of  best EDA outputs of image \emph{woods}. $\mathcal{Q}_{\mathfrak{B}}$ and PFoM scores correspond to the best edge map according to $\mathcal{C}$.(Maximum scores by column are highlighted in bold typeface.)}
\label{tabla1}
\end{table}

The analysis of the scores given by PFoM and  $\mathcal{Q}_{\mathfrak{B}}$ to the edge maps selected by $\mathcal{C}$, (Table \ref{tabla1}, sixth to seventh column) reveals that the  reference-based measures heavily  penalize  the false positives that are introduced by outlining the woods and country road surrounding the buildings in the image, which are not depicted in the GT. Table~\ref{tabla1}, third to fifth column,  show $\mathcal{E}$, $\mathcal{H}$ and $\mathcal{C}$ scores, revealing that the most balanced map related to $\mathcal{E}$ is given by LoG EDA; the map with most edge entropy, i.e.  related to $\mathcal{H}$, is given by Canny EDA and the one that shows better statistical complexity,i.e. related to $\mathcal{C}$, is also Canny EDA. The scores given by PFoM and $\mathcal{Q}_{\mathfrak{B}}$ to such maps select as the best map the one given by Prewitt EDA.
 An interesting comparison is given by the scores of $\mathcal{E}$ and $\mathcal{Q}_{\mathfrak{B}}$. The \emph{Equilibrium} index is based on $\mathcal{Q}_{\mathfrak{B}}$ by replacing the GT with a family of preselected local patterns, sought in the whole image. The provided GT, which does not outline all the objects in the image, misleads the reference-based measures towards lighter maps while the use of pre-defined edge patterns helps preventing such problem.

\begin{figure}[h!]
\[\begin{array}{cccc}
 \ \includegraphics[scale = 0.2]{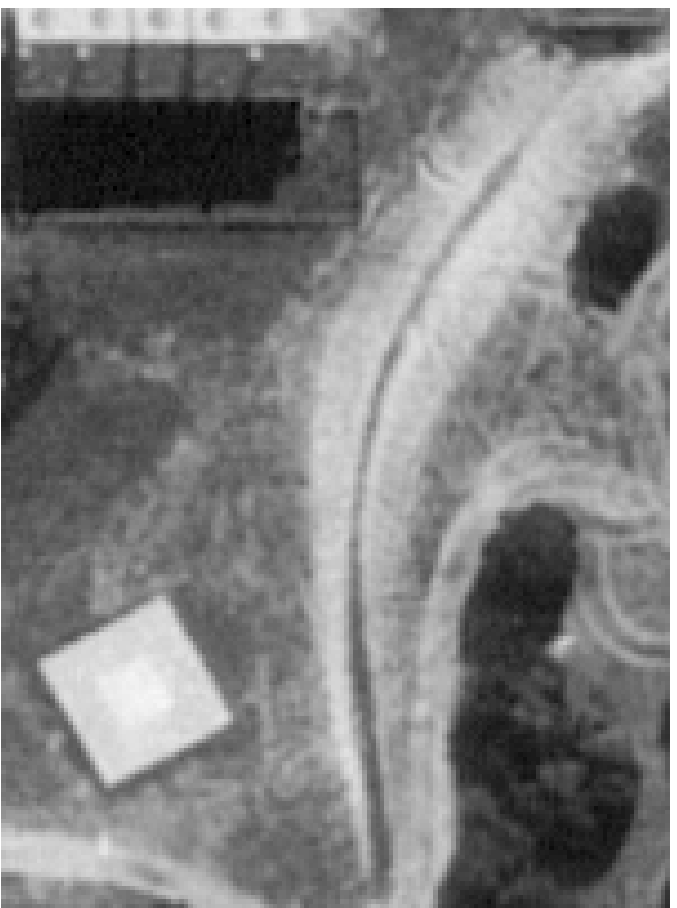} \ &\includegraphics[scale = 0.2]{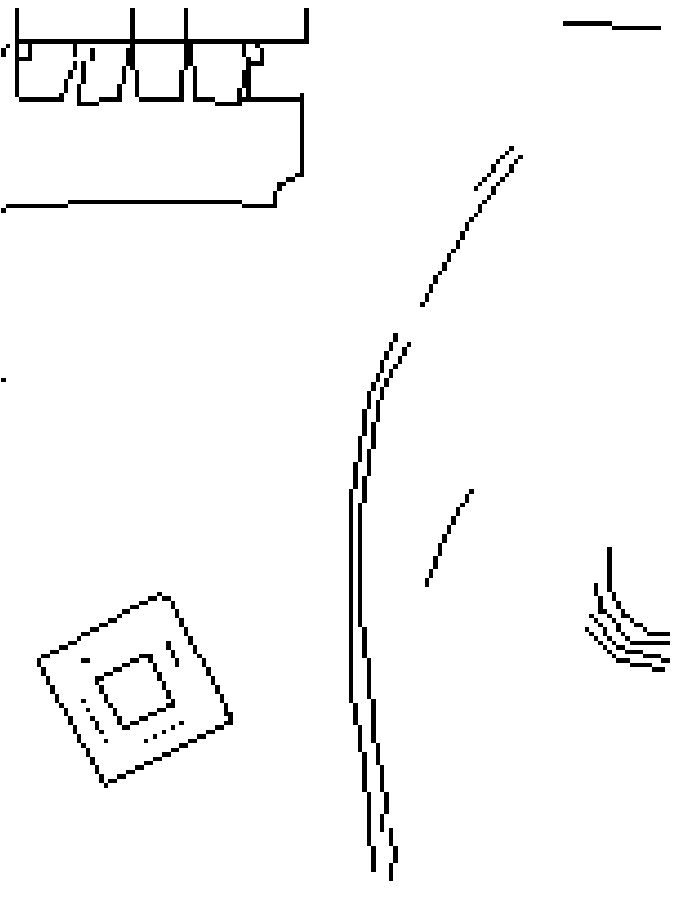} \
& \ \includegraphics[scale = 0.2]{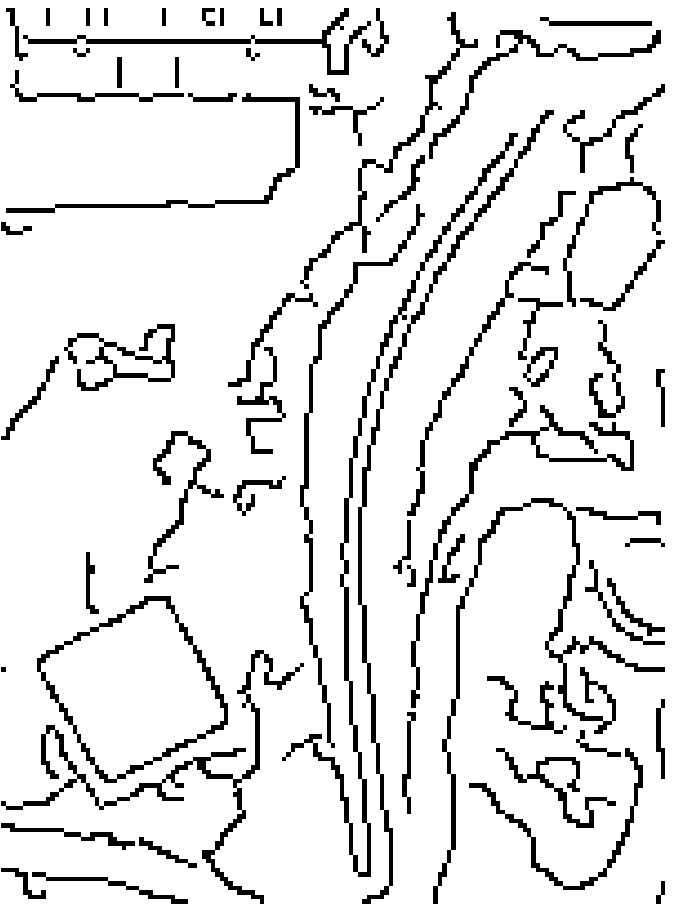} \ & \ \includegraphics[scale = 0.2]{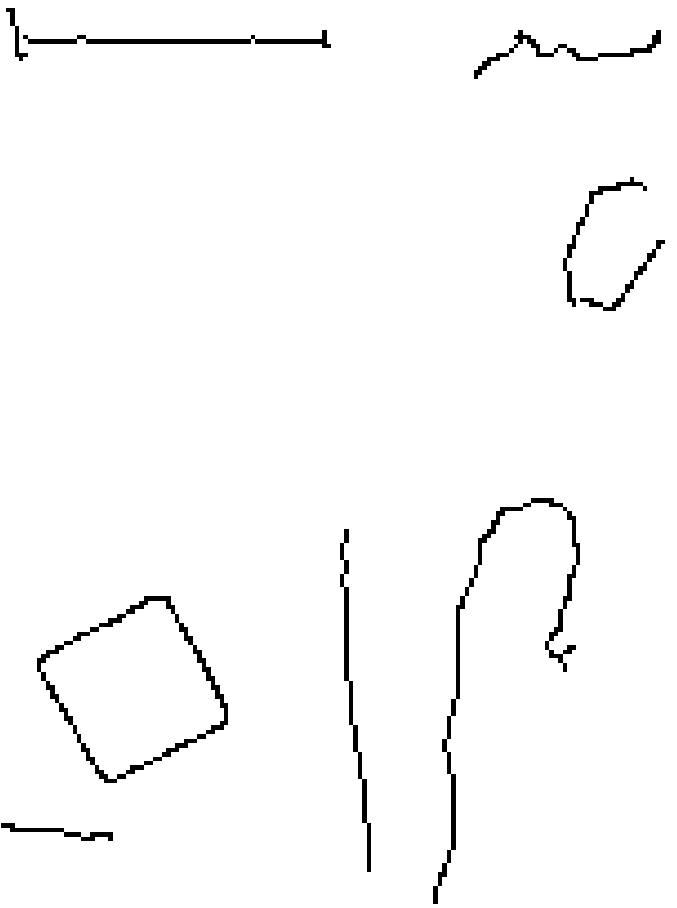} \ \\
(a)&(b)&(c)&(d)\\
\includegraphics[scale = 0.2]{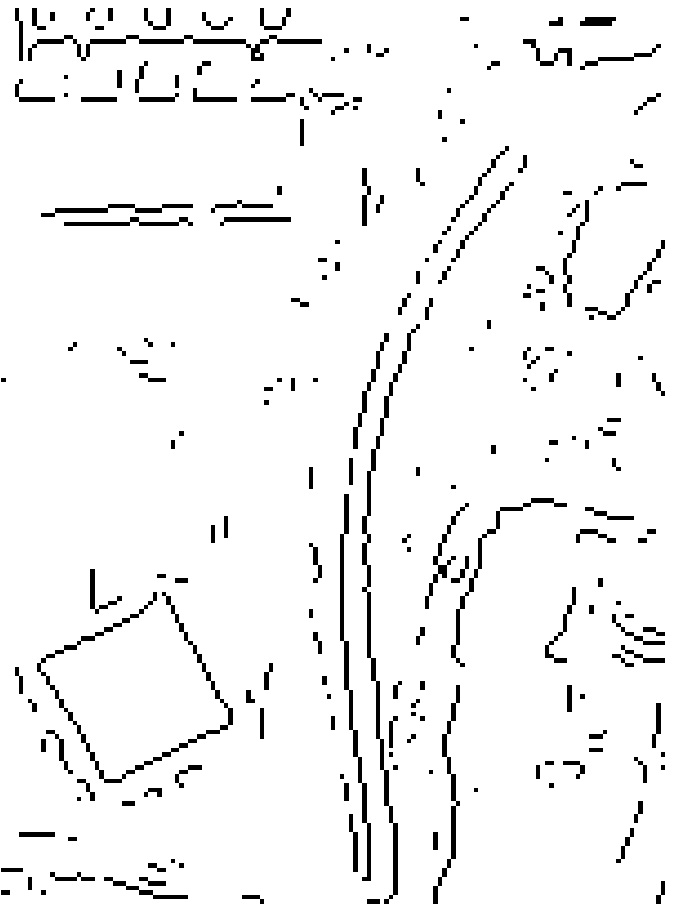} \ & \ \includegraphics[scale = 0.2]{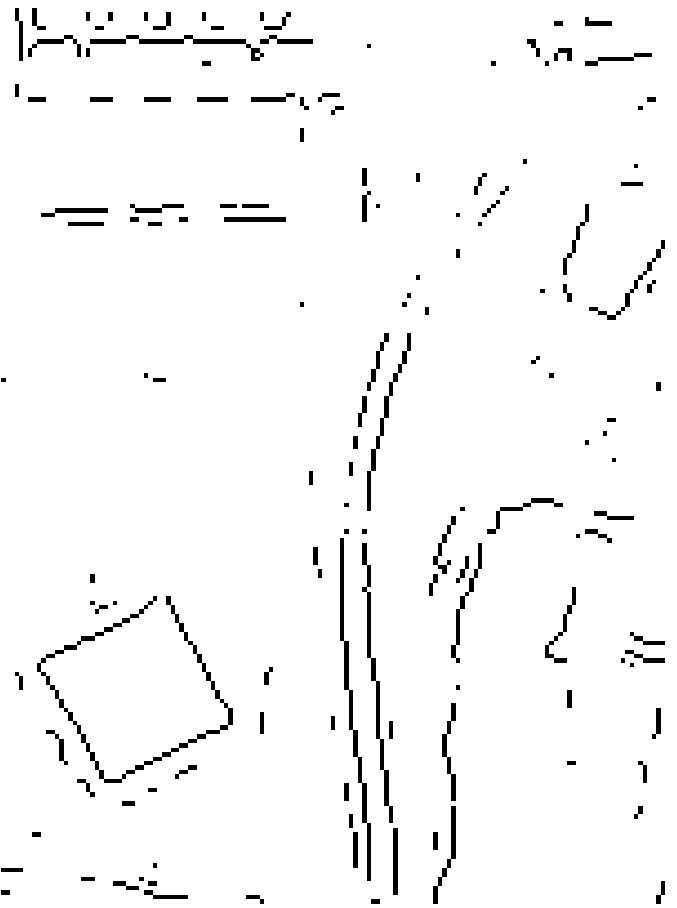} \
& \ \includegraphics[scale = 0.2]{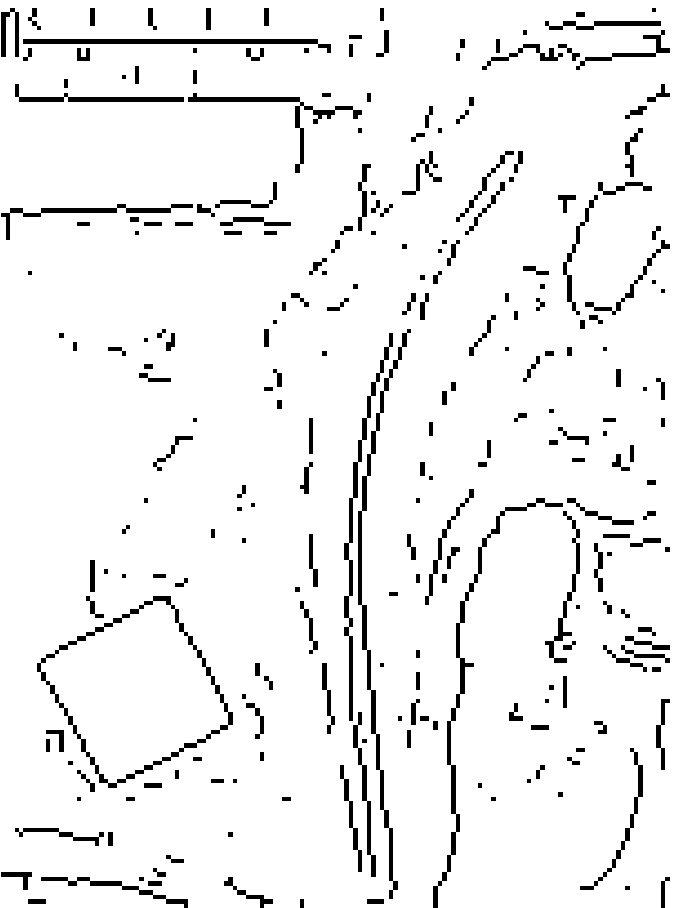} \ & \ \includegraphics[scale = 0.2]{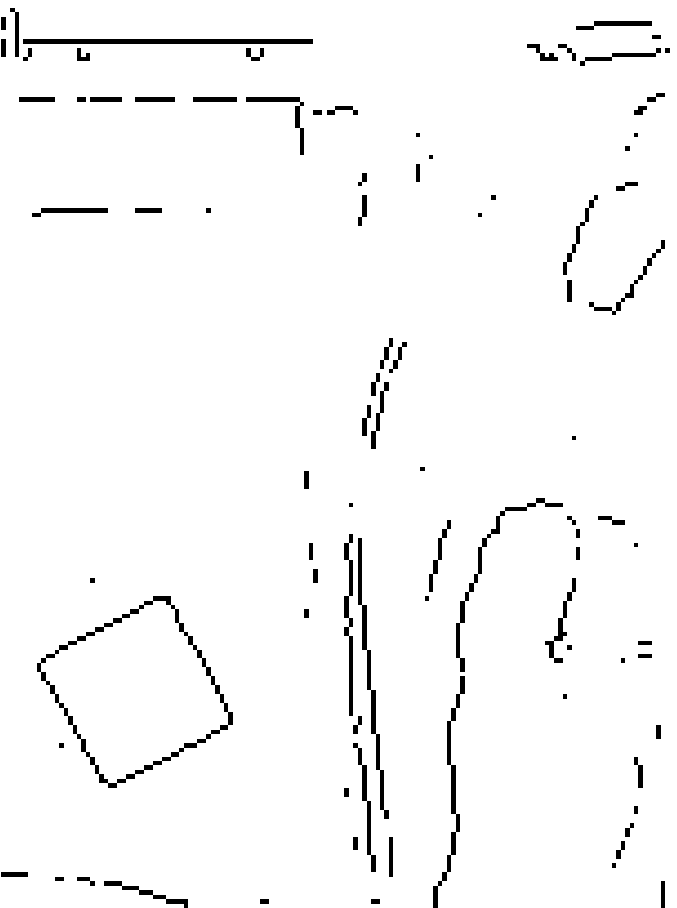} \ \\
(e)&(f)&(g)&(h)\\
 \ \includegraphics[scale = 0.2]{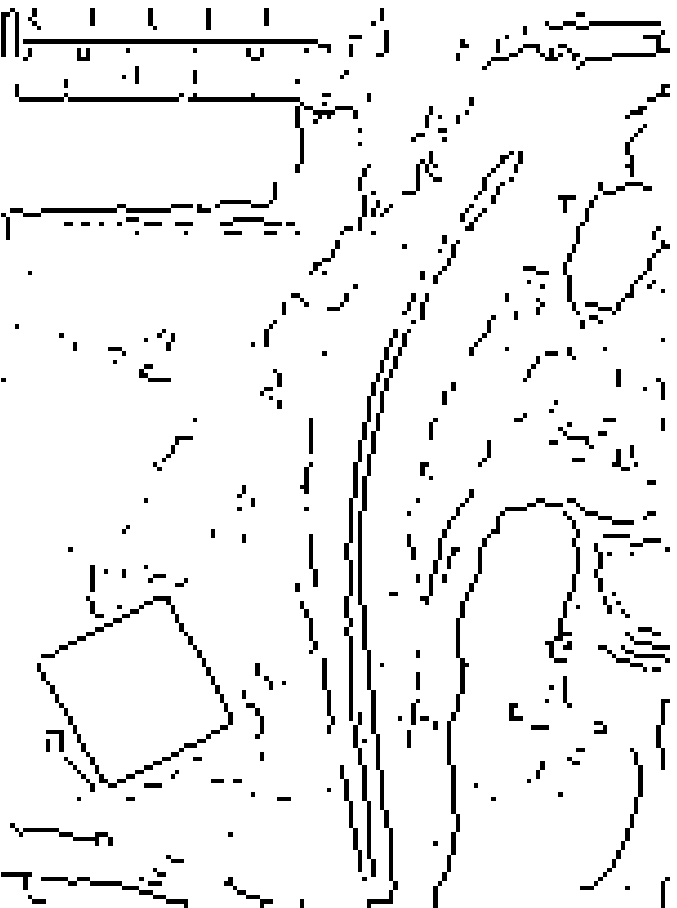} \ & \ \includegraphics[scale = 0.2]{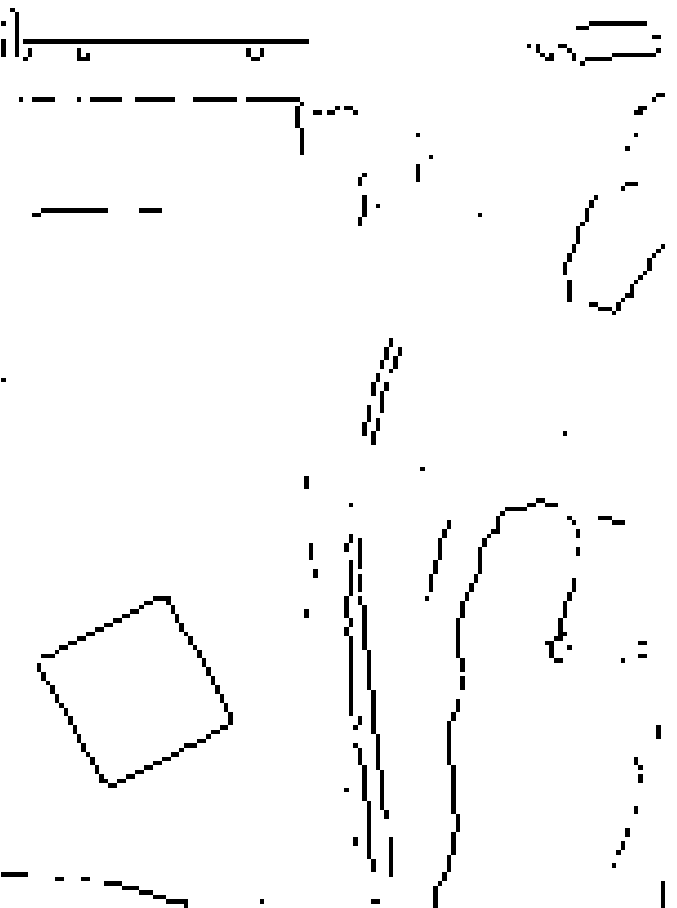} \
& \ \includegraphics[scale = 0.2]{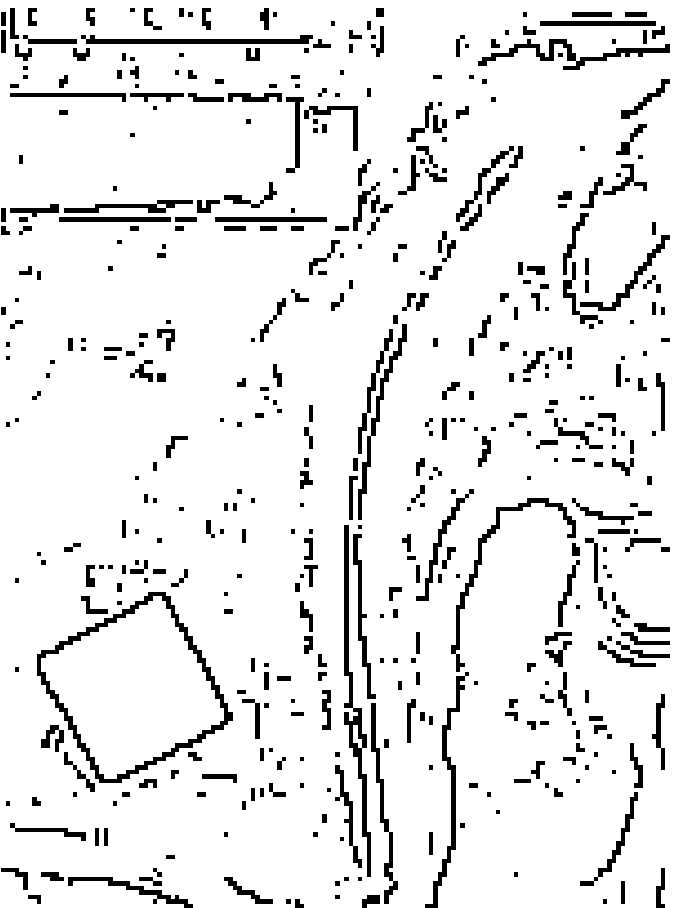} \ & \ \includegraphics[scale = 0.2]{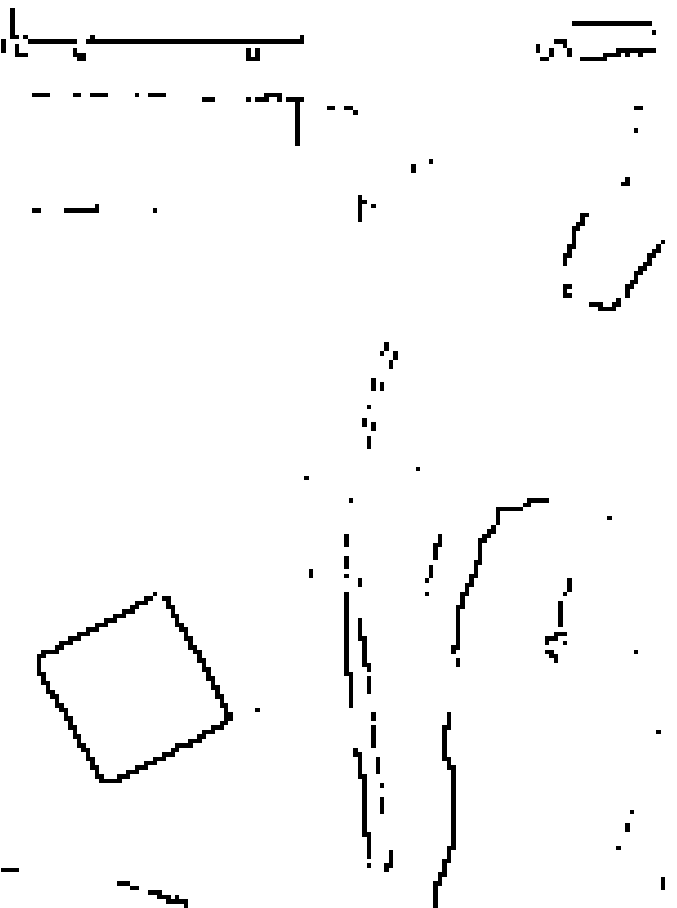} \ \\
(i)&(j)&(k)&(l)\\
\end{array} \]
\caption{$(a)$ Enlarged view of the southeast corner of the image \emph{woods}  depicting a country road. $(b)$ GT from $(a)$.  $(c)$ and $(d)$ are Canny edge maps, $(e)$ and $(f)$ LoG edge maps, $(g)$ and $(h)$ Prewitt edge maps, $(i)$ and $(j)$ Sobel edge maps, $(k)$ and $(l)$ Roberts edge maps. The first of each pair was selected by  $\mathcal{C}$ and the second of each pair by PFoM.}
\label{zoom}
\end{figure}

An enlarged view of the southeast corner of the image \emph{woods} depicting a country road is shown in Figure \ref{zoom})$(a)$ with its corresponding GT in  panel $(b)$.  Panel $(c)$ and $(d)$ are Canny edge maps, $(e)$ and $(f)$ LoG edge maps, $(g)$ and $(h)$ Prewitt edge maps, $(i)$ and $(j)$ Sobel edge maps, $(k)$ and $(l)$ Roberts edge maps. The first of each pair was selected by  $\mathcal{C}$ and the second of each pair by PFoM.

In each edge map selected by the reference-based measures, the country road is poorly defined. Instead, every edge map selected by $\mathcal{C}$, Figure~\ref{zoom} $(c)$, $(e)$, $(g)$, $(i)$ and $(k)$ have the country road well defined, as well as the vegetation surrounding it.

\begin{figure}[h!]
\centering
 \[\begin{array}{cc}
 \includegraphics[scale = 0.40]{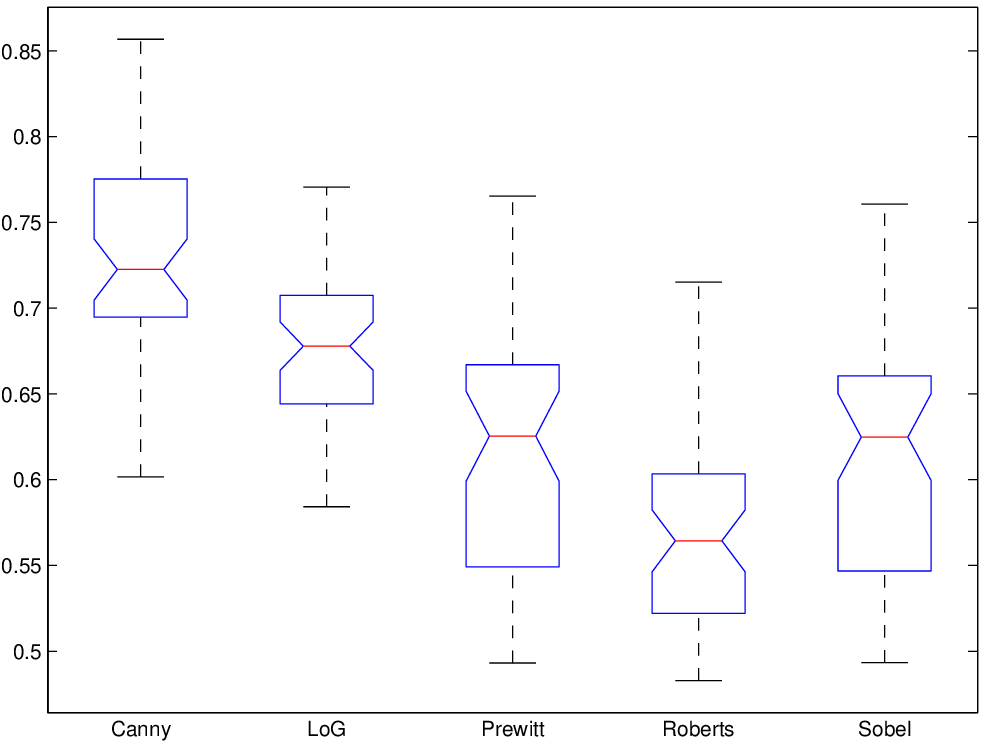}&\includegraphics[scale = 0.40]{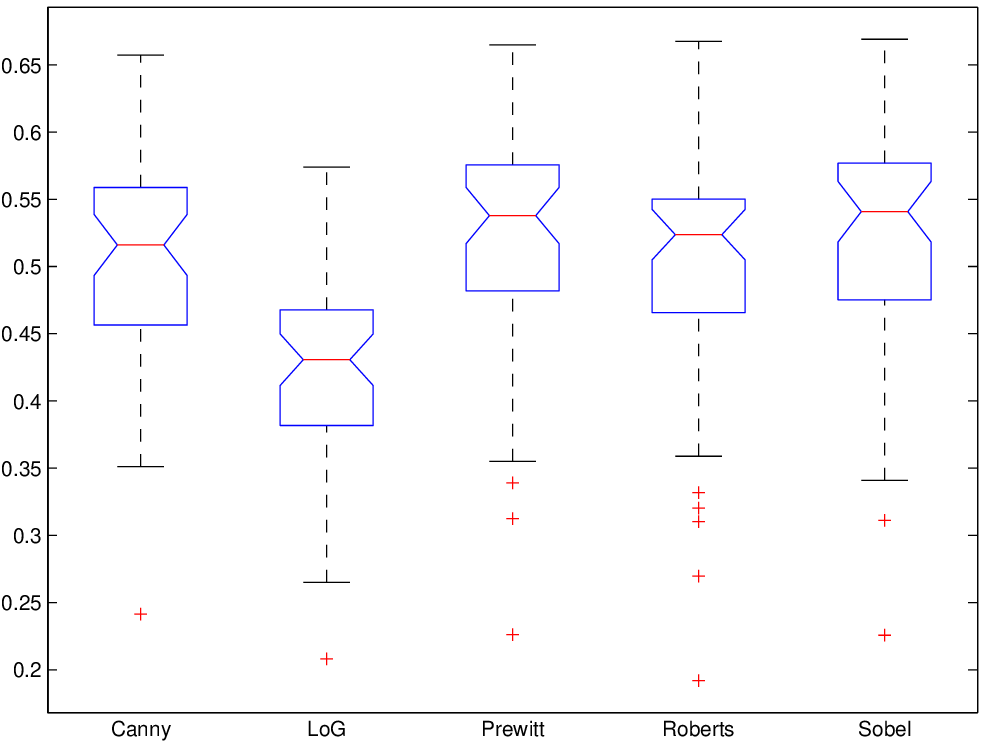}\\
(a)&(b)\\
 \end{array} \]
  \vspace{-0.5cm}
 \caption{$(a)$ Boxplot of  maximum $\mathcal{C}$ score values, $(b)$  Boxplot of PFoM scores computed on edge maps selected by $\mathcal{C}$. Scores are computed over all 50 images of the first collection of South Florida database.}
 \label{boxplot}
\end{figure}

Finally, we discuss the empirical statistical distribution of the maximum $\mathcal{C}$ scorings on all EDA maps computed with all South Florida images. Boxplots of such values are shown in Figure \ref{boxplot}~along with boxplots of PFoM scores computed on the same edge maps, the ones selected by our $\mathcal{C}$ measure. We infer from comparison between all $\mathcal{C}$ empirical distributions  that Canny EDA produces slightly better maps than the other detectors. PFoM score differently such maps giving the same moderately low mean (around 0.5 value) to all gradient detectors but LoG EDA. We show examples of such images and best EDA maps in Figure~ref{otras}.
Each row displays outputs from a different EDA. Last panel of each row shows a plot of the $\mathcal{C}$ curve as a function of high threshold. The different shapes of the $\mathcal{C}$ curves can be accounted for the differences in the sampling of each EDA parameter space.

\subsection{Second experiment: Image with multiple GT images}
\begin{figure}[h]
 \[\begin{array}{ccccc}
 \includegraphics[scale = 0.25]{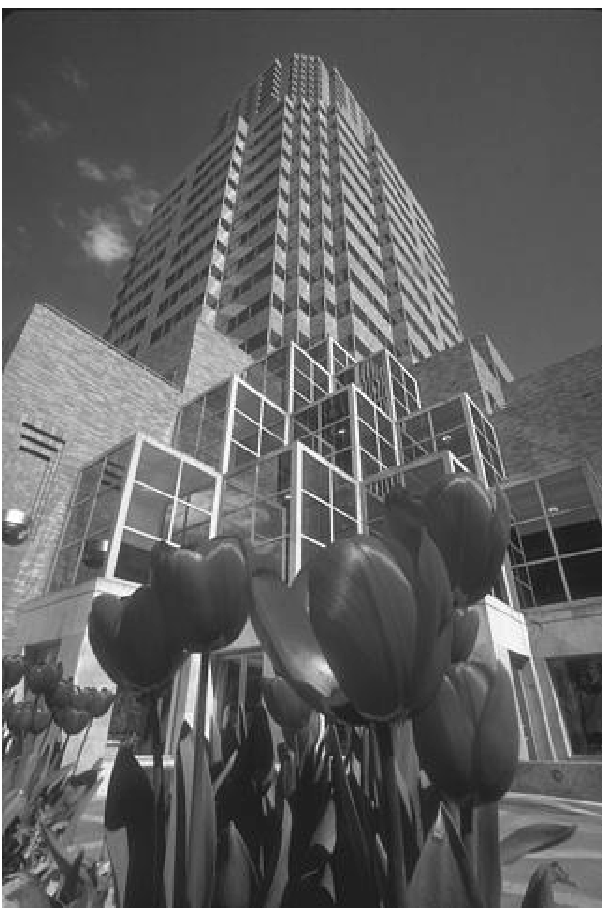}&\includegraphics[scale = 0.25]{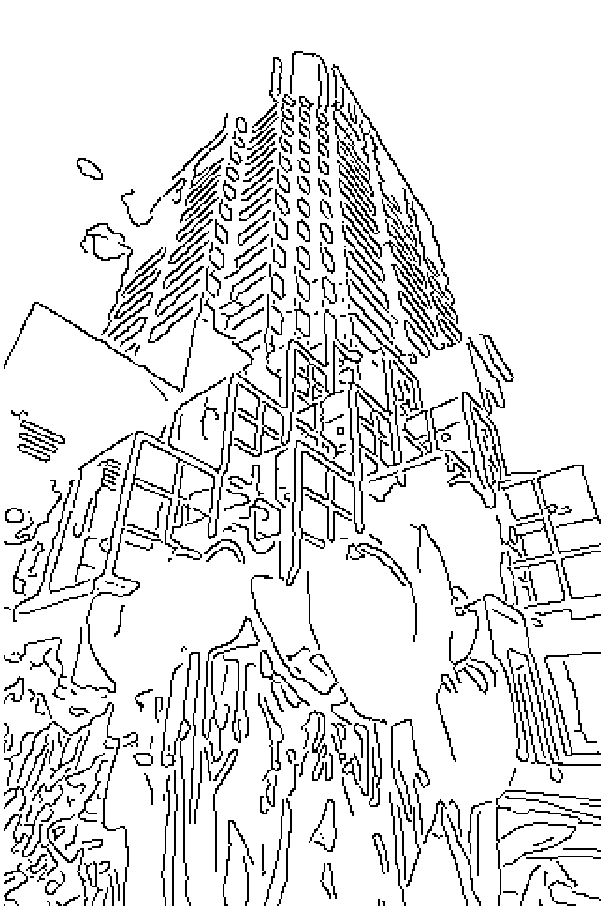}&\includegraphics[scale = 0.25]{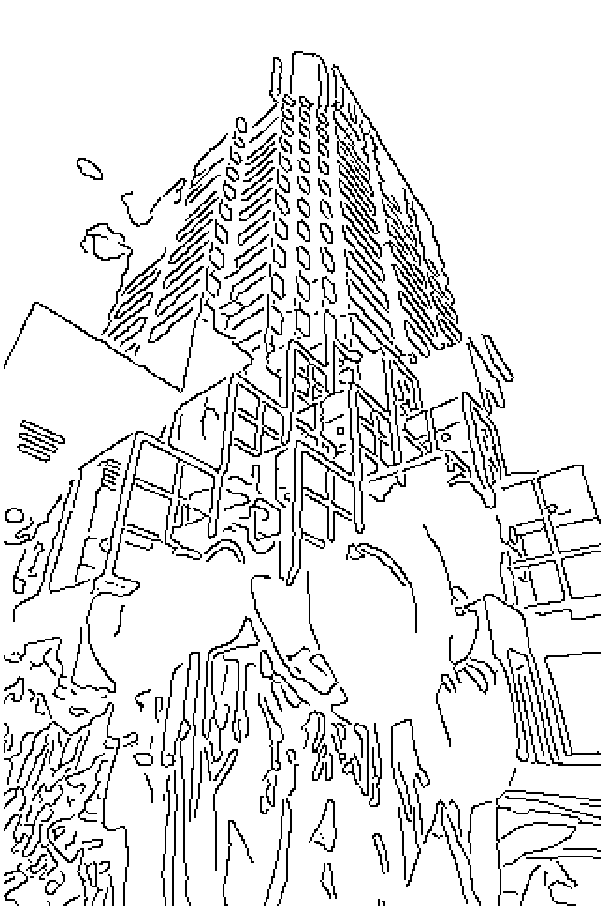}&\includegraphics[scale = 0.25]{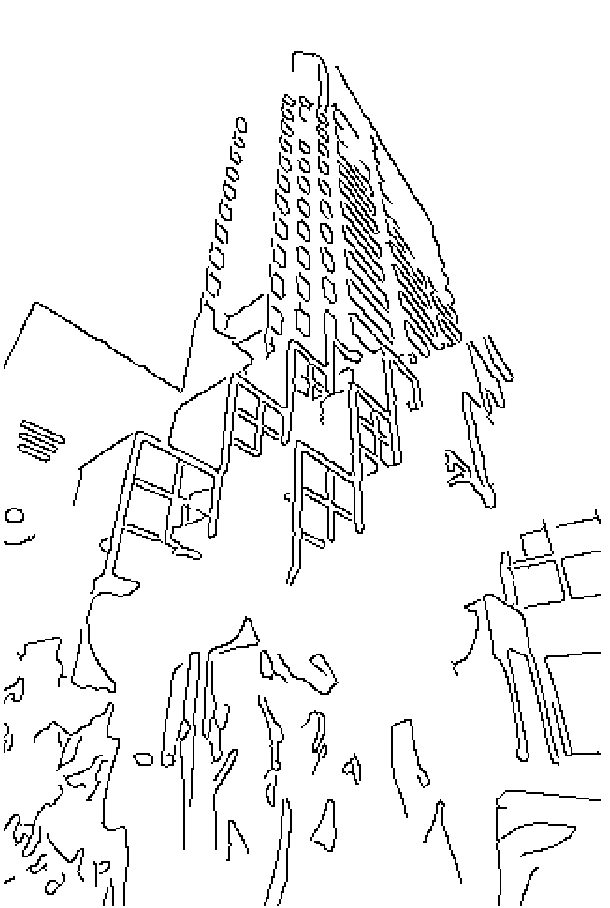}&\includegraphics[scale = 0.25]{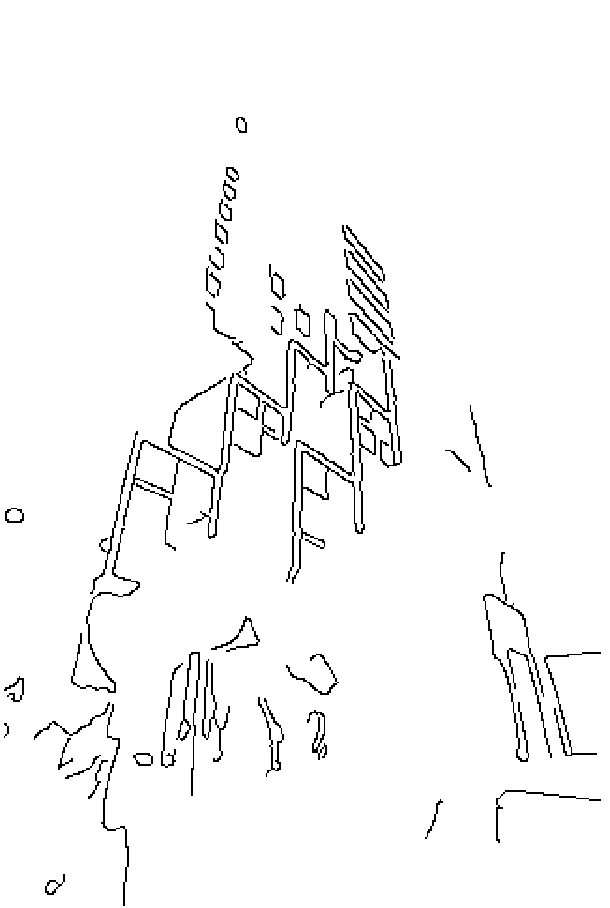}\\
 (a)&(b)&(c)&(d)&(e)\\
 \includegraphics[scale = 0.25]{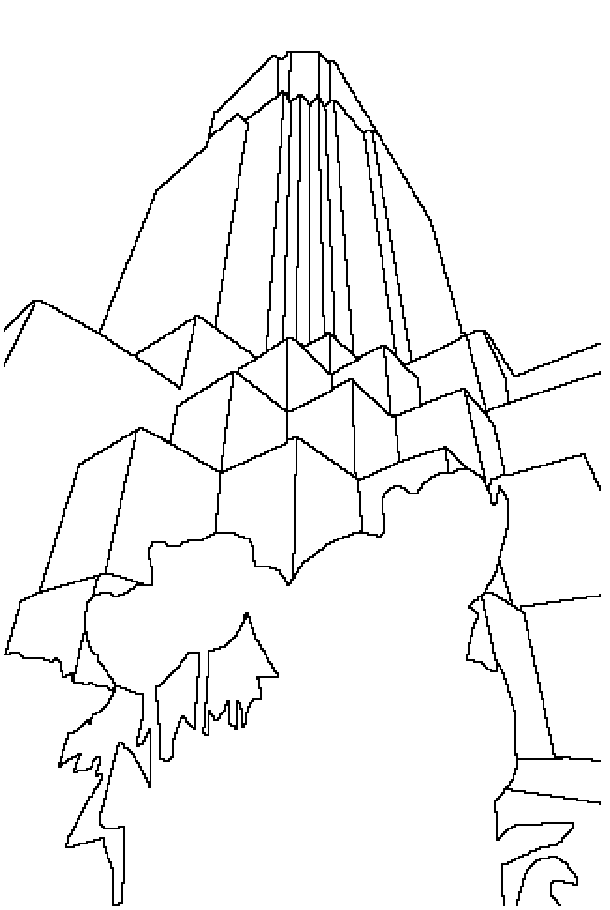} &\includegraphics[scale = 0.25]{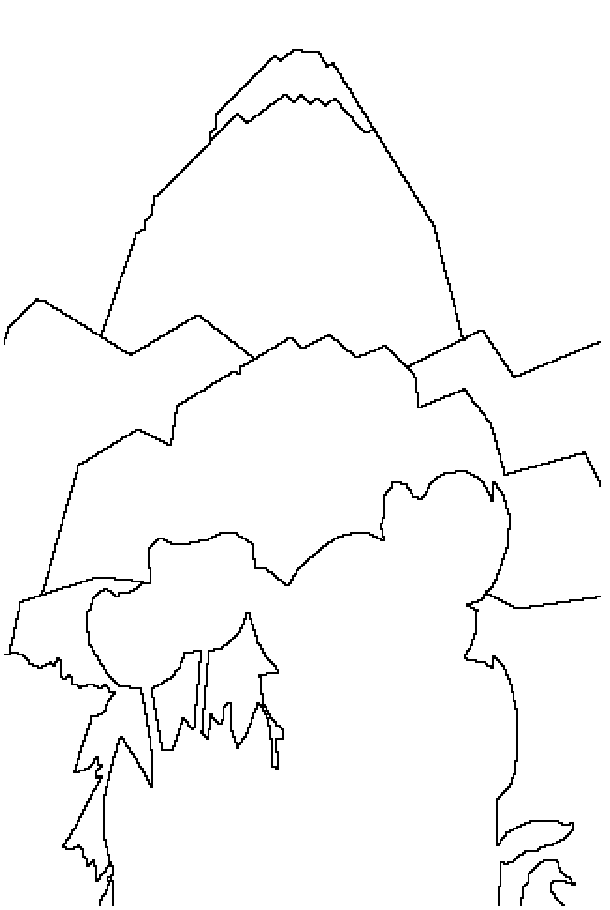}&\includegraphics[scale = 0.25]{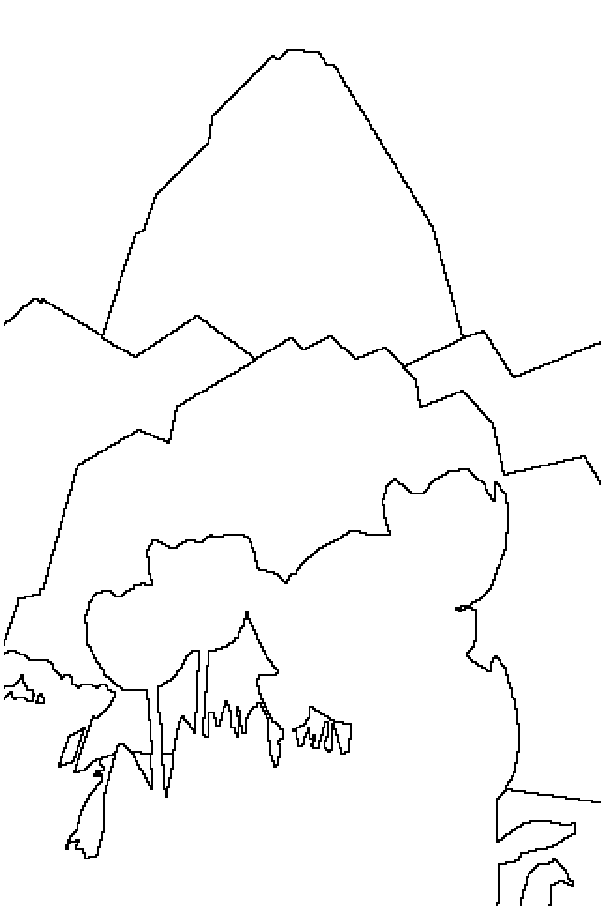}&\includegraphics[scale = 0.25]{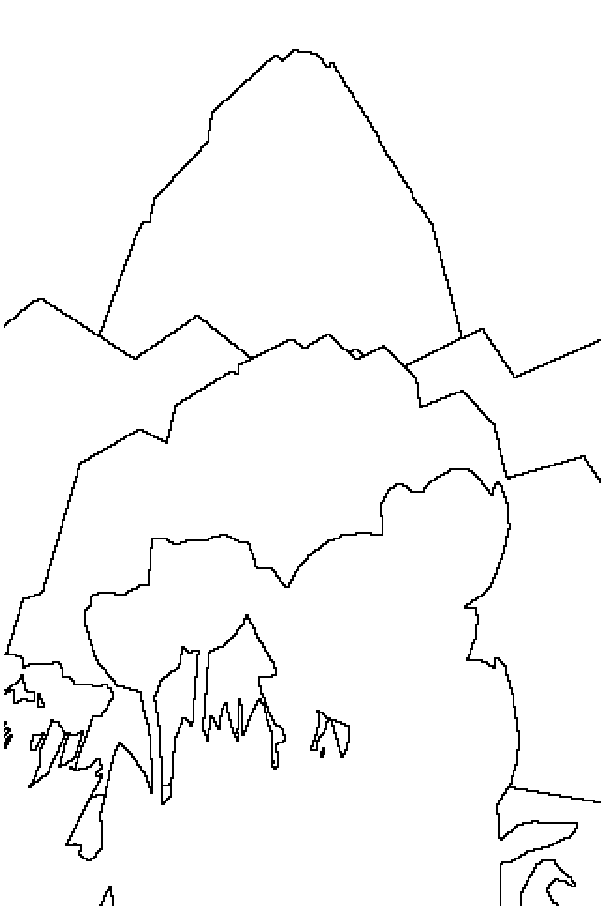}&\includegraphics[scale = 0.25]{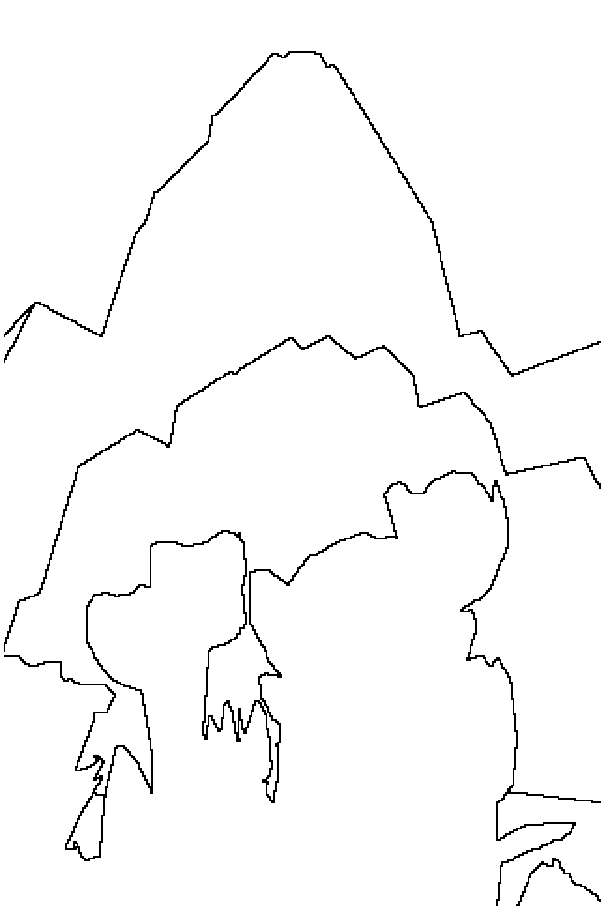}\\
(f) \ \ GT1&(g) \ \ GT2 &(h) \ \ GT3&(i) \ \ GT4&(j) \ \ GT5\\
 \end{array} \]
 \vspace{-0.5cm}
 \caption{ $(a)$ Image \emph{86000}; $(b)$ best map selected with measure $\mathcal{C}$; $(c)$ best map selected by PFoM with GT1; $(d)$ best map selected by $\mathcal{Q}_{\mathfrak{B}}$ with all GT; $(e)$ best map selected by PFoM with GT3. $(f)$-$(j)$ are respectively (GT1)-(GT5) GT images from the Berkeley database.}
 \label{edificio}
\end{figure}

Some authors believe that the manual GT approach is essential for performance characterization  because most researchers
do not regard results on synthetic images as convincing and still wish to see results on real images, see \cite{Fernandez2008} and references therein. Unfortunately,
 manual GT annotation is dubious, tedious and time-consuming. In addition, different annotators
often give different GT or the same annotator can give different GT to the same real image at
different times \cite{Berkeley_database}. The use of \emph{consensus} GT for real images avoids
both (1) the subjective generation of manual GT for real images and (2) the generation
of artificial GT for artificial images, which probably do not faithfully represent the real scenes. On the
other hand, the consensus GT allows many real images to be used for performance characterization, \cite{Fernandez2008, Yitzhaky2003}.

In this experiment, we show that the use of our measure gives results similar to the ones obtained when a pool of GT images are used in the intra-technique and in the inter-technique evaluation problem.

We selected an image from the Berkeley Segmentation Database \cite{Berkeley_database}, a benchmark database for boundary detection algorithms that provides images with several hand made segmentations offered as GT. Thus, the level of detail of the  different GT segmentations is diverse, and it represents the human opinion on what the boundary edges of the objects in images are. Besides, edge detection is not the same as boundary detection; boundary maps show only the outline of main objects while edge maps show the whole structure of the image. In Figure~\ref{edificio}  five different GT images available for the image \emph{86000} are shown.

All supervised measures depend on the level of detail of the GT, thus any supervised measure computed with  GT1 (highly detailed) will give high marks to a more cluttered edge map, but if computed by using GT5 (little detailed) it will certainly give a maximum score to a map with very few edge points.

Unsupervised measures score differently; they search general characteristics in the map, determined in this case by the specific database of edge patterns and the KS statistic. This example aims at exploring the degree of matching of $\mathcal{C}$  scoring with human observation and supervised measures. Thus we use only the gold standard supervised measure, PFoM, and the gold standard EDA, Canny EDA, to elaborate the example.

We obtained $\mathcal{C}$ scores  for the five GT images and the 100 edge maps outputs of Canny EDA computed with parameters described in the previous subsection. The same 100 Canny edge maps were scored with PFoM, by using all different hand-made GT images and they were scored with $\mathcal{Q}_{\mathfrak{B}}$ using different collections of GT images. Our reference-based measure $\mathcal{Q}_{\mathfrak{B}}$ provides a \emph{consensus score} in this framework.   In Figure~\ref{edificio} $(a)$,  the best map selected by $\mathcal{C}$ is shown. In panel $(b)$ the optimal edge map selected using PFoM with GT1 image is shown. That map was also selected with the supervised measure $\mathcal{Q}_{\mathfrak{B}}$ using the collection of all man-made GT images available.  In panel $(c)$, the  edge map selected by $\mathcal{Q}_{\mathfrak{B}}$ by using GT2, GT3, GT4 and GT5 is shown. In panel $(d)$,  the edge map using PFoM with GT5 is shown.

Visual inspection tells us that the maps selected by PFoM and $\mathcal{C}$ are almost identical when the GT is the most detailed one (GT1). But the differences are very striking when PFoM is using GT5 (the least detailed one) as GT, i.e. the map selected lost the structure of the building.

In Table~\ref{table3}, the values of $\mathcal{E}$, $\mathcal{C}$ and $\mathcal{H}$  over the collection of five GT images are shown. Our~$\mathcal{C}$ measure gives the maximum scoring to the most detailed GT.
\begin{table}[h]
\centering
\begin{tabular}{|c|c|c|c|}
  \hline
  GT           & $\mathcal{E}$           & $\mathcal{H}$       & $\mathcal{C}$        \\ \hline
  GT1                 & \textbf{0.8146}        &  \textbf{0.8612}       & \textbf{0.7015 }    \\
  GT2                 & 0.7730        &  0.8292       & 0.6410     \\
  GT3                  & 0.7599        &  0.7852       & 0.5966     \\
  GT4                 & 0.7597        &  0.7817       & 0.5939     \\
 GT5                  & 0.7778        &  0.8040       & 0.6253     \\
  \hline
\end{tabular}
\caption{$\mathcal{E}$, $\mathcal{H}$ and $\mathcal{C}$ scores of all GT images available for the image \emph{86000} .(The best results are highlighted in bold typeface.) }
\label{table3}
\end{table}
In this example, three mayor conclusions are drawn:
\begin{itemize}
\item By using supervised measures, the degree of details of the GT impacts on the quality of the edge map selected. PFoM selects a better map using a detailed GT than using a less detailed GT. Our reference-based measure $\mathcal{Q}_{\mathfrak{B}}$ selects edge maps that are as good as the ones selected by PFoM, and it accommodates the use of a whole collection of GT images to select the best edge map when moving parameters in a fixed range.
\item $\mathcal{C}$ selects edge maps that are as good as the ones selected by PFoM at its best (when the GT is detailed enough) but selects better maps than PFoM when the GT is very simple (almost a boundary). The \emph{Equilibrium} index $\mathcal{E}$ is based on the $\mathcal{Q}_{\mathfrak{B}}$ measure computed over a rich database of patterns; this operation is better than correlating with a simple (or inaccurate)  GT.
\end{itemize}


\begin{figure}[h]
 \[\begin{array}{cccc}
 \includegraphics[scale = 0.18]{./i109.eps}&\includegraphics[scale = 0.18]{./verdad_i109.eps}&\includegraphics[scale = 0.18]{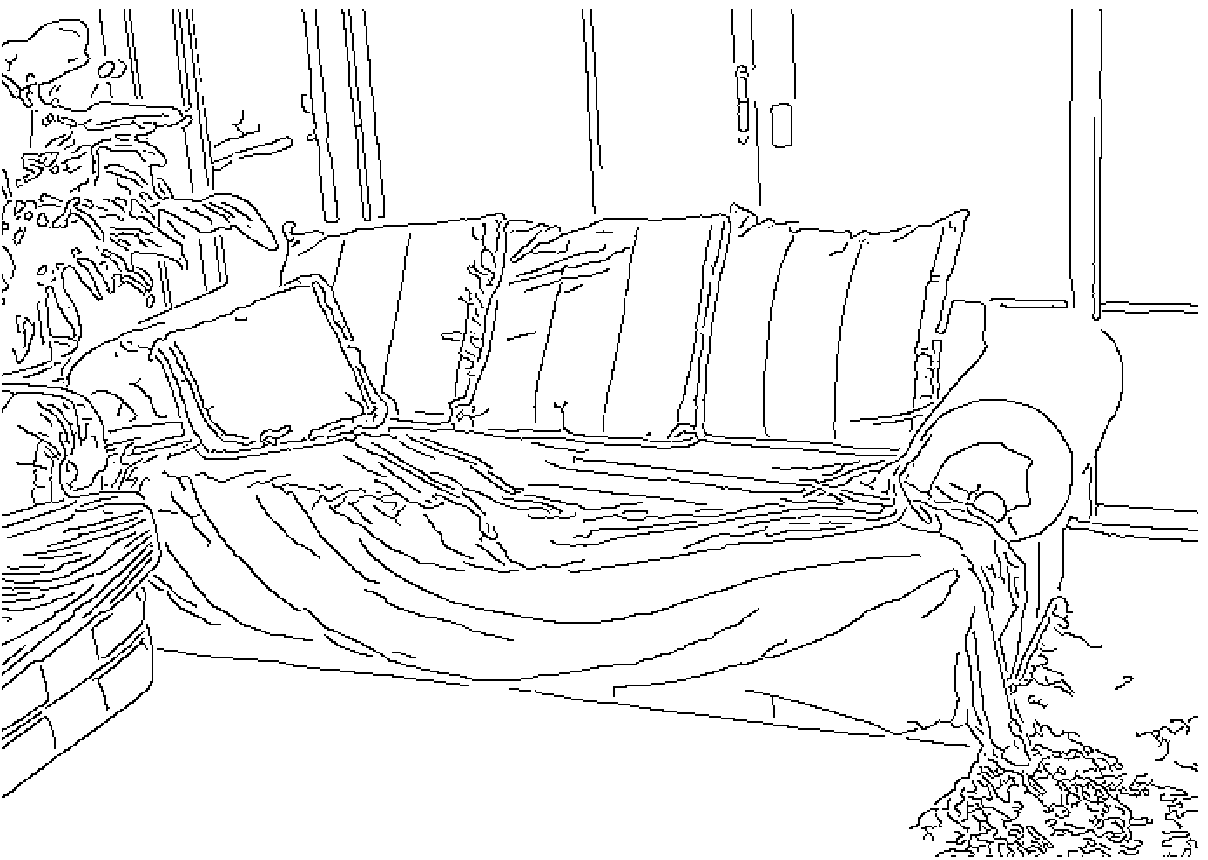}
 &\includegraphics[scale = 0.21]{./graficas_EHC_i109.eps}\\
 \includegraphics[scale = 0.20]{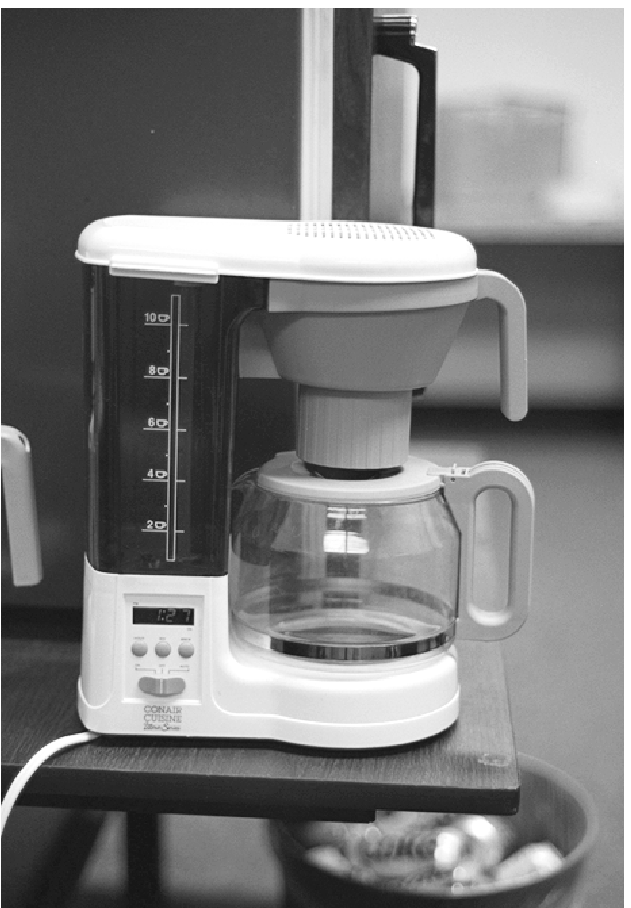}&\includegraphics[scale = 0.20]{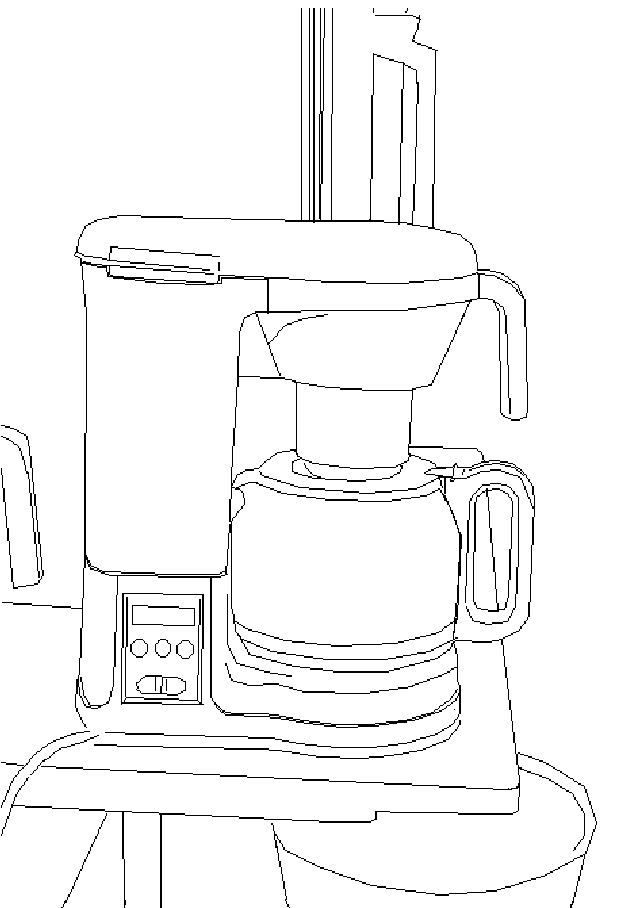}& \includegraphics[scale = 0.20]{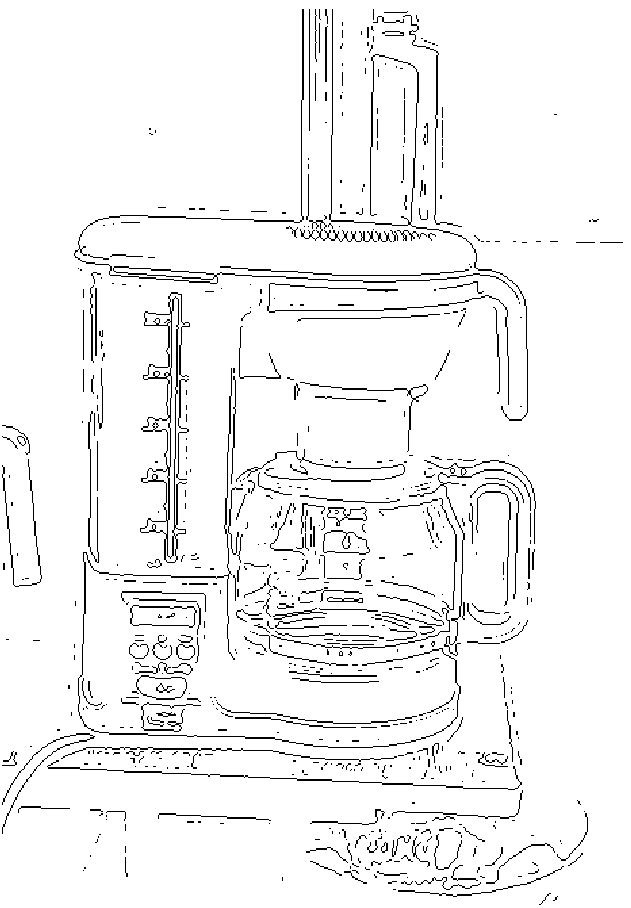}
 &\includegraphics[scale = 0.21]{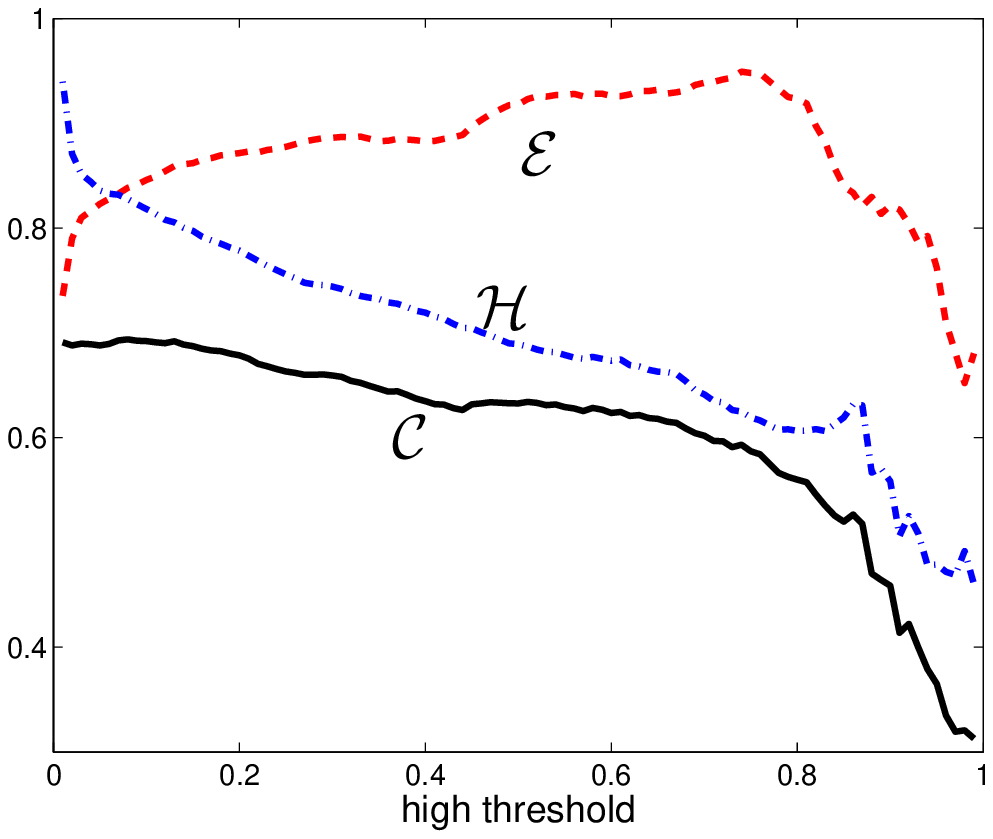}\\
 \includegraphics[scale = 0.18]{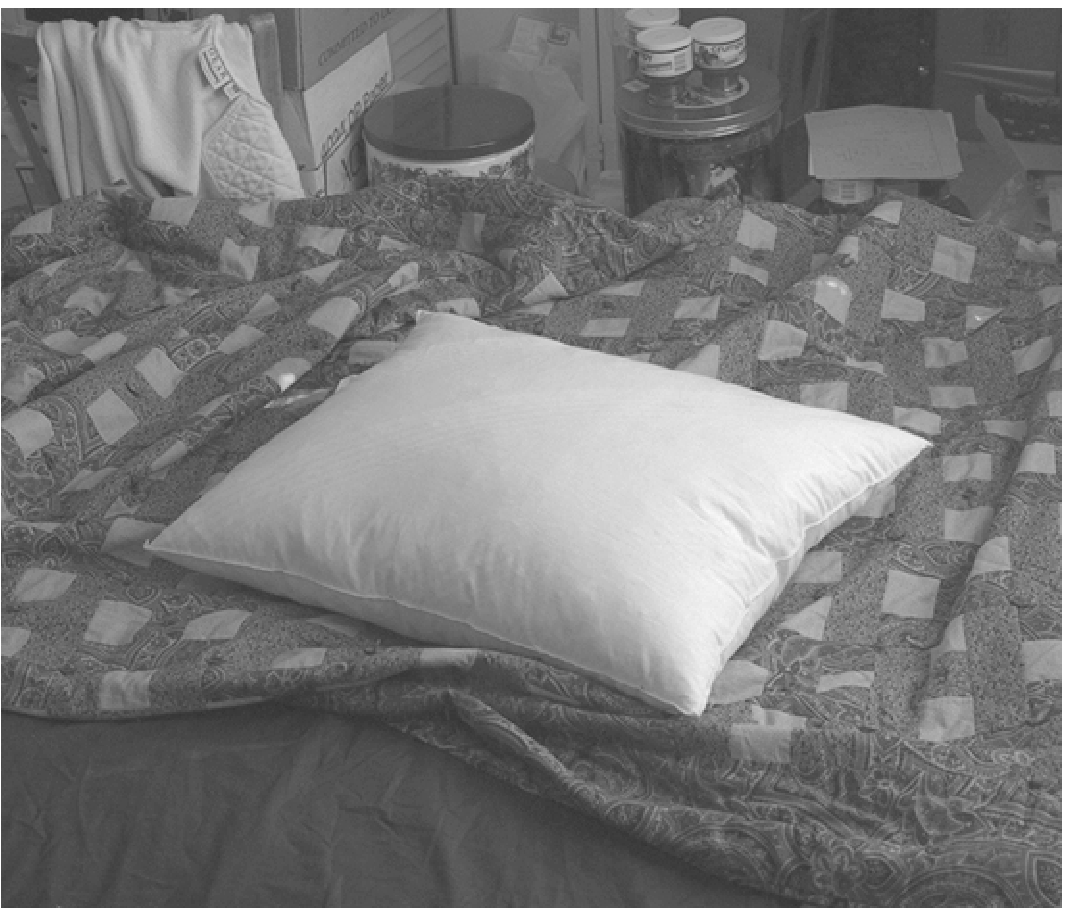}&\includegraphics[scale = 0.18]{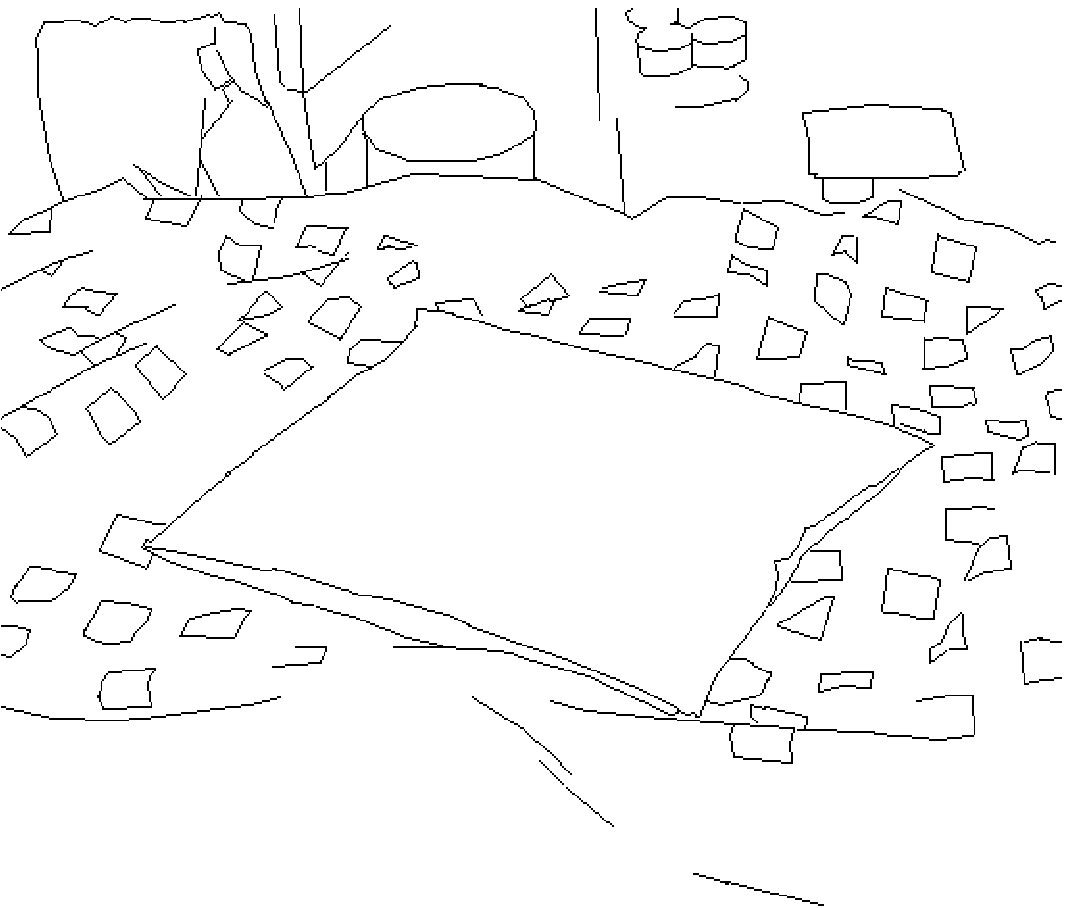}& \includegraphics[scale = 0.18]{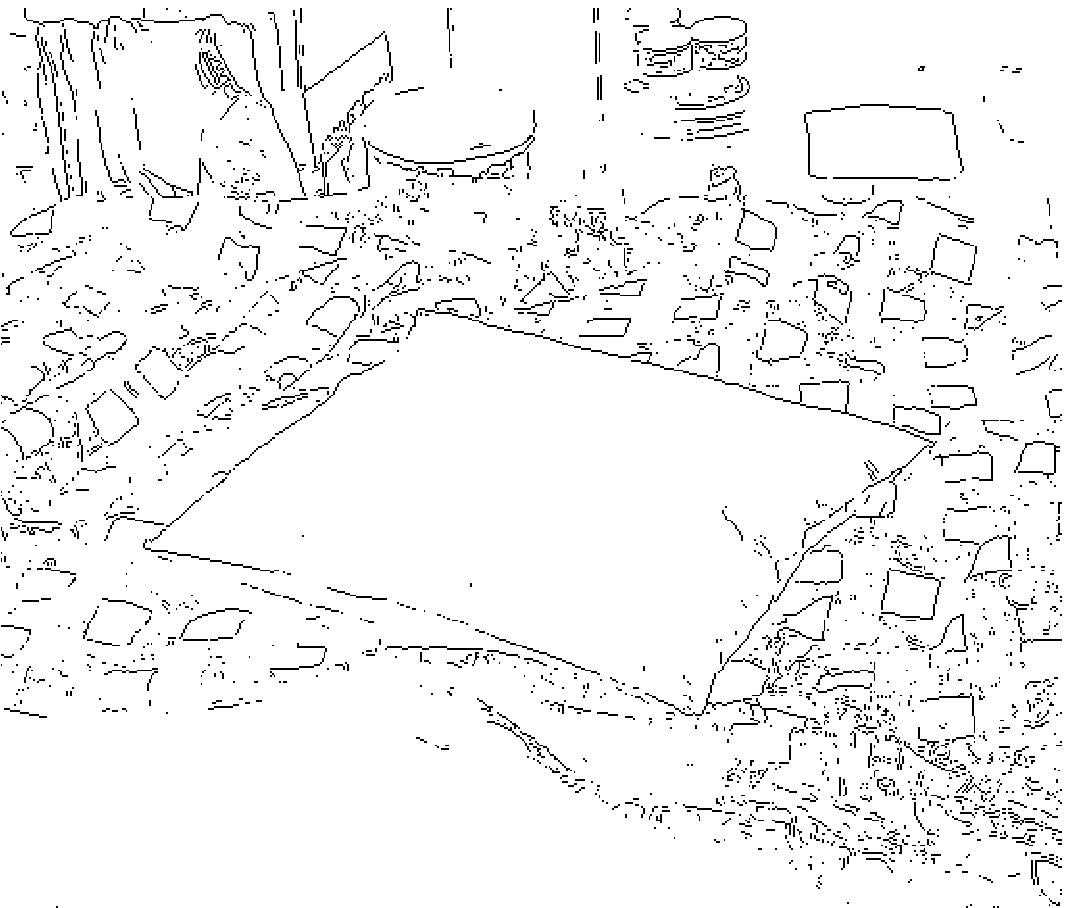}
 &\includegraphics[scale = 0.21]{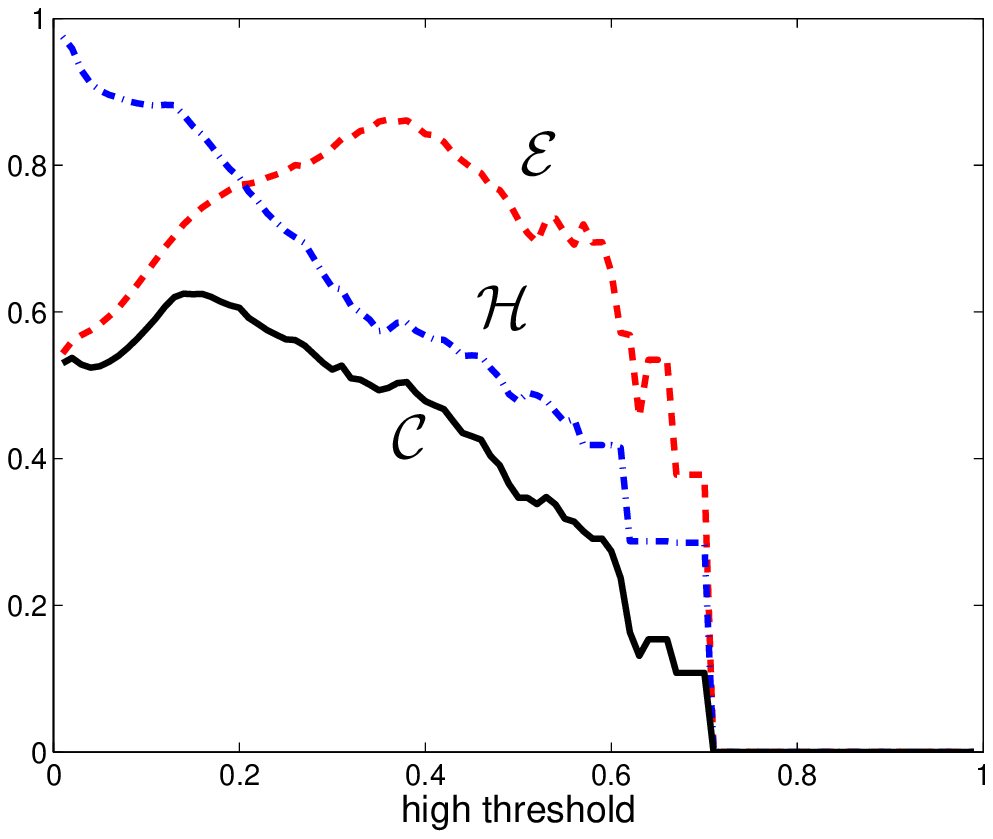}\\
 \includegraphics[scale = 0.18]{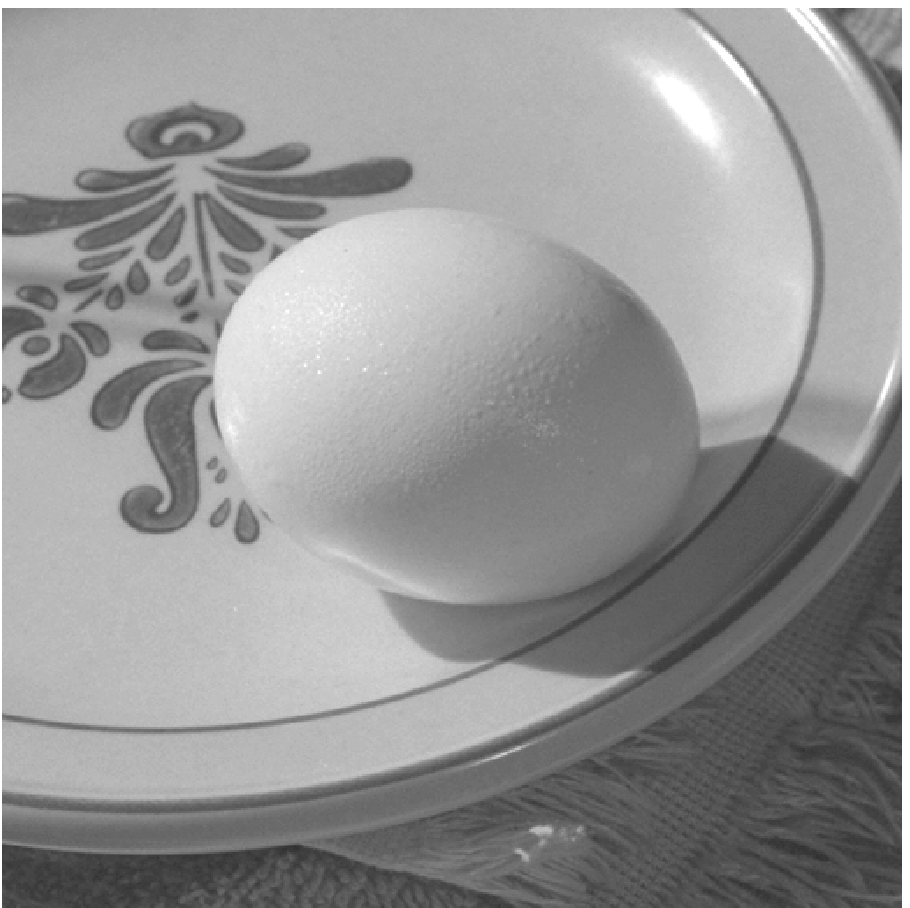}&\includegraphics[scale = 0.18]{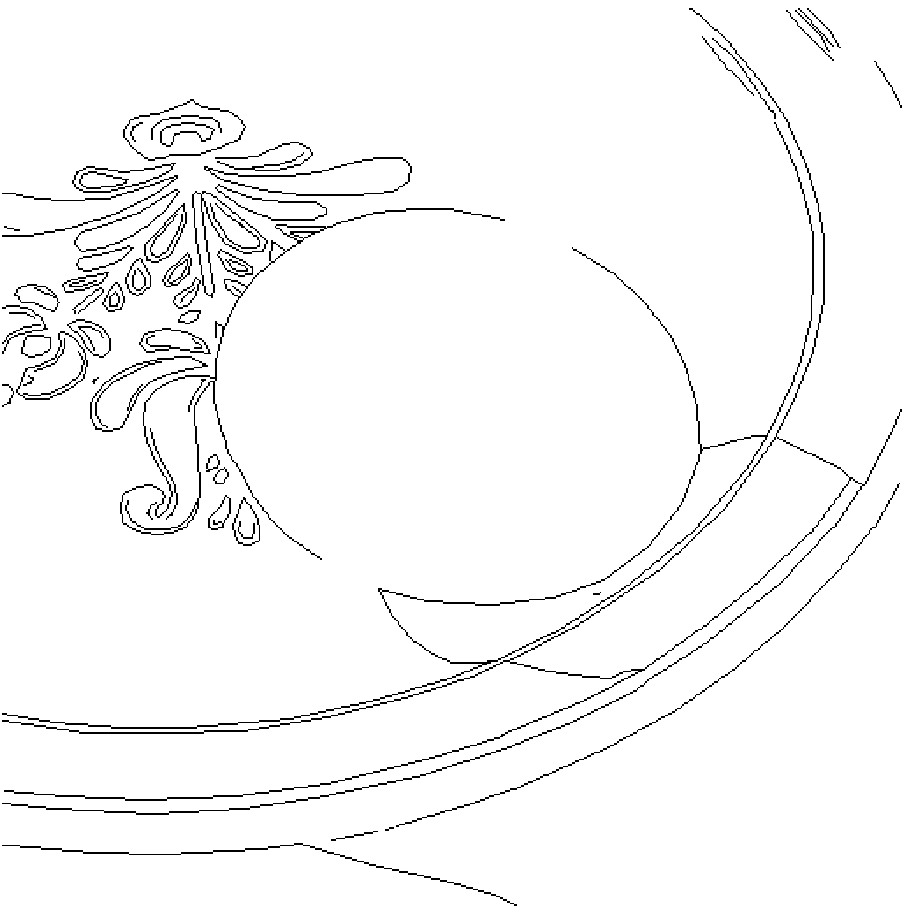}& \includegraphics[scale = 0.18]{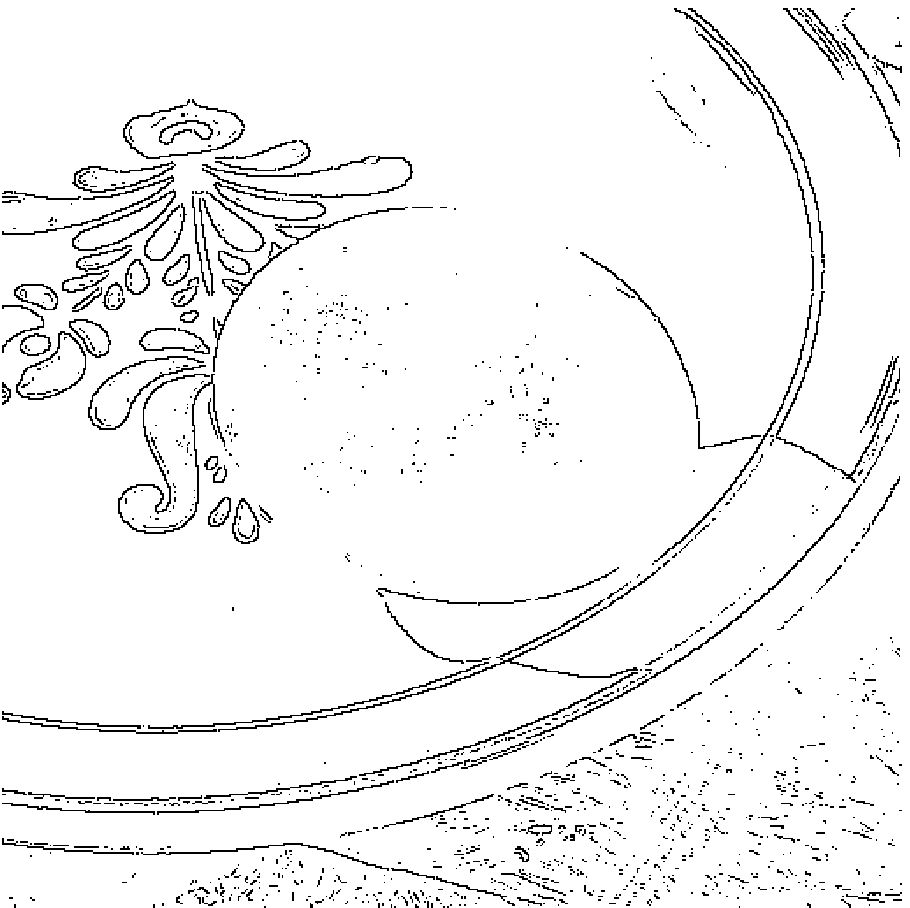}
 &\includegraphics[scale = 0.21]{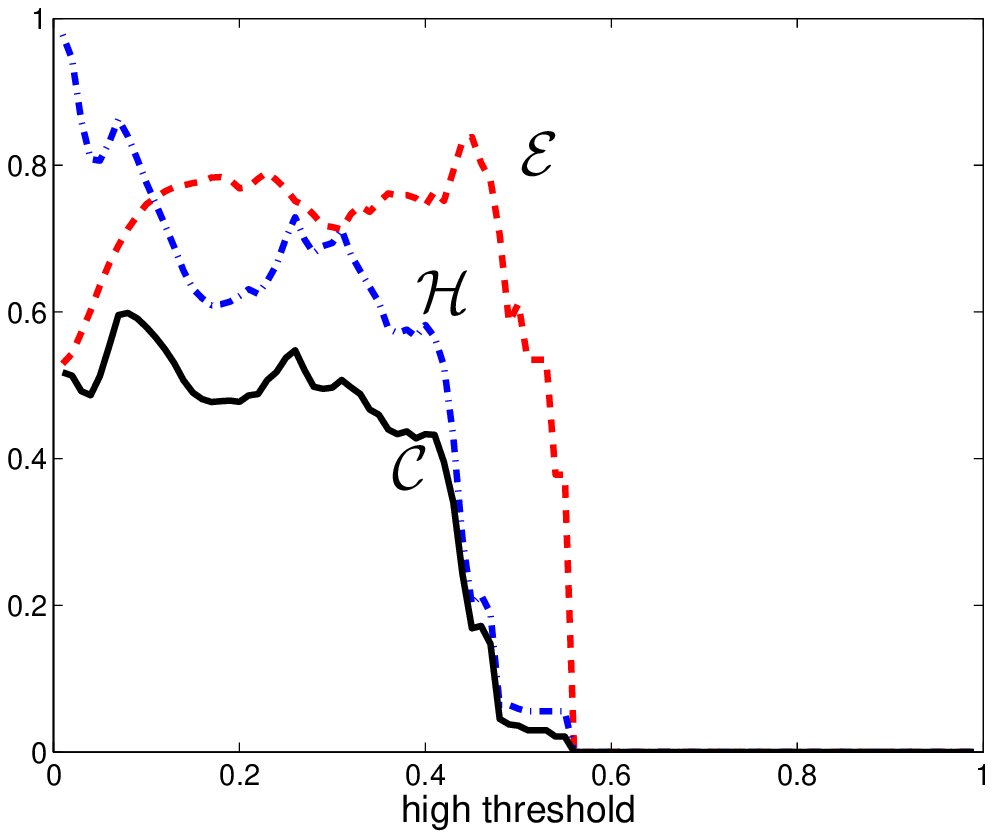}\\
 \includegraphics[scale = 0.18]{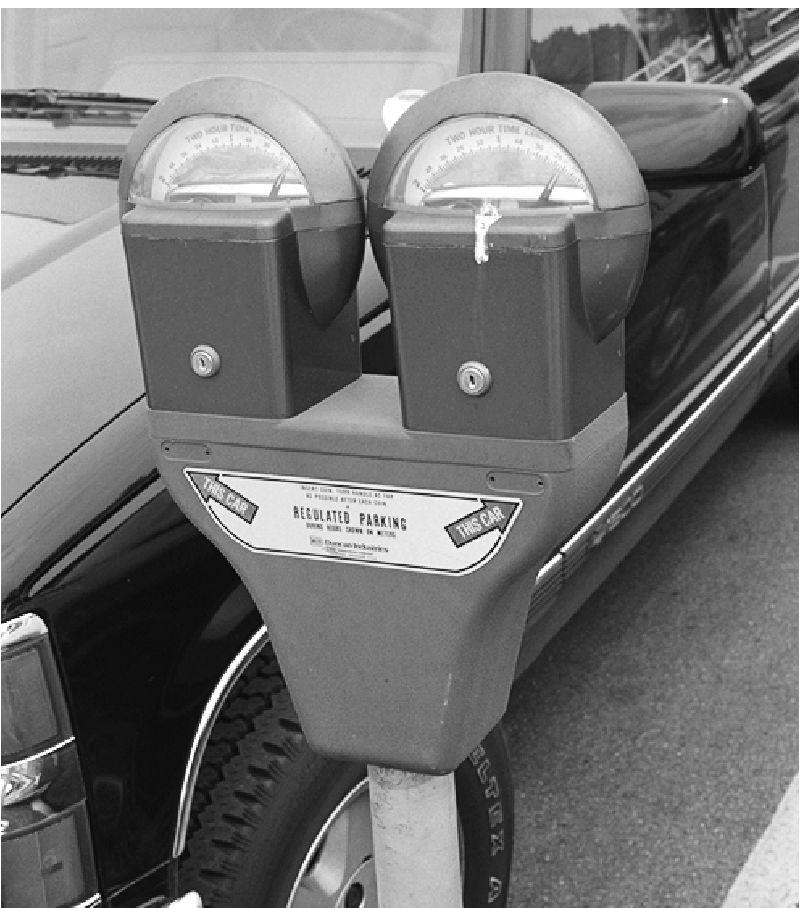}&\includegraphics[scale = 0.18]{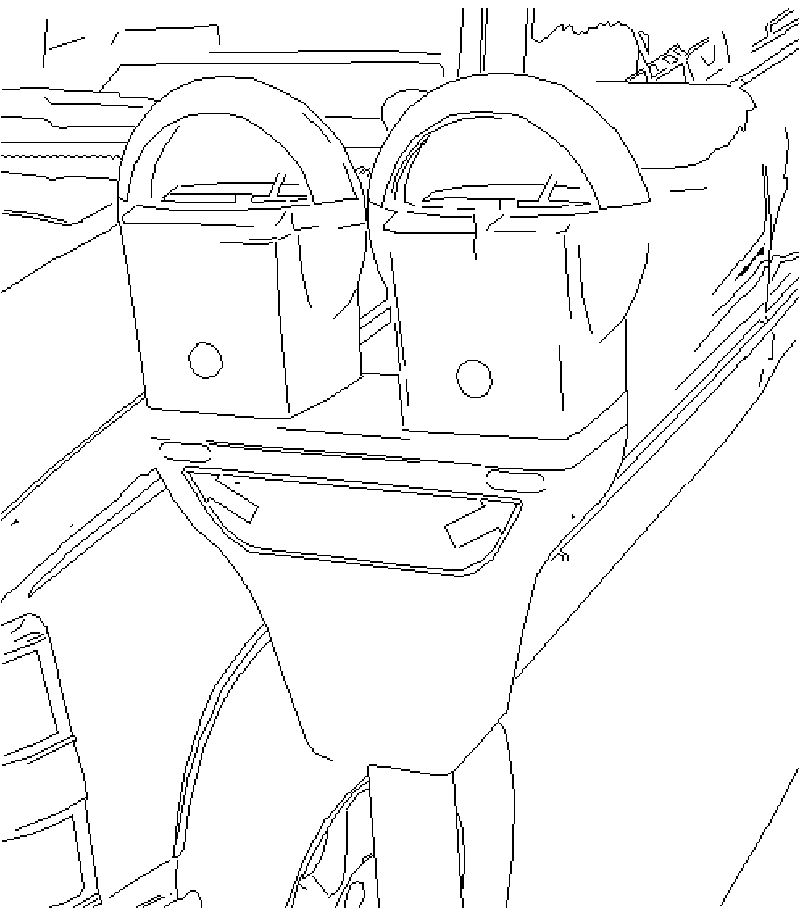}& \includegraphics[scale = 0.18]{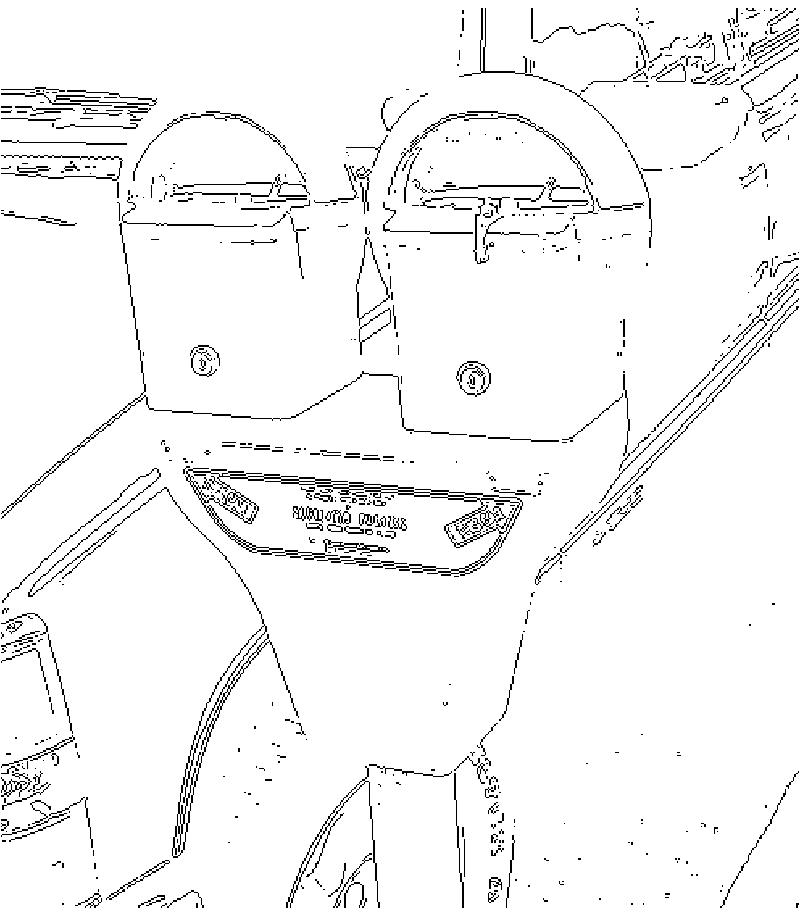}
 &\includegraphics[scale = 0.21]{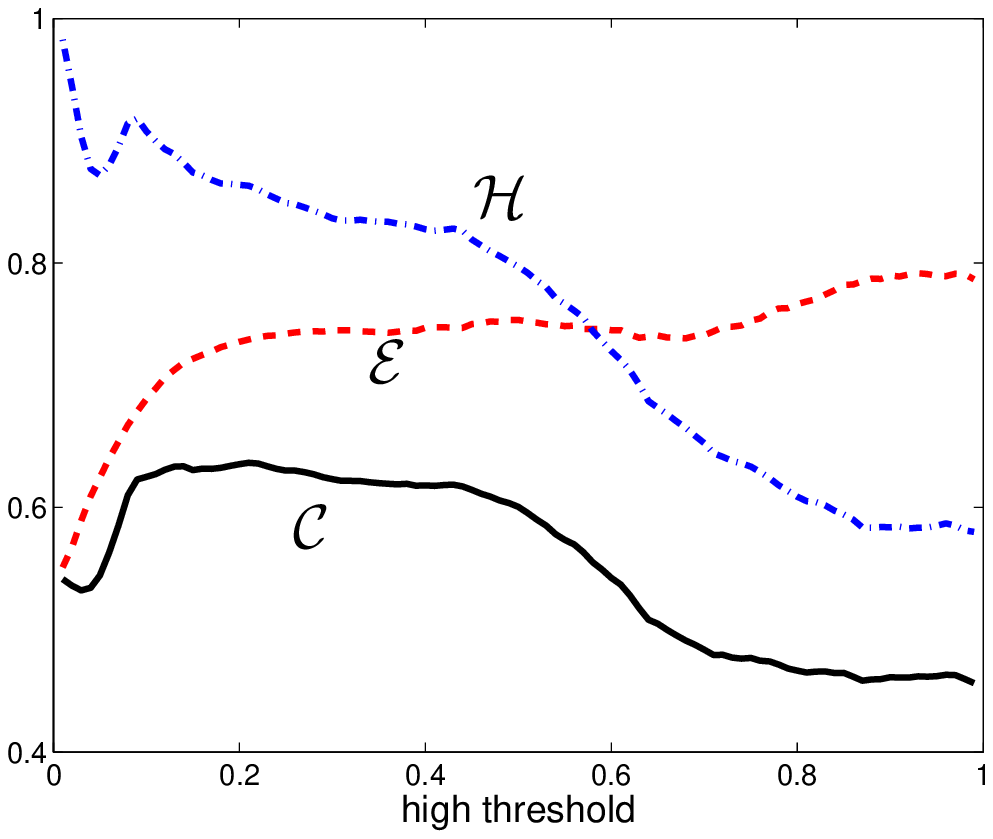}\\
 \end{array} \]
 \vspace{-0.5cm}
 \caption{From top to bottom, images and detectors, \emph{109}, Canny; \emph{coffee}, LoG; \emph{218}, Prewitt; \emph{egg}, Roberts; \emph{parkingmeter},  Sobel. The last panel of each row shows plots of the $\mathcal{C}$ measure vs the threshold values of the corresponding detector.  }
 \label{otras}
\end{figure}

\section{Conclusions}
\label{conclusion}
In this paper, new ideas of edge \emph{Equilibrium} and edge \emph{Information} are discussed.  They lead to the definition of a  new SCM for scoring binary maps.  To measure edge \emph{Equilibrium}, a similarity index  was defined by projecting the edge map into a family of edge patterns that scores the continuity and width of edges in fixed size windows of the edge map. To measure  \emph{Information},   a new {\emph{Entropy} index based on the KS statistic was defined. The SCM is the product of the  \emph{Equilibrium} and  \emph{Entropy} indices and it is effectively used for performance characterization which includes: (i) the specific evaluation of an algorithm (intra-technique process) in order to identify its best parameters, and (ii) the comparison of different algorithms (inter-technique process) in order to classify them according to their quality.

Our experiments were made with common edge detectors that are used by a large number of practitioners. More complex edge detectors aim at specific characteristics in the images, thus the measure should be modified accordingly with a pattern database that accommodates those general characteristics. Active contour methods as applied in~\cite{Giron2012}
are based on the statistical distribution of the noise present in PolSAR images. A measure like ours must carefully be  modified to score such EDA outputs, which is the scope of another paper. We are also studying alternative definitions for the \emph{Entropy} index based on  edge map  histogram functionals  that could be tailored to measure the  performance of boundary detection algorithms more accurately.

%
%
%

\section*{Acknowledgments}

This paper has been partially supported by the Argentinean Grants PICT 2008-00291, and SeCyT-UNC. JM was partially supported by a SGCyT-UNS grad student travel fellowship.  JM wants to thanks Famaf-UNC for its hospitality while the preparation of this manuscript. JG was supported by a Conicet graduate student fellowship.  The authors wants to thank Prof. Alejandro Frery for enriching discussions that lead to the definition of the measure. All measure related Matlab code was written by the authors and it is available to be downloaded from the Reproducible Research repository of AGF at University of Cordoba.

\bibliographystyle{elsart-num}
\bibliography{biblio_final}

\end{document}